\documentclass{article}
\PassOptionsToPackage{numbers, compress}{natbib}
 \usepackage[main, final]{neurips_2025}
\usepackage[utf8]{inputenc} 
\usepackage[T1]{fontenc}
\usepackage[breaklinks]{hyperref}
\usepackage{url}
\usepackage{booktabs}
\usepackage{amsfonts} 
\usepackage{nicefrac}
\usepackage{xcolor}
\usepackage{color_edits}
\usepackage{float}
\usepackage{microtype}
\usepackage{upgreek}
\usepackage{amsbsy}
\usepackage{amssymb}
\usepackage{cancel}
\usepackage{amsmath}
\usepackage{amsthm}
\usepackage{graphicx}
\usepackage{setspace}
\usepackage{titlesec}
\usepackage{enumitem}
\usepackage{caption}
\usepackage{subcaption}
\usepackage{adjustbox,multirow}
\usepackage[noabbrev]{cleveref}
\usepackage{algorithm}
\usepackage{algpseudocode}
\usepackage{wrapfig}
\usepackage{titletoc}
\usepackage{bbm}
\usepackage{mathtools}
\usepackage{mleftright}

\makeatletter
\newcommand{\newreptheorem}[2]{%
    \newtheorem*{rep@#1}{\rep@title}%
    \newenvironment{rep#1}[1]{%
        \def\rep@title{#2 \ref{##1}}%
        \begin{rep@#1}}%
    {\end{rep@#1}}}
\makeatother

\newtheorem{theorem}{Theorem}
\newreptheorem{theorem}{Theorem}

\newreptheorem{corollary}{Corollary}

\newtheorem{lemma}[theorem]{Lemma}

\newtheorem{proposition}[theorem]{Proposition}
\newreptheorem{proposition}{Proposition}

\theoremstyle{definition}
\newtheorem{definition}{Definition}
\newreptheorem{definition}{Definition}

\newtheorem{assumption}{Assumption}

\crefname{equation}{}{}
\crefname{proposition}{Proposition}{Propositions}
\crefname{section}{Section}{Sections}
\crefname{appendix}{Appendix}{Appendices}
\crefname{lemma}{Lemma}{Lemmas}
\crefname{assumption}{Assumption}{Assumptions}
\crefname{algorithm}{Algorithm}{Algorithms}
\crefname{theorem}{Theorem}{Theorems}
\crefname{corollary}{Corollary}{Corollaries}
\crefname{insight}{Insight}{Insights}
\crefname{definition}{Definition}{Definitions}
\crefname{figure}{Figure}{Figures}
\crefname{table}{Table}{Tables}

\DeclareMathOperator{\E}{\mathbb{E}}
\DeclareMathOperator{\var}{\mathrm{Var}}
\newcommand{\R}{\mathbb{R}}

\newcommand{\dist}{P}

\def\ddefloop#1{\ifx\ddefloop#1\else\ddef{#1}\expandafter\ddefloop\fi}
\def\ddef#1{\expandafter\def\csname bb#1\endcsname{\ensuremath{\mathbb{#1}}}}
\ddefloop ABCDEFGHIJKLMNOPQRSTUVWXYZ\ddefloop
\def\ddefloop#1{\ifx\ddefloop#1\else\ddef{#1}\expandafter\ddefloop\fi}
\def\ddef#1{\expandafter\def\csname b#1\endcsname{\ensuremath{\mathbf{#1}}}}
\ddefloop ABCDEFGHIJKLMNOPQRSTUVWXYZ\ddefloop
\def\ddef#1{\expandafter\def\csname c#1\endcsname{\ensuremath{\mathcal{#1}}}}
\ddefloop ABCDEFGHIJKLMNOPQRSTUVWXYZ\ddefloop
\def\ddef#1{\expandafter\def\csname h#1\endcsname{\ensuremath{\hat{#1}}}}
\ddefloop ABCDEFGHIJKLMNOPQRSTUVWXYZ\ddefloop
\def\ddef#1{\expandafter\def\csname wh#1\endcsname{\ensuremath{\widehat{#1}}}}
\ddefloop abcdefghijklmnopqrstuvwxyz\ddefloop
\def\ddef#1{\expandafter\def\csname hc#1\endcsname{\ensuremath{\widehat{\mathcal{#1}}}}}
\ddefloop ABCDEFGHIJKLMNOPQRSTUVWXYZ\ddefloop
\def\ddef#1{\expandafter\def\csname t#1\endcsname{\ensuremath{\widetilde{#1}}}}
\ddefloop ABCDEFGHIJKLMNOPQRSTUVWXYZ\ddefloop
\def\ddef#1{\expandafter\def\csname tc#1\endcsname{\ensuremath{\widetilde{\mathcal{#1}}}}}
\ddefloop ABCDEFGHIJKLMNOPQRSTUVWXYZ\ddefloop

\usepackage{bm}
\makeatletter
\newcommand{\vect}[1]{%
  \ifcat\noexpand#1\relax
    \bm{#1}
  \else
    \mathbf{#1}
  \fi
}
\makeatother

\usepackage{prettyref}
\usepackage{xcolor}

\newcommand{\midsem}{\nonscript\;;\nonscript\;\mathopen{}}

\newcommand*\diff{\mathop{}\!\mathrm{d}}

\newcommand{\kl}[2]{\mathrm{D}_\mathrm{KL}\left(#1 \ \middle\| \ #2\right)}

\newcommand{\data}{\mathcal{D}}

\newcommand{\indep}{\perp\!\!\!\!\perp} 
\newcommand{\given}{\nonscript\;\middle|\nonscript\;\mathopen{}}

\makeatletter
\DeclarePairedDelimiter{\absplain}{\lvert}{\rvert}
\DeclareRobustCommand{\abs}{\@ifstar{\absplain}{\absplain*}}
\makeatother
\DeclareRobustCommand{\lpar}{\mleft(}
\DeclareRobustCommand{\rpar}{\mright)}
\DeclareRobustCommand{\lbar}{\mleft[}
\DeclareRobustCommand{\rbar}{\mright]}
\DeclareRobustCommand{\lcbar}{\mleft\{}
\DeclareRobustCommand{\rcbar}{\mright\}}

\newcommand{\xhdr}[1]{\textbf{#1.}}

\newcommand{\obs}{\mathrm{obs}}

\newcommand{\bernoulli}{\mathrm{Bernoulli}}
\newcommand{\sigmoid}{\mathrm{Sigmoid}}

\hypersetup{
    colorlinks,
    linkcolor={red!100!black},
    citecolor={blue!100!black},
    urlcolor={blue!80!black}
}
\usepackage[most]{tcolorbox} 
\definecolor{orangesignal}{HTML}{FE6100}
\definecolor{highlightsignal}{HTML}{FFB000}
\definecolor{purplesignal}{HTML}{785ef0}
\definecolor{bluesignal}{HTML}{648fff}
\definecolor{pinksignal}{HTML}{dc267f}

\newcommand{\firstbest}[1]{\textbf{\underline{\color{bluesignal} #1}}}
\newcommand{\secondbest}[1]{\textbf{{\color{purplesignal} #1}}}

\newcommand{\stepindicator}[1]{%
  {\textbf{\textit{(#1)}}}%
}

\definecolor{citationcolor}{HTML}{648fff} 
\definecolor{linkcolorcustom}{HTML}{648fff}

\hypersetup{ 
    colorlinks=true,       
    linkcolor=pinksignal,
    citecolor=citationcolor, 
    urlcolor=linkcolorcustom 
}

\newcommand{\method}{CausalPFN}

\title{\method:~Amortized Causal Effect Estimation\\via In-Context Learning}

\author{%
  \hspace{-6pt}Vahid Balazadeh\thanks{Equal Contribution}\;\,{}$^{1}$$^{2}$ \ \ \ Hamidreza Kamkari$^*$$^{3}$ \ \ \ Valentin Thomas$^{3}$ \ \ \ Benson Li$^{1}$$^{2}$ \ \ \ Junwei Ma$^{3}$ \\ \textbf{Jesse C. Cresswell$^{3}$ \ \ \ Rahul G. Krishnan$^{1}$$^{2}$}\\
  \texttt{ vahid@cs.toronto.edu, hamid@layer6.ai} \\
  {}$^1$ University of Toronto\quad {}$^2$ Vector Institute \quad 
  {}$^3$ Layer 6 AI
}

\begin{document}

\maketitle
\begin{abstract}
    Causal effect estimation from observational data is fundamental across various applications. However, selecting an appropriate estimator from dozens of specialized methods demands substantial manual effort and domain expertise. We present \method, a single transformer that \emph{amortizes} this workflow: trained once on a large library of simulated data-generating processes that satisfy ignorability, it infers causal effects for new observational datasets out of the box. \method\ combines ideas from Bayesian causal inference with the large-scale training protocol of prior-fitted networks (PFNs), learning to map raw observations directly to causal effects without any task-specific adjustment. Our approach achieves superior average performance on heterogeneous and average treatment effect estimation benchmarks (IHDP, Lalonde, ACIC). Moreover, it shows competitive performance for real-world policy making on uplift modeling tasks. \method\ provides calibrated uncertainty estimates to support reliable decision-making based on Bayesian principles. This ready-to-use model requires no further training or tuning and takes a step toward automated causal inference (\href{https://github.com/vdblm/CausalPFN/}{https://github.com/vdblm/CausalPFN}).
\end{abstract}

\section{Introduction}
\begin{wrapfigure}{r}{0.53\linewidth}\captionsetup{font=footnotesize, labelformat=simple, labelsep=colon}
  \vspace{-35pt}  
  \centering
  \includegraphics[width=\linewidth]{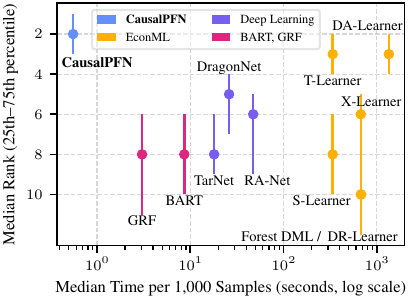}
  \vspace{-12pt}  
  \caption{{\bf Time vs. Performance.} Comparison across 310 causal inference tasks from IHDP, ACIC, and Lalonde. \method\ achieves the best average rank (by precision in estimation of heterogeneous effect) while being much faster in \emph{wall-clock time from data to estimates}.}
  \label{fig:teaser}
  \vspace{-18pt}   
\end{wrapfigure}
Causal inference—estimating the effects of interventions from data—is fundamental across numerous domains, including public policy, economics, and healthcare \citep{manski1993identification, angrist2014mastering, imbens2015causal}. The central challenge lies in estimating causal quantities from observational data: records collected without explicit interventions, where confounding factors can obscure true causal effects. Various causal identification settings have emerged to address this challenge~\citep{angrist1995identification,angrist1996identification,balke1997bounds,mackinnon2007mediation}. Perhaps the most common one is to assume no unobserved confounding (ignorability or backdoor)~\citep{rubin1974estimating,pearl2009causality}.

Even within the conceptually straightforward ignorability framework, researchers have developed dozens of specialized causal estimators over the past four decades. Prominent examples include Meta-Learners~\citep{kunzel2019metalearners}, doubly robust methods~\citep{funk2011doubly,kennedy2020optimal}, double/debiased machine learning (DML)~\citep{chernozhukov2017doubledebiased,foster2023orthogonal}, and neural network approaches~\citep{shalit2017estimating,shi2019adapting,curth2021catenets,curth2021inductive,curth2021nonparametric,curth2021really}, among others~\citep{rosenbaum1983central,konstantinov2023heterogeneous,ma2024diffpo,liu2024dag}. This large number of estimators creates practical challenges as domain expertise is required to select, tune, or design the most appropriate estimator for each application~\citep{shimoni2018benchmarking,schuler2018comparison,dorie2019automated,alaa2019validating,neal2020realcause,mahajan2024empirical}. 

The Bayesian paradigm offers an elegant framework to address these challenges~\citep{rubin1978bayesian,imbens1997bayesian,imbens2015causal,hill2011bayesian,balazadeh2024sequential}; rather than manually designing or selecting the best estimator, one can: (1) parameterize an appropriate prior distribution over plausible underlying causal mechanisms, i.e., the data-generating processes (DGPs), (2) define the causal estimand as a functional of the DGP parameters, (3) compute a posterior distribution over DGPs conditioned on observed data, and (4) derive the posterior predictive distribution (PPD) of the causal estimand. 
However, the practical adoption of Bayesian methods remains limited. Computing posterior distributions typically requires expensive sampling methods~\citep{oganisian2021practical,hill2011bayesian}, which often leads researchers to make specific assumptions about the DGPs or priors that are not necessarily reflective of the complexity of the downstream tasks~\citep{hahn2020bayesian,li2023bayesian}.

Meanwhile, an emerging area in deep in-context learning suggests using large models that can approximate PPDs by taking the entire list of observations as context and amortize the expensive process of posterior inference~\citep{garnelo2018neural,garnelo2018conditional,kim2019attentive}. A successful example is the prior-fitted network (PFN) \citep{müller2023transformers} that achieved remarkable performance in tabular prediction tasks~\citep{hollmann2023tabpfn, ma2024tabdpt, helli2024drift, hollmann2025accurate, ye2025closer, mccarter2025exactly, liu2025tabpfn}. PFNs employ transformer architectures trained on large-scale simulated DGPs, representing a rich prior, to perform posterior predictive inference via in-context learning; given a dataset of input-output examples as context, they can predict outputs for new inputs. PFNs shift the computational burden from inference time to (pre-)training, producing a single set of model parameters that can make fast and accurate predictions on unseen datasets. However, they are only designed for regression and classification, not for causal inference.

We propose to bridge the large-scale training of amortized models with Bayesian causal inference and introduce \method, a transformer model for causal effect estimation via in-context learning. Our framework leverages a general-purpose prior, based on the \emph{ignorability} assumption, to generate a vast collection of simulated DGPs. By training on these diverse DGPs, our method learns to infer the causal estimands directly from observational data. While our approach requires an expensive pre-training phase, once complete it is ready for inference on new datasets with no further training, fine-tuning, or hyperparameter optimization. Hence, \method\ is easy-to-use, efficient for inference, and shows remarkably strong performance as an estimator. 
\cref{fig:teaser} illustrates the relative performance and efficiency of our method compared to standard baselines. For inference on an unseen dataset, \method\ requires only forward passes, whereas baseline methods have additional costs including hyperparameter tuning or cross-validation. We therefore report the computational time for all of these stages for the baselines to reflect the total costs of predicting on a new dataset.

\begin{figure}\captionsetup{font=footnotesize, labelformat=simple, labelsep=colon}
    \centering
    \vspace{-20pt}
    \includegraphics[width=1\linewidth]{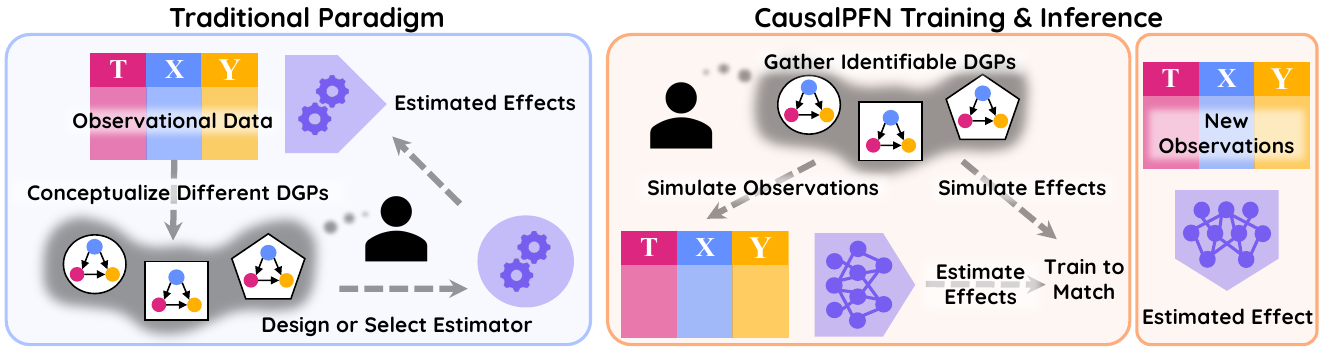}
    \caption{\textbf{Traditional Causal Inference vs. \method}. \textit{(Left)}: A domain expert selects or tunes an estimator for a DGP that they deem appropriate for the given data. \textit{(Right)}: The domain expert simulates diverse DGPs for pre-training, and a transformer learns to amortize causal inference automatically.}
    \label{fig:main_figure}
    \vspace{-20pt}
\end{figure}

We show \method's workflow compared to traditional causal inference in \cref{fig:main_figure}. Our key contributions are: 
\stepindicator{i} To our knowledge, for the first time, we demonstrate that a single transformer-based model trained on a diverse library of simulated DGPs can match or surpass specialized estimators across multiple datasets without task-specific tuning. Specifically, \method\ achieves the best average rank on CATE across IHDP, ACIC, and Lalonde, and competitive ATE performance, without task-specific tuning.
\stepindicator{ii} We highlight \method's competitive out-of-the-box performance for real-world policy making on various uplift modeling tasks.
\stepindicator{iii} We theoretically characterize the assumptions under which \method's estimates are asymptotically consistent. 
\stepindicator{iv} We develop a principled uncertainty quantification framework for \method\ to produce finite-sample calibrated credible intervals for the estimates. 
\stepindicator{v} Finally, we release our model's weights with a user-friendly API, streamlining the adoption of \method\ as a capable estimator. \method\ is fast, ready-to-use, and does not require any further training or hyperparameter tuning.

\section{Background} \label{sec:background}

\xhdr{Causal Effect Estimation} 
We adopt the potential–outcomes framework for causal inference \citep{rubin2005causal}.  
Let $T \in \cT$ denote the treatment from a finite treatment set $\cT$, and $\vect{X} \in \cX$ the observed covariates. For every $t\in\cT$, $Y_t \in \R$ is the potential outcome under treatment $t$, while the observed (factual) outcome is $Y\coloneqq Y_T$.  
We call the joint distribution $\dist(\vect{X},T,\{Y_t\}_{t\in\cT},Y)$ the \emph{data-generating process} (DGP), and denote by $\dist_\obs$ the marginal distribution over observed triples $(\vect{X}, T, Y)$. Given samples from $\dist_\obs$, a central goal is to recover the \emph{conditional expected potential outcomes} (CEPOs):
\begin{equation}
    \mu_t(\vect{x}) \coloneqq \E\lbar Y_t \given \vect{X} = \vect x \rbar, \qquad \forall t \in \cT,\; \vect x \in \cX.
    \label{eq:cepo}
\end{equation}
For binary treatments, two common estimands, average treatment effect (ATE), and conditional average treatment effect (CATE) follow directly from the CEPOs. We refer to CEPOs, CATE, and ATE collectively as \emph{causal effects}. 
\begin{align}
    \text{ATE}: \quad & \lambda  \coloneqq \E\lbar Y_1 - Y_0 \rbar  = \E\lbar \mu_1(\vect{X}) - \mu_0(\vect{X}) \rbar , \label{eq:ate} \\
    \text{CATE}: \quad & \tau(\vect{x}) \coloneqq \E\lbar Y_1 - Y_0 \given \vect{X} = \vect x \rbar  = \mu_1(\vect{x}) - \mu_0(\vect{x}). \label{eq:cate}
\end{align}
Estimating causal effects from observational data is impossible without further assumptions: different DGPs can induce the same $\dist_\obs$ but have different causal effects~\citep{pearl2009causality,hernan2010causal,imbens2015causal}. We thus define:
\begin{definition}[CEPO-Identifiability]\label{def:identification}
For each $t \in \cT$, CEPO-identifiability holds when $\mu_t$ can be written as a functional of the observational distribution $\dist_\obs$.
\end{definition}
Throughout, we assume \emph{strong ignorability}, a standard \emph{sufficient} assumption that makes CEPOs identifiable. Strong ignorability posits that, conditional on observed covariates, treatment assignment has positive probability for all $t\in \cT$ and is independent of all potential outcomes~\citep{rubin1974estimating, rosenbaum1983central, peters2017elements}:
\begin{assumption}[Strong Ignorability]\label{assumption:ignorability}
    \stepindicator{i} $Y_{t} \indep T \mid \vect{X}$ for all $t \in \cT$ (Unconfoundedness), and \stepindicator{ii} $\dist\lpar T=t \given \vect{X}\rpar > 0$ a.e. for all $t \in \cT$ (Positivity).
\end{assumption}

\xhdr{Bayesian Causal Inference}
A Bayesian formulation of causal inference considers an explicit likelihood model for the DGP~\citep{rubin1978bayesian,oganisian2021practical,li2023bayesian}. Let $\psi$ be the parameter that indexes the DGPs $\dist^\psi\lpar \vect{X}, T, \{Y_t\}_{t \in \cT}, Y\rpar$. A prior $\pi(\psi)$ encodes domain knowledge on parameters $\psi$. Given i.i.d. observations $\data_\obs = \lcbar (\vect{x}^{(n)}, t^{(n)}, y^{(n)}) \rcbar_{n=1}^N$ coming from the observational distribution $\dist^\psi_\obs$, Bayes' rule yields the posterior $\pi \lpar \psi \given \data_\obs \rpar$. For any functional $g(\psi)$—for example $g(\psi) =  \E^\psi\lbar Y_1 - Y_0 \rbar $ for ATE—the posterior predictive distribution (PPD) 
\begin{equation}
    \pi^g \lpar \cdot \given \data_\obs \rpar \coloneqq \lbar B \;\mapsto\;  \int \bbI \left( g(\psi) \in B \right) \pi\lpar \psi \given \data_\obs\rpar \diff\psi\rbar, \qquad B \in \cB,
\end{equation}
is induced by the posterior distribution $\pi \lpar \psi \given \data_\obs \rpar$ ($\cB$ denotes the Borel $\sigma\text{-algebra}$ over $\mathbb{R}$). Point estimates (posterior means) and credible intervals therefore arise automatically from these induced posteriors. Because the posterior is rarely available in closed form, one resorts to approximate inference such as Markov-chain Monte-Carlo (MCMC)~\citep{hill2011bayesian} or variational inference~\citep{louizos2017causal,jesson2020identifying}. Such techniques have been applied with flexible priors including nonparametric BART models~\citep{hill2011bayesian,hahn2020bayesian}, Dirichlet processes~\citep{linero2023and} and Gaussian processes~\citep{alaa2017bayesian}. In summary, the Bayesian paradigm offers a unified framework for inference on causal estimands and provides automatic uncertainty quantification.

\xhdr{Amortizing Posterior Predictive Inference with Prior-Fitted Networks}
Running a new posterior inference for every dataset is computationally demanding, especially with high-dimensional covariates~\citep{hahn2020bayesian,li2023bayesian}. Recent work shows that in-context transformers can \emph{amortize} Bayesian prediction: instead of sampling from the posterior at test time, a single network is trained to map a context set directly to the PPD~\citep{garnelo2018conditional, garnelo2018neural, kim2019attentive,müller2023transformers,helli2024drift}. PFNs instantiate this idea for supervised learning~\citep{hollmann2023tabpfn}.

Consider a supervised dataset $\data^{\text{SL}} = \{(\vect{x}^{(n)}, y^{(n)})\}_{n=1}^N$ and a prior $\pi^{\text{SL}}$ on parameters $\phi$ indexing $\dist^\phi\lpar \vect{X}, Y\rpar$. The Bayesian approach to predict the output for a new input $\vect{x}$ is to use the PPD
\begin{equation}\label{eq:ppd}
   \mathrm{PPD} \lpar Y \given \vect{X} = \vect{x}, \data^{\text{SL}} \rpar := \int \dist^\phi \lpar Y \given \vect{X} = \vect{x} \rpar \pi^{\text{SL}}\lpar \phi \given \data^{\text{SL}}\rpar \diff \phi.
\end{equation}
Rather than approximating the posterior distribution $\pi^{\text{SL}}\lpar \phi \given \data^{\text{SL}} \rpar$ with MCMC or variational inference~\citep{jordan1999introduction,andrieu2003introduction,neal2012bayesian}, PFNs directly parameterize the PPD using a single transformer model $q_\theta\lpar Y \given \vect{X}, \data^{\text{SL}}\rpar$ by minimizing the \emph{data-prior loss}
\begin{equation} \label{eq:data-prior-loss}
    \ell_\theta := \E_{\phi \sim \pi^{\text{SL}},\ \data^{\text{SL}} \cup \{ \vect{X}, Y\} \sim \dist^\phi} \lbar -\log q_\theta \! \lpar Y \given \vect{X}, \data^{\text{SL}}  \rpar \rbar.
\end{equation}
Crucially, training requires only \emph{prior} samples $(\phi,\data^{\text{SL}})$; no posterior sampling is needed.  
With a suitably rich prior, a single PFN can be applied \emph{off-the-shelf} to diverse predictive problems~\citep{ma2024tabdpt,hollmann2025accurate}.

\section{The Mathematical Framework of \method} \label{sec:method}
Our primary estimands of interest are the CEPOs from \cref{eq:cepo}. As shown in \cref{eq:ate} and \cref{eq:cate}, CEPOs directly enable estimation of both ATE and CATE. Therefore, we focus on developing an estimator that can accurately infer these quantities from the observational data. Specifically, we follow the Bayesian paradigm for causal inference, as introduced in \cref{sec:background}, and parameterize CEPOs as $\mu_t\lpar \vect x\midsem \psi\rpar$.
Given a suitably rich prior distribution $\pi$ over the DGPs, which we will explicitly design in \cref{sec:implementation}, we define our target as the posterior predictive distribution of CEPOs:

\begin{definition}[CEPO-PPD] \label{def:cepo-ppd}
For each $t \in \cT$ and covariate vector $\vect{x}$, the \emph{CEPO-PPD} is
\begin{equation} \label{eq:cppd}
\pi^{\mu_t}\lpar \cdot \given \vect{x}, \data_\obs \rpar
\;\coloneqq\;
\lbar B \mapsto \int \bbI\lpar \mu_t(\vect{x}\midsem \psi) \in B \rpar \, \pi\lpar \psi \given \data_\obs\rpar \,\diff\psi\rbar,
\qquad B \in \cB.
\end{equation}
\end{definition}

\xhdr{Consistent Estimation of CEPOs} 
The CEPO-PPD captures the epistemic uncertainty about the CEPO encoded in the posterior. A concentrated distribution $\pi^{\mu_t}$ indicates that the observations $\data_\obs$ are informative and sufficiently large to accurately pin down the true CEPO, whereas a high-variance distribution implies that the data is insufficient for estimation. With that in mind, we now study under which conditions increasing the size of the observations $\data_\obs$ allows us to accurately recover the true CEPO from the CEPO-PPD. This is given through the following informal result (re-stated and proven formally in \cref{appx:theory}) which provides necessary and sufficient conditions on the prior $\pi$ under which the CEPO-PPDs enable consistent estimation of the CEPOs:
\begin{proposition}[Informal]\label{prop:cepo-ppd-consistency}
Under mild regularity assumptions (\cref{assump:reg-measurability,assump:reg-integrability} in \cref{appx:theory}), for almost all $\psi^\star \sim \pi$ and any set of i.i.d.\ samples
$\data_\obs\sim\dist^{\psi^\star}_\obs$, we have that as $|\data_{\obs}|\to\infty$,
\begin{equation}
  \E_{\mu \sim \pi^{\mu_t} \lpar \cdot \given \vect x,\data_\obs\rpar} [\mu]
  \overset{a.s.}{\longrightarrow}
  \mu_t(\vect{x}\midsem\psi^\star), \quad \forall t \in \cT,
  \text{ and almost all } \vect{x} \in \cX,
\end{equation}
\textbf{if and only if} the prior $\pi$ is \emph{CEPO-identifiable}, that is for almost all $\psi \sim \pi$, the CEPOs $\mu_t\lpar \cdot \midsem \psi \rpar$ only depend on the observational distribution $\dist_\obs^\psi$ (\cref{def:identification}). 
\end{proposition}

\textit{(Proof sketch)} We group all DGPs $\psi$ that share the same observational distribution $\dist_\obs^\psi$ into an equivalence class and induce a prior obtained from $\pi$ on the resulting quotient space. By Doob’s theorem \cite{doob1949application}--a classical result from Bayesian consistency theory--the posterior on this new prior almost surely concentrates on the true equivalence class once asymptotically many observations are given. Consequently, for any functional of the observations that is constant within each equivalence class, its posterior predictive converges almost surely to its true value. Importantly, the causal functional of interest, $\mu_t$, can be written as a functional of the observations if and only if the corresponding DGP has identifiable CEPOs. Thus, identifiability is both necessary and sufficient to ensure that $\mu_t$ is constant throughout the equivalence class, and for the consistency result to hold.

\textit{(Remark 1)}
While the algorithms in our paper use \emph{strong ignorability}, \cref{prop:cepo-ppd-consistency} itself is an entirely general result and can be extended to DGPs that are not necessarily ignorable, but whose CEPOs satisfy identifiability in \cref{def:identification}. Importantly for our practical setting, when the prior~$\pi$ enforces strong ignorability, \cref{prop:cepo-ppd-consistency} suggests that the CEPO-PPDs consistently recover the true CEPO.

\textit{(Remark 2)}
\cref{prop:cepo-ppd-consistency} highlights two key design principles for the prior $\pi$: $(i)$ $\pi$ must rule out non-identifiable cases, and, once identifiability is secured, $(ii)$ broadening $\pi$ increases the chance that a particular $\psi^\star$ lies within its support, thus enabling consistent recovery of the true CEPO for that $\psi^\star$. 

\xhdr{Learning the CEPO-PPD}
Having shown that CEPO-PPDs are useful for estimating the true CEPOs, we now describe how to learn them.  Inspired by PFNs, we train a single transformer $q_\theta$ to approximate the full predictive distribution $\pi^{\mu_t}$. To fit this model, we introduce the following loss:

\begin{definition}[Causal Data-Prior Loss]\label{def:causal-data-prior-loss}
For any $t \in \cT$, we define the {causal data-prior loss} as
\begin{align} \label{eq:causal-data-prior-loss}
\cL_{t}(\theta)
\;\coloneqq\;
\E_{\psi \sim \pi,\ \data_\obs \cup \{\vect{x}\} \ \sim\ \dist^\psi_\obs} \lbar - \log {q_\theta\lpar \mu_t(\vect{x}\midsem \psi) \given \vect{x}, t, \data_\obs\rpar} \rbar.
\end{align}
\end{definition}

In \cref{appx:theory2}, we show that minimizing $\cL_t(\theta)$ also minimizes the KL-divergence between the true CEPO-PPD and $q_\theta$,  leading to 
$q_\theta \lpar \cdot \given \vect{x},t,\data_\obs\rpar \approx \pi^{\mu_t} \lpar \cdot \given \vect{x},\data_\obs\rpar$ for all $t\in\cT$. This entire training process shifts the computational burden from inference to pre-training: rather than evaluating the posterior $\pi\lpar \psi \given \data_\obs\rpar$ at test time, the model learns to map observational data directly to the corresponding predictive distribution. When the model is well-fitted, the prior satisfies the assumptions of \cref{prop:cepo-ppd-consistency}, and $\data_\obs$ is sufficiently large, the predicted $q_\theta$ accurately pins down the true CEPO.

\begin{figure}\captionsetup{font=footnotesize, labelformat=simple, labelsep=colon}
    \centering
    \includegraphics[width=1\linewidth]{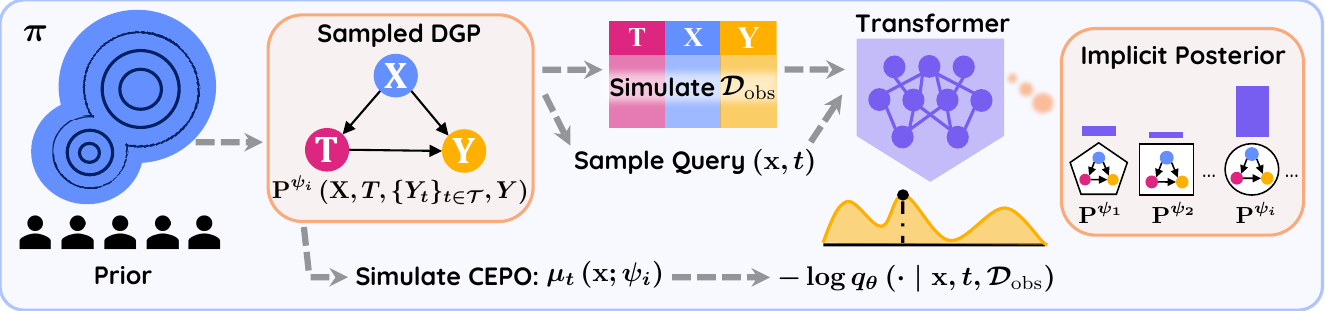}
    \caption{
    \textbf{Causal Data-Prior Training}. 
    At each iteration an index $\psi_i \sim \pi$ is sampled \textit{(left)}, yielding the DGP $\dist^{\psi_i}(\vect X,T,\{Y_t\}_{t\in\cT},Y)$. 
    From this DGP we simulate an observational context $\data_\obs$ and a query $(\vect{x},t)$ with its true $\mu_t(\vect{x}\midsem\psi_i)$ \textit{(center)}. 
    Passing $(\vect{x},t,\data_\obs)$ through the transformer predicts the CEPO‑PPD $q_\theta\lpar \cdot\given \vect{x},t,\data_\obs\rpar$ \textit{(in yellow)}, which is derived from an implicit posterior $\pi\lpar \cdot \given \data_\obs\rpar$ that is \emph{never} explicitly computed \textit{(right)}. 
    We train $\theta$ to minimize the causal data‑prior loss \textit{(bottom)}.
    }
    \label{fig:training_figure}
    \vspace{-0.8cm}
\end{figure}

\cref{fig:training_figure} visually illustrates optimizing the causal data-prior loss using stochastic gradient descent: at each iteration, we sample a DGP $\psi_i \sim \pi$, generate an observational dataset $\data_\obs$ from this DGP, and select a query point $(\vect{x}, t)$. We compute (simulate) the ground-truth CEPO $\mu_t\lpar \vect{x} \midsem \psi_i\rpar$ and feed both the observational data and query to the model. The model outputs a CEPO-PPD, and we update $\theta$ using gradient descent to increase the probability assigned to the true CEPO value. Through training, $\theta$ minimizes the data-prior loss and implicitly learns to perform posterior predictive inference, and estimate the predictive distribution $\pi^{\mu_t}$, without ever explicitly computing the posterior.

\xhdr{Point \& Distributional Estimation of Causal Effects} 
Given observational data $\data_\obs$ from an underlying $\psi^\star$, a natural point estimate for CEPOs is the mean of the predicted CEPO-PPD,
$\E_{\mu \sim q_{\theta}\lpar \cdot \given \vect{x}, t, \data_\obs  \rpar} \lbar  \mu  \rbar \approx \mu_t(\vect{x} \midsem \psi^\star)$. 
These CEPO estimates can also form point estimates for CATEs using \cref{eq:cate}, and for ATEs using \cref{eq:ate} by empirical averaging across units in $\data_\obs$. 

Beyond point estimation, the estimated CEPO-PPDs can also capture the epistemic uncertainty about the causal effects. We can use $q_\theta$ to construct credible intervals around CEPOs, CATEs, and ATEs via sampling from $q_\theta\lpar \cdot \given \vect{x}, t=1, \data_\obs\rpar$ and $q_\theta\lpar \cdot \given \vect{x}, t=0, \data_\obs\rpar$. We can then use these intervals to quantify the uncertainty of our estimated causal effects.

\section{Implementing \method} \label{sec:implementation}

\begin{wrapfigure}{r}{0.378\linewidth}\captionsetup{font=footnotesize, labelformat=simple, labelsep=colon}
  \vspace{-18pt}
  \centering
  \includegraphics[width=\linewidth]{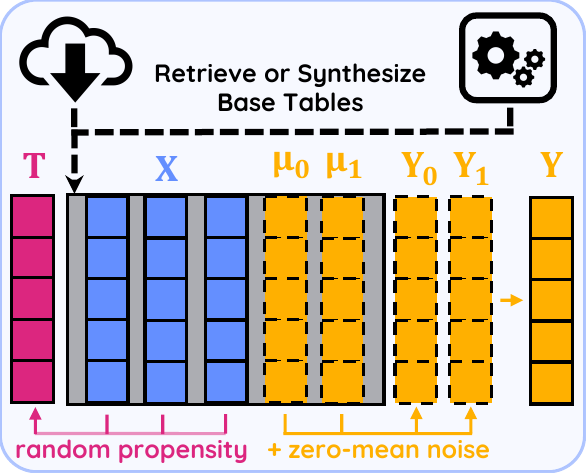}
    {\captionsetup{font=footnotesize, aboveskip=2pt, belowskip=-20pt}
    \captionof{figure}{\textbf{Prior construction}. Sample diverse base tables (OpenML or synthetic TabPFN), select covariates $X$, draw treatment $T$ with a random propensity model, select columns $\mu_0, \mu_1$ and add zero‑mean noise to form $Y_0,Y_1,$ and $Y$.}\label{fig:prior-sampling}
    }
\end{wrapfigure}
While \cref{sec:method} presents the framework in general form (arbitrary finite $\cT$ and identifiability), for implementation we focus on binary treatments $\cT = \{0, 1\}$ under strong ignorability. These assumptions reflect the most common settings encountered by practitioners and serve as a natural starting point. Extending the implementation and algorithms to more general settings is left for future work.

\xhdr{A Scalable Prior} 
Here, we focus on designing an appropriate prior $\pi$ over DGPs that satisfies the theoretical requirements established in \cref{prop:cepo-ppd-consistency}. This prior must balance two factors: First, it should contain a rich set of DGPs with sufficient coverage to approximate real-world scenarios—similar to the priors used in successful tabular predictive models like TabPFN \citep{hollmann2023tabpfn, hollmann2025accurate}, TabDPT \citep{ma2024tabdpt}, and TabICL \citep{qu2025tabicl}. Second, and uniquely for causal inference, all DGPs in our prior must satisfy strong ignorability which directly implies identifiability of the prior. Moreover, the generated DGPs must allow us to access the ground-truth CEPOs, as required by the causal data-prior loss in \cref{def:causal-data-prior-loss} for training.

To address these requirements, we develop a procedure that can transform \emph{any} base table from standard tabular priors into a valid causal dataset, illustrated by \cref{fig:prior-sampling}:
\stepindicator{i} retrieve a base table with $N$ rows from either a large library of tabular data\footnote{We use 337 OpenML tables \citep{bischl2021cc18}, checked to avoid leakage, totaling over $10^9$ feature values.} or synthesize it (details in \cref{appx:prior_generation});
\stepindicator{ii} randomly select columns with a varying number of covariates as $\vect{X}$;
\stepindicator{iii} pick two other columns, relabel them as $\mu_0(\vect{X}), \mu_1(\vect{X})$;
\stepindicator{iv} optionally add zero-mean noise to $\mu_0(\vect{X})$ and $\mu_1(\vect{X})$ to obtain $Y_0$ and $Y_1$, or simply set $Y_0 = \mu_0(\vect{X})$ and $Y_1 = \mu_1(\vect{X})$; these four steps simulate samples from the joint distribution $(\vect{X}, Y_0, Y_1)$;
\stepindicator{v} generate a random function $f$, leveraging similar synthetic functions as in \citet{hollmann2023tabpfn} to map covariates to their treatment logits;
\stepindicator{vi} sample binary treatments $T \sim \bernoulli \lpar \sigmoid\lpar f(\vect{X})\rpar \rpar$;
\stepindicator{vii} finally, form the observed outcomes $Y := Y_T$.

\looseness=-1 The procedure above ``simulates'' a collection $\{t^{(n)}, \vect{x}^{(n)}, \mu_0^{(n)}, \mu_1^{(n)}, y^{(n)}\}_{n=1}^N$ from an underlying DGP that can be used to sample the observational data and obtain CEPOs necessary for training (recall \cref{fig:training_figure}). This approach guarantees strong ignorability \emph{by design}: since treatment $T$ is determined solely from $\vect{X}$, it is conditionally independent from the potential outcomes $Y_0, Y_1$. Additionally, by applying the sigmoid function, we ensure $0 < \dist\lpar T=1 \given \vect{X}\rpar < 1$, satisfying positivity. While this procedure primarily targets binary treatments, it can naturally extend to finite discrete treatments.

For the diversity aspect of $\pi$, we rely on the empirical success of existing tabular foundation models and the deliberate design in our generation process. Sampling covariates directly from a mix of real and synthetic tables yields data that is more likely to reflect the scenarios the model will face at inference. We assume no distributional assumptions on covariates and potential outcomes. \cref{appx:prior_generation} details additional mechanisms for controlling treatment effect heterogeneity and positivity in our synthetic DGPs, as well as the detailed configurations of the prior-generation process.

\xhdr{Model Architecture \& Parallel Training}
We model $q_\theta$ using a PFN‑style transformer encoder that receives a sequence of row tokens as \emph{context} (i.e., $\data_\obs$), where each token embeds a triplet $(t^{(n)},\vect{x}^{(n)},y^{(n)})$. At every iteration, we embed $B_Q$ batched \emph{query} tokens $(t,\vect{x})$. We then apply 20 layers of self‑attention and MLP layers, followed by a final projection layer to get $q_\theta\lpar \cdot \given \vect{x}, t, \data_\obs \rpar$ for all the $(t,\vect{x})$ pairs in the batched query. 
The transformer uses the asymmetric masking: query tokens cannot attend to each other and attend only to context tokens, so predictions are conditionally independent given the context.

To model each CEPO‑PPD, we approximate it with a quantized histogram. We discretize the outcome axis into $L=1024$ bins and let the network project the query tokens into $L$ logits. We then apply SoftMax to turn the logits into a quantized distribution $q_\theta\lpar \cdot\given\vect{x},t,\data_\obs\rpar[\ell], \forall \ell \in [L]$. At each round of gradient update, we place a Gaussian with a small $\sigma$ at the true CEPO ${\mu}_t(\vect{x})$ and integrate it over bins to obtain Gaussian quantized probabilities ${\cN({\mu}_t(\vect{x}), \sigma^2)}[\ell]$  and minimize the \emph{histogram loss}:
\begin{equation}
    \texttt{HL}\lbar {\mu}_t(\vect{x}) \,\Vert\, q_\theta\rbar
=-\sum_{\ell=1}^{L}
{\cN({\mu}_t(\vect{x}), \sigma^2)}[\ell]\cdot
\log q_\theta[\ell].
\label{eq:hl-loss}
\end{equation}
This loss is an approximation to the causal data-prior loss in \cref{eq:causal-data-prior-loss}; it coincides in the limit $\sigma \to 0$ and $L \to \infty$. The histogram loss formulation affords a tractable proxy for the continuous CEPO‑PPD. 

\begin{figure}
    \centering
    \captionsetup{font=footnotesize}
    \includegraphics[width=0.9\linewidth]{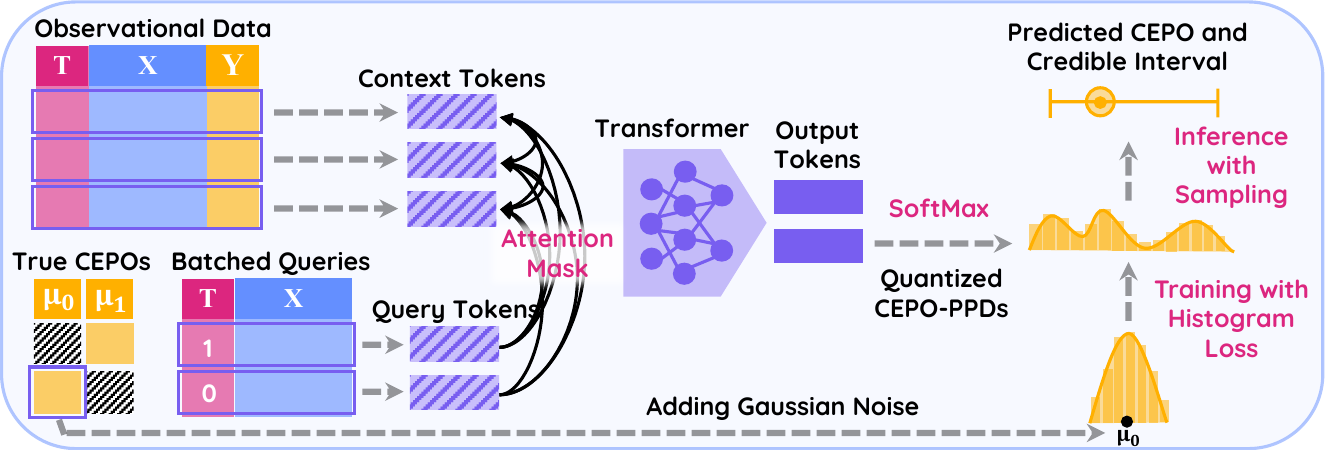}
    \caption{{\bf Architecture, Training, and Inference Details.} \emph{(Left)}: An observational dataset, and a batch of queries along with their true CEPO values are sampled from the prior. Each observational row forms a context token, while query tokens consist of only the treatment and covariates. \emph{(Middle)}: The context and query tokens are fed into a transformer encoder with an asymmetric attention masking, where both context and query tokens attend only to the context tokens. \emph{(Bottom-Right)}: The output tokens are projected into a 1024-dimensional logit vector and softmaxed to form a discretized CEPO-PPD. Then, the true CEPO value corresponding to each output token is smoothed by adding narrow-width Gaussian, and training is done by minimizing the cross-entropy (histogram) loss.  \emph{(Top-Right)}: At inference time, the CEPO-PPD mean is used as the point estimate.}
    \label{fig:arch-details}
    \vspace{-0.3cm}
\end{figure}

A more detailed overview of the architecture and procedures for point and interval estimates is illustrated in \cref{fig:arch-details}; further details (e.g., parameter counts, compute, inference-time techniques, number of prior datasets, scalability, and speed) are available in Appendices  \ref{appx:arch}, \ref{appx:sensitivity}, and \ref{appx:speed}.

\section{Experiments}
\label{sec:experiments}

\xhdr{Baseline Causal Effect Estimators}
We compare to a broad suite of baselines. This includes double machine learning (DML)~\citep{chernozhukov2017doubledebiased, athey2019generalized,foster2023orthogonal}, doubly robust learner (DR-Learner)~\citep{kunzel2019metalearners, kennedy2020optimal}, as well as the T-, S-, X-, and domain adaptation learner (DA-Learner), all part of the \texttt{EconML} package~\citep{econml}. Moreover, we include deep-learning–based methods such as TarNet~\citep{shalit2017estimating}, DragonNet~\citep{shi2019adapting}, and RA-Net~\citep{curth2021inductive}, implemented via the \texttt{CATENets} library~\citep{curth2021catenets}. Finally, we compare to inverse propensity weighting (IPW)~\citep{rosenbaum1983central}, Bayesian regression trees (BART)~\citep{hill2011bayesian,chan2022data}, and generalized random forests (GRF)~\citep{athey2019generalized}. All the baselines, except for IPW, provide both CATE and ATE estimates.

\emph{Importantly, we tune most of the baselines with cross-validation via grid search. The set of hyperparameter, along with the results with default hyperparameters are all detailed in \cref{appx:hyper-setup}}.

\xhdr{Benchmarks with Ground-Truth Effects \href{https://github.com/vdblm/CausalPFN/blob/main/notebooks/causal_effect_full.ipynb}{
    \includegraphics[height=1.4em]{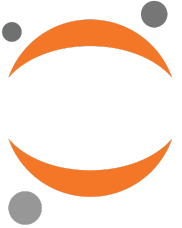}
}}
A handful of benchmarks provide ground-truth causal effects, allowing us to directly measure estimation errors. 
Given a dataset of $N$ units with covariates and ground-truth CATE values
$\{(\mathbf{x}^{(n)},\,\tau(\mathbf{x}^{(n)}))\}_{n=1}^N$, and a
ground-truth ATE $\lambda$, we evaluate models using the relative ATE error and
the precision in estimation of heterogeneous effects (PEHE) \citep{hill2011bayesian}:
\begin{footnotesize}
\begin{equation}
\mathrm{Relative Error}(\hat{\lambda}) = \frac{|\hat{\lambda}-\lambda|}{|\lambda|},
\qquad
\mathrm{PEHE}(\hat{\tau}) = \sqrt{\frac{1}{N}\sum_{n=1}^N
\bigl(\tau(\mathbf{x}^{(n)})-\hat{\tau}(\mathbf{x}^{(n)})\bigr)^2 }.
\end{equation}
\vspace{-1mm}
\end{footnotesize}\\
Here, $\hat{\tau}$ and $\hat{\lambda}$ denote the estimated CATE and ATE, respectively.
\cref{tab:causal_effect_results} compares \method\ to all baselines on four standard set of datasets: 100 realizations of IHDP~\citep{ramey1992infant, hill2011bayesian}, 10 realizations of ACIC 2016~\citep{dorie2019automated}, and the Lalonde {\tiny CPS} and Lalonde {\tiny PSID} cohorts~\citep{lalonde1986evaluating} with their causal effects provided by \texttt{RealCause} (each with 100 realizations)~\citep{neal2020realcause}. Our model demonstrates superior performance on both CATE and ATE tasks, remaining within the top models across most benchmarks. To assess the overall performance of each method for CATE estimation, we calculate the average rank of each method across all 310 realizations based on PEHE. For ATEs, we calculate the average rank of each method based on relative errors. \method\ outperforms all baselines in terms of average CATE rank, while being competitive for average ATE rank. Notably, our model is trained entirely on simulated data and \emph{never} sees the evaluation data during pre-training. While some baseline estimators in \cref{tab:causal_effect_results} perform well on specific datasets, they underperform on others. In contrast, the consistent performance of \method\ suggests that amortized approaches can potentially eliminate the manual burden of task-specific estimator design.

\begin{table}[t]
\centering
\captionsetup{font=footnotesize}
\caption{\textbf{CATE \& ATE results.} Columns correspond to benchmark suites: IHDP, ACIC~2016, Lalonde {\tiny CPS/PSID}. \textit{(left half)} mean PEHE and the average rank when pooling all tasks. \textit{(right half)} mean ATE relative error and its average across all tasks. Lalonde PEHE is in thousands. The \firstbest{best} and \secondbest{second best} columns are highlighted. Cells with ``—'' indicate that the method is not applicable.} 
\label{tab:causal_effect_results}
\begin{adjustbox}{max width=\textwidth}
\LARGE
\begin{tabular}{lccccc|ccccc}
\toprule
\multirow{2}{*}{\textbf{Method}} &
\multicolumn{5}{c|}{\textbf{Mean PEHE $\pm$ Standard Error} $(\downarrow \text{better})$} &
\multicolumn{5}{c}{\textbf{Mean ATE Relative Error $\pm$ Standard Error} $(\downarrow \text{better})$} \\
\cmidrule(r){2-6}\cmidrule(l){7-11}
& IHDP & ACIC 2016 & \multicolumn{1}{c}{Lalonde {\footnotesize CPS}} & \multicolumn{1}{c}{Lalonde {\footnotesize PSID}} & \multicolumn{1}{c|}{Avg.} & 
IHDP & ACIC 2016 & Lalonde {\footnotesize CPS} & Lalonde {\footnotesize PSID} & \multicolumn{1}{c}{Avg.} \\
& & & ($\times10^{3}$) & ($\times10^{3}$) & Rank & & & & & Rank\\
\midrule
\textbf{CausalPFN} &
\firstbest{0.58$\pm$0.07}&
{0.92$\pm$0.11}&
\firstbest{8.96$\pm$0.02}&
\firstbest{14.40$\pm$0.20}&
\firstbest{2.30$\pm$0.10}& 
0.20$\pm$0.04& 
{0.05$\pm$0.01}& 
\firstbest{0.13$\pm$0.01}& 
{0.22$\pm$0.02}& 
{4.45$\pm$0.19}\\

T-Learner&
\secondbest{1.73$\pm$0.30}&
{0.76$\pm$0.07}& 
\secondbest{9.22$\pm$0.04}& 
{15.16$\pm$0.46}& 
\secondbest{3.57$\pm$0.16}&
0.21$\pm$0.04& 
{0.03$\pm$0.01}& 
{0.24$\pm$0.02}& 
\secondbest{0.16$\pm$0.03}& 
\firstbest{4.31$\pm$0.18}\\

DA-Learner&
{2.07$\pm$0.36}&
{0.72$\pm$0.08}& 
{9.39$\pm$0.06}& 
\secondbest{14.55$\pm$0.24}& 
{3.60$\pm$0.16}&
0.23$\pm$0.04& 
{0.03$\pm$0.01}& 
{0.27$\pm$0.02}& 
{0.20$\pm$0.03}& 
{4.83$\pm$0.19}\\

DragonNet &
2.16$\pm$0.25& 
2.11$\pm$0.19& 
10.93$\pm$0.15& 
16.45$\pm$0.29& 
5.99$\pm$0.18& 
0.20$\pm$0.04& 
0.06$\pm$0.02& 
0.55$\pm$0.03& 
0.47$\pm$0.03& 
6.26$\pm$0.17 \\

IPW &
—& 
—&
—&
—&
—& 
0.24$\pm$0.04 & 
0.21$\pm$0.05 & 
\secondbest{0.17$\pm$0.01}& 
\firstbest{0.10$\pm$0.01}& 
\secondbest{4.41$\pm$0.21}\\

RA-Net &
2.35$\pm$0.19 &
2.35$\pm$0.25 & 
11.74$\pm$0.09 & 
18.33$\pm$0.43 & 
7.15$\pm$0.16& 
0.20$\pm$0.04 & 
0.07$\pm$0.03 & 
0.74$\pm$0.02 & 
0.50$\pm$0.04 & 
6.78$\pm$0.17 \\

X-Learner&
3.31$\pm$0.51& 
\firstbest{0.60$\pm$0.08}& 
12.15$\pm$0.15&
20.28$\pm$0.49& 
7.46$\pm$0.19&
\secondbest{0.16$\pm$0.04}& 
0.03$\pm$0.01&
0.84$\pm$0.03&
0.72$\pm$0.03& 
7.31$\pm$0.19\\

TarNet &
1.82$\pm$0.14& 
2.20$\pm$0.21& 
12.88$\pm$0.02& 
19.19$\pm$0.18& 
8.38$\pm$0.14& 
0.20$\pm$0.04& 
0.05$\pm$0.02& 
1.00$\pm$0.00& 
0.78$\pm$0.01& 
8.83$\pm$0.15\\

S-Learner &
2.57$\pm$0.41 &
0.85$\pm$0.13 & 
12.66$\pm$0.05 & 
21.80$\pm$0.18 & 
8.43$\pm$0.18&
0.20$\pm$0.04 & 
0.03$\pm$0.01 & 
0.97$\pm$0.01 & 
0.90$\pm$0.02 & 
8.85$\pm$0.18 \\

BART &
2.50$\pm$0.39& 
\secondbest{0.68$\pm$0.11}& 
12.81$\pm$0.05& 
21.36$\pm$0.16& 
8.55$\pm$0.16& 
0.44$\pm$0.09& 
0.04$\pm$0.01& 
0.99$\pm$0.01& 
0.86$\pm$0.01& 
8.99$\pm$0.18 \\

GRF &
3.67$\pm$0.61& 
1.32$\pm$0.30& 
12.33$\pm$0.06& 
22.91$\pm$0.17& 
8.82$\pm$0.18& 
0.18$\pm$0.03& 
0.07$\pm$0.02& 
0.82$\pm$0.02& 
0.85$\pm$0.02& 
8.02$\pm$0.18\\

Forest DML&
4.53$\pm$0.73& 
1.48$\pm$0.31& 
12.95$\pm$0.04& 
22.99$\pm$0.15& 
9.83$\pm$0.17& 
\firstbest{0.08$\pm$0.01}&
0.05$\pm$0.01& 
1.03$\pm$0.01& 
1.05$\pm$0.01& 
9.60$\pm$0.21\\

Forest DR Learner&
4.02$\pm$0.67&
1.34$\pm$0.29& 
15.98$\pm$0.68& 
22.78$\pm$0.54& 
10.00$\pm$0.17& 
0.17$\pm$0.03&
0.04$\pm$0.02&
1.20$\pm$0.23& 
3.64$\pm$2.78& 
8.38$\pm$0.18\\

\bottomrule
\end{tabular}
\end{adjustbox}
\end{table}

\xhdr{Policy Evaluation on Marketing Randomized Trials \href{https://github.com/vdblm/CausalPFN/blob/main/notebooks/qini.ipynb}{
    \includegraphics[height=1.4em]{assets/jupyter_logo.pdf}
}} 
Ground-truth CATEs are only available for synthetic or semi-synthetic datasets. However, if a randomized controlled trial (RCT) is available, we can still evaluate the quality of a CATE estimator by assessing the performance of policies derived from it.
A common tool for evaluating such policies is the \emph{Qini curve}~\citep{radcliffe2007using}, which plots the cumulative treatment effect when units are ranked in descending order of their predicted CATE. 

Formally, let $(y^{(n)},t^{(n)})_{n=1}^N$ denote outcomes and binary treatments from an RCT, and let $\widehat{\tau}_n$ be the corresponding CATE estimates, ordered so that $\widehat{\tau}_{1}\ge\cdots\ge\widehat{\tau}_{N}$.  Define  
\begin{equation}
\lambda(q) \coloneqq \sum_{n=1}^{\lfloor qN\rfloor}
\lpar \tfrac{t^{(n)}y^{(n)}}{r(q)}-\tfrac{(1-t^{(n)})y^{(n)}}{1 - r(q)}\rpar, \qquad
Q(q) \coloneqq q \cdot \lambda(q) / \lambda(1),
\qquad
0\le q\le 1,
\end{equation}
where $r(q)=\tfrac{1}{\lfloor qN \rfloor}\sum_{n}^{\lfloor qN \rfloor} t^{(n)}$ is the empirical treatment rate for the first $q$-quantile of units.
Because the data comes from an RCT, $\lambda(q)$ unbiasedly estimates the ATE for the top $q$‑quantile of units ranked by predicted CATEs.  
Plotting $Q(q)$ against the treated fraction $q$ yields the (normalized) Qini curve, and the area under this curve is called the \emph{Qini score}.  
A random ranking produces a baseline curve as a straight line from $(0,0)$ to $(1,1)$.  
The higher the Qini curve lies above this line, the better the model prioritizes high‑impact units with larger CATE values, leading to greater lift and policy benefit. 

\begin{figure}[t]
  \centering
  \begin{minipage}[t]{0.38\linewidth}
    \vspace{0pt}
    \centering
    \includegraphics[width=\linewidth]{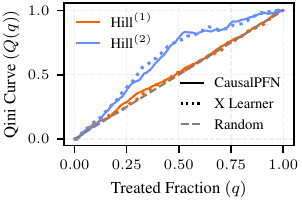}
    \captionof{figure}{\footnotesize Hill$^{(1)}$ \& Hill$^{(2)}$ Qini curves.}
    \label{fig:hillstrom}
  \end{minipage}\hfill
  \begin{minipage}[t]{0.6\linewidth}
    \vspace{0pt}%
    \centering
    \captionof{table}{\footnotesize \textbf{Normalized Qini scores} ($\uparrow$ better).  All datasets use 50k stratified subsamples, except Hill$^{(1)}$ and Hill$^{(2)}$, which use the full 64k rows.  Columns are normalized to 1.0 for \firstbest{the best model}.}
    \label{tab:uplift}
    \footnotesize
    \setlength{\tabcolsep}{4pt}
    \resizebox{\linewidth}{!}{%
      \begin{tabular}{lccccccc}
        \toprule
        \textbf{Method} &
        \textbf{Hill$^{(1)}$} &
        \textbf{Hill$^{(2)}$} &
        \textbf{Criteo} &
        \textbf{X5} &
        \textbf{Lenta} &
        \textbf{Mega} &
        \textbf{Avg.} \\
        \midrule
        \textbf{\method}  & 0.992 & 0.968 & 0.859 & 0.922 & \firstbest{1.000} & 0.970 & \firstbest{0.952}\\
        X~Learner         & 0.975 & 0.980 & \firstbest{1.000} & 0.937 & 0.771 & \firstbest{1.000} & 0.944\\
        S~Learner         & \firstbest{1.000} & \firstbest{1.000} & 0.881 & \firstbest{1.000} & 0.651 & 0.941 & 0.912\\
        DA~Learner        & 0.985 & 0.964 & 0.626 & 0.929 & 0.781 & 0.998 & 0.881\\
        T~Learner         & 0.991 & 0.972 & 0.701 & 0.964 & 0.644 & 0.986 & 0.876\\
        \bottomrule
      \end{tabular}}
  \end{minipage}
  \vspace{-0.5cm}
\end{figure}

We benchmark \method\ on five large marketing RCTs from the \texttt{scikit‑uplift} library~\citep{user-guide-for-uplift-modeling}.
The first dataset, Hillstrom~\citep{hillstrom2008mining}, includes 64,000 customers randomly assigned to one of three treatments: no e‑mail, an e‑mail advertising men’s merchandise, or an e‑mail advertising women’s merchandise. The outcome is whether a website visit occurred within two weeks (binary).
We consider two causal tasks:
\textbf{Hill$^{(1)}$} – Men’s‑merchandise e‑mail (treatment) vs. no e‑mail (control), and
\textbf{Hill$^{(2)}$} – Women’s‑merchandise e‑mail vs. no e‑mail.
We estimate CATEs using \method~(five‑fold honest splitting) and X Learner. \cref{fig:hillstrom} shows Qini curves where \method\ closely matches X Learner across the targeting range. Notably, Hill$^{(2)}$ shows much greater gains, \emph{suggesting focusing on women’s‑merchandise ad campaigns, compared to men's, can drive more gains in the number of website visits}.
We also evaluate \method\ on four larger campaigns—\textbf{Lenta}, Retail Hero (\textbf{X5}), Megafon (\textbf{Mega}), and \textbf{Criteo}~\citep{lenta2020,retailhero2020,megafon2020,diemert2018criteo}—each with $\sim$$10^{6}$ rows. For tractability, we compute Qini scores on stratified $50$k subsamples; \cref{tab:uplift} shows \method\ achieves the best mean performance. 
However, when we run it on full tables (see \cref{tab:uplift-full} of \cref{appx:qini}), we observe a drop in performance, which aligns with known context-length limitations of PFN-style transformers on large tables~\citep{thomas2024retrieval}. Still, the strong subsample results highlight the potential of scaling \method\ to longer contexts, which remains an important future direction.

\xhdr{Uncertainty \& Calibration \href{https://github.com/vdblm/CausalPFN/blob/main/notebooks/calibration.ipynb}{
    \includegraphics[height=1.4em]{assets/jupyter_logo.pdf}
}} 
Recall from \cref{sec:method} that for each unit covariate $\vect{x}$, \method\ can produce both point estimates and credible intervals for the CATE and CEPOs. We do so by drawing 10,000 samples from the quantized distributions $q_\theta \lpar \cdot \given \vect{x}, t, \data_\obs\rpar$ and construct credible intervals at any desired significance level $\alpha$. Here, we evaluate these intervals, focusing on the model's calibration. We also assess a key assumption from \cref{prop:cepo-ppd-consistency}—whether the inference-time DGP $\psi^\star$ lies within the support of the prior $\pi$, and how the model behaves when this assumption is violated. 

We define families of synthetic DGPs to simulate both in-distribution and out-of-distribution (OOD) scenarios.  
Each DGP samples covariates $\vect x$ from a uniform distribution, defines a treatment logit function $f$ and CEPO functions $\mu_t$ for $t \in \{0, 1\}$, assigns treatment via 
$T \sim \bernoulli \lpar \sigmoid \lpar f(\vect{x})\rpar\rpar$,  
and generates potential outcomes as  
$y_t = \mu_t(\vect{x}) + \epsilon_t$, where $\epsilon_t$ is drawn from a standard Uniform, Gaussian, or Laplace.  
We consider two DGP families;
\textbf{Sinusoidal}, where $f$ and $\mu_t$ are functions with sinusoidal components, and  
\textbf{Polynomial}, where the functions $f$ and $\mu_t$ are polynomials of varying degree (see \cref{appx:uncertainty} for detailed configurations).
\method~is trained either on the same family it is tested on, or on a different one (OOD).

\begin{figure}
    \centering
    \includegraphics[width=0.9\linewidth]{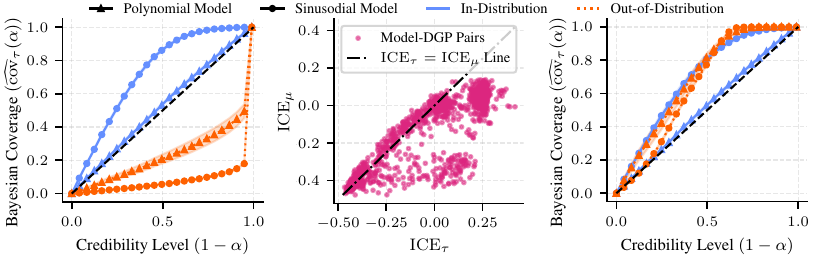}
    \captionsetup{font=footnotesize}
    \caption{ \looseness=-1 \textbf{Calibration.} \textit{(Left)}: CATE coverage vs. nominal credibility. In-distribution DGPs (blue) lie on or above the diagonal (calibrated/conservative), while OOD DGPs (orange) fall below it (overconfident).
    \textit{(Middle)}: Across model–DGP pairs, CATE ICE (x-axis) exceeds regression ICE (y-axis).
    \textit{(Right)}: Temperature scaling based on regression ICE ensures the model is either calibrated or conservative for both in- and out-of-distribution DGPs.}
    \label{fig:calibration}
    \vspace{-15pt}
\end{figure}

For a unit with covariates $\vect{x}$ and significance level $\alpha$, we say the true CATE is \emph{covered} if $\tau(\vect{x})$ lies within the predicted $100(1-\alpha)\%$ interval obtained using samples from $q_\theta$.  
Plotting Bayesian coverage against nominal levels of $\alpha$ yields the CATE calibration curve. 
As shown in \cref{fig:calibration} (left), \method\ is reliably calibrated under in-distribution settings but becomes severely overconfident when evaluated on OOD DGPs ($\psi^\star \not\sim \pi$).  
This aligns with prior observations that neural models often exhibit pathological overconfidence under distribution shift~\citep{guo2017calibration, ovadia2019can}.

To correct this, we apply a temperature parameter $\theta_T$ to the SoftMax that outputs the quantized CEPO‑PPD from the logits of the model.  
We aim to tune $\theta_T$ to minimize the calibration error. However, direct CATE calibration is impossible because $\tau(\vect{x})$ is never observed at test-time.
Instead, we introduce the \emph{regression calibration} based on observational data: an observed triple $(t,\vect{x},y)$ is covered by the predicted credible interval when $y$ lies inside the model’s predicted interval for the CEPO-PPD $\mu_t\lpar \vect x \midsem \psi^\star \rpar$.
With that in mind, we let $\widehat{\mathrm{cov}}_\mu(\alpha)$ and $\widehat{\mathrm{cov}}_\tau(\alpha)$ denote the Bayesian coverage at level $\alpha$ for the regression and CATE calibration curves, respectively, and define  
\begin{equation}
\mathrm{ICE}_{\mu} \coloneqq \int_{0}^{1} \lpar \widehat{\mathrm{cov}}_\mu(\alpha) - \alpha\rpar\,\diff\alpha, \text{ and } \qquad \mathrm{ICE}_{\tau} \coloneqq \int_{0}^{1} \lpar \widehat{\mathrm{cov}}_\tau(\alpha) - \alpha\rpar\,\diff\alpha,
\end{equation}
as the integrated coverage error (ICE) for regression and CATE (negative values $=$ overconfidence). 

Note that we do not expect $\widehat{\mathrm{cov}}_\mu$ to be calibrated: regression intervals combine epistemic uncertainty of the CEPO with the irreducible (aleatoric) noise in $Y$, so $\mathrm{ICE}_{\mu}$ is biased. Still, it holds a useful signal. Across all model--DGP pairs in \cref{fig:calibration} (middle), we consistently observe $\mathrm{ICE}_{\mu} \le \mathrm{ICE}_{\tau}$: the regression curve sits at or below the CATE curve. While $\mathrm{ICE}_{\tau}$ is inaccessible without having the true CATE, $\mathrm{ICE}_{\mu}$ is computable from observational data. Because we empirically observe $\mathrm{ICE}_\mu \le \mathrm{ICE}_\tau$ across model–DGP pairs, temperature scaling that brings $\mathrm{ICE}_\mu$ to zero guarantees that CATE intervals are calibrated or conservative. We fit the temperature by 5-fold calibration on the observational data (no access to ground-truth counterfactuals).
The calibrated curves in \cref{fig:calibration} (right) confirm that, after temperature scaling, \method's overconfidence on the OOD test-sets disappears. Additional synthetic and real‑world data experiments appear in \cref{appx:uncertainty}.

\xhdr{Comparison to TabPFN} We also compare against the latest version of TabPFN \citep{hollmann2025accurate}, plugging its regression output as a proxy for CEPO. As \cref{tab:tabpfn_comparison} shows, TabPFN is surprisingly competitive without any causal tuning, yet \method\ outperforms it on every benchmark except ACIC 2016. To isolate the benefit of training on a causal prior, compared to the predictive \emph{non-identifiable} prior in TabPFN, we fine-tune it on our prior for 48 hours on an H100 GPU. This causal fine-tuning boosts the performance and confirms the added value of identifiable priors for causal effect estimation.

\begin{table}[t]
\centering
\captionsetup{font=footnotesize}
\caption{\textbf{TabPFN Comparison.} PEHE \textit{(left half)} alongside ATE relative error \textit{(right half)}. TabPFN$^\star$ is the latest TabPFN model \citep{hollmann2025accurate} tuned with our prior. Best numbers are \firstbest{highlighted}.}
\label{tab:tabpfn_comparison}
\begin{adjustbox}{max width=\textwidth}
\Large
\begin{tabular}{lcccc|cccc}
\toprule
\multirow{2}{*}{\textbf{Method}} &
\multicolumn{4}{c|}{\textbf{PEHE $\pm$ Standard Error} ($\downarrow$ better)} &
\multicolumn{4}{c}{\textbf{ATE Relative Error $\pm$ Standard Error} ($\downarrow$ better)} \\
\cmidrule(r){2-5}\cmidrule(l){6-9}
& IHDP & ACIC 2016 & \multicolumn{1}{c}{Lalonde {\footnotesize CPS}} & \multicolumn{1}{c|}{Lalonde {\footnotesize PSID}} & 
IHDP & ACIC 2016 & Lalonde {\footnotesize CPS} & Lalonde {\footnotesize PSID}\\
& & & ($\times10^{3}$) & ($\times10^{3}$) & & & & \\
\midrule

\textbf{\method~(Ours)} &
\firstbest{0.58$\pm$0.07} & 
0.92$\pm$0.11 & 
\firstbest{8.96$\pm$0.02} & 
\firstbest{14.40$\pm$0.20} &
\firstbest{0.20$\pm$0.04} & 
{0.05$\pm$0.01} & 
\firstbest{0.13$\pm$0.01} & 
\firstbest{0.22$\pm$0.02} \\

\textbf{TabPFN$^\star$ (Ours)} &
{0.90$\pm$0.16} &
\firstbest{0.47$\pm$0.05} & 
{8.97$\pm$0.06} & 
{14.90$\pm$0.95} &
{0.21$\pm$0.04} & 
\firstbest{0.03$\pm$0.01} & 
{0.17$\pm$0.02} & 
{0.22$\pm$0.08} \\

TabPFN &
0.95$\pm$0.20 & 
{0.54$\pm$0.08} & 
9.45$\pm$0.19 & 
18.7$\pm$0.83 &
{0.21$\pm$0.04} & 
\firstbest{0.03$\pm$0.01} & 
0.32$\pm$0.05 & 
0.60$\pm$0.07 \\

\bottomrule
\end{tabular}
\end{adjustbox}
\vspace{-10pt}
\end{table}

\section{Related Work}

\xhdr{Single-Dataset Estimators}
Common methods for causal effect estimation are trained and applied on a single dataset. Representative examples include the X‑, S‑, DR‑, and RA‑Learners, as well as IPW and DML~\citep{econml}.
Alongside these approaches, several neural variants such as TARNet~\citep{shalit2017estimating}, DragonNet~\citep{shi2019adapting}, CEVAE~\citep{louizos2017causal}, and NCMs~\citep{xia2021causal,xia2022neural} have been proposed; however, all of them still require per‑dataset training and do not amortize across various datasets.

\xhdr{Amortized Causal Inference}
Amortized methods train a \emph{single} network that maps observational data to causal quantities across \emph{multiple} DGPs. Existing approaches fall into two groups: (i) methods that first recover a causal graph from observational data and then compute interventions on that graph \citep{scetbon2024fixed, mahajan2024zero}, following ideas from causal discovery \citep{peters2014causal, zheng2018dags, khemakhem2021causal, lorch2022amortized, ke2022learning, kamkari2023order}; and (ii) methods that infer causal effects end-to-end \citep{nilforoshan2023zero, zhang2023towards, bynum2025black}. Amortization has also been explored in decision-making, where the aim is to learn policies that generalize across environments or tasks \citep{lee2023supervised, lau2024personalized}. While closely related, none of these methods provides a ready-to-use estimator that consistently surpasses specialized single-dataset estimators on standard benchmarks. In contrast, our method is trained once and produces causal effects without any access to or adaptation on the test-time DGPs. Through large-scale training, \method\ delivers out-of-the-box performance that exceeds specialized single-dataset estimators. Recently, concurrent work by \citet{robertson2025pfn} also applies PFNs to causal effect estimation but lacks a procedure to guarantee the identifiability of the prior data; additionally, we observe relatively poor empirical performance compared to \method. For further discussion and comparison with this method, refer to \cref{appx:concurrent}.

\xhdr{Scaling In-Context Transformers}
In-context learning with transformers has shown impressive results across a range of domains~\citep{brown2020language, xie2021explanation, coda2023meta, dong2024survey,liang2024transformers,vetter2025effortless}. Although the underlying mechanisms responsible for this success remain an active area of research~\citep{akyurek2023learning, dai2022can, olsson2022context, von2023transformers, li2023transformers, yadlowsky2023pretraining, von2023uncovering, bai2024transformers,peyrard2025meta}, increasing model size and training data have consistently and undoubtedly led to stronger performance. This success has recently extended to tabular prediction with models such as TabPFN~\citep{hollmann2023tabpfn, hollmann2025accurate}, TabDPT~\citep{ma2024tabdpt}, and TabICL~\citep{qu2025tabicl}, which are trained on broad prior distributions and perform well on real-world data without fine-tuning. \method~complements these works, demonstrating that—with sufficient scale and training—in-context learning can also be effectively adapted to causal inference.

\section{Conclusions, Limitations, and Future Work}\label{sec:conclusion}

In this paper, we introduced a practical paradigm for amortized causal effect estimation that combines Bayesian causal inference with large-scale tabular training. Despite learning solely from simulated data, \method\ matches, and often outperforms, specialized causal estimators across diverse real-world domains. Through amortization, we significantly reduce the burden of estimator selection at inference time, and to foster adoption, we have open-sourced the code and presets.

That said, several important limitations remain:
\stepindicator{i} Our approach fundamentally assumes strong ignorability, which is an untestable assumption in practice. Without this condition, \method\ has no guarantees of validity. Domain expertise still remains essential to determine whether this method is appropriate or whether alternative approaches should be used.
\stepindicator{ii} Our theoretical guarantees rely on idealistic assumptions: a well-specified prior and asymptotically large datasets. We lack finite-sample theory characterizing the estimator's behavior in practical settings. Investigating robustness to prior misspecification and developing finite-sample guarantees remain open problems. Recent work on theory of valid adjustment sets \citep{choo2024probably} may offer promising directions for addressing these challenges.
\stepindicator{iii} Performance degradation is evident on the largest marketing tables (\cref{tab:uplift-full}), reflective of the known size-scalability trade-off inherent to PFN-style models \citep{hollmann2025accurate}.
\stepindicator{iv} While \method\ already supports multi-arm discrete treatments with a finite set $\cT$, we have only implemented it for the binary $\cT$. Additionally, extending to the continuous treatment setting where $\cT$ is not finite remains fully unexplored. 
\stepindicator{v} Finally, our entire implementation relies on the strong ignorability or backdoor assumption. Extending our framework to richer domain-informed priors like instrumental variables can broaden the framework’s reach, although designing scalable priors for such cases is non-trivial.

\clearpage

\section*{Acknowledgements}
{
We would like to thank Mouloud Belbahri for his suggestions regarding the uplift modeling experiments. RGK gratefully acknowledges support from the Canada Research Chairs Program (CRC-2022-00049) and the Canada CIFAR AI Chairs Program, This research was funded in part by a NFRF Special Call Award (NFRFR-2022-00526) and NSERC Discovery Grant (RGPIN-2022-04546). Resources used in preparing this research were provided, in part, by the Province of Ontario, the Government of Canada through CIFAR, and companies sponsoring the Vector Institute.
}

\bibliographystyle{plainnat}
\bibliography{main}

\begin{thebibliography}{123}
\providecommand{\natexlab}[1]{#1}
\providecommand{\url}[1]{\texttt{#1}}
\expandafter\ifx\csname urlstyle\endcsname\relax
  \providecommand{\doi}[1]{doi: #1}\else
  \providecommand{\doi}{doi: \begingroup \urlstyle{rm}\Url}\fi

\bibitem[Aky{\"u}rek et~al.(2023)Aky{\"u}rek, Schuurmans, Andreas, Ma, and
  Zhou]{akyurek2023learning}
Ekin Aky{\"u}rek, Dale Schuurmans, Jacob Andreas, Tengyu Ma, and Denny Zhou.
\newblock What learning algorithm is in-context learning? investigations with
  linear models.
\newblock In \emph{The Eleventh International Conference on Learning
  Representations}, 2023.

\bibitem[Alaa and Van Der~Schaar(2019)]{alaa2019validating}
Ahmed Alaa and Mihaela Van Der~Schaar.
\newblock Validating causal inference models via influence functions.
\newblock In \emph{International Conference on Machine Learning}, pages
  191--201. PMLR, 2019.

\bibitem[Alaa and van~der Schaar(2017)]{alaa2017bayesian}
Ahmed~M. Alaa and Mihaela van~der Schaar.
\newblock Bayesian inference of individualized treatment effects using
  multi-task gaussian processes.
\newblock In \emph{Advances in Neural Information Processing Systems},
  volume~30, 2017.

\bibitem[Andrieu et~al.(2003)Andrieu, De~Freitas, Doucet, and
  Jordan]{andrieu2003introduction}
Christophe Andrieu, Nando De~Freitas, Arnaud Doucet, and Michael~I Jordan.
\newblock {An introduction to MCMC for machine learning}.
\newblock \emph{Machine Learning}, 50:\penalty0 5--43, 2003.

\bibitem[Angrist and Pischke(2014)]{angrist2014mastering}
Joshua~D Angrist and J{\"o}rn-Steffen Pischke.
\newblock \emph{Mastering 'Metrics: The path from cause to effect}.
\newblock Princeton University Press, 2014.

\bibitem[Angrist et~al.(1996)Angrist, Imbens, and
  Rubin]{angrist1996identification}
Joshua~D Angrist, Guido~W Imbens, and Donald~B Rubin.
\newblock Identification of causal effects using instrumental variables.
\newblock \emph{Journal of the American statistical Association}, 91\penalty0
  (434):\penalty0 444--455, 1996.

\bibitem[Athey et~al.(2019)Athey, Tibshirani, and Wager]{athey2019generalized}
Susan Athey, Julie Tibshirani, and Stefan Wager.
\newblock Generalized random forests.
\newblock \emph{The Annals of Statistics}, 47\penalty0 (2):\penalty0
  1148--1178, 2019.
\newblock \doi{10.1214/18-AOS1709}.

\bibitem[Bai et~al.(2023)Bai, Chen, Wang, Xiong, and Mei]{bai2024transformers}
Yu~Bai, Fan Chen, Huan Wang, Caiming Xiong, and Song Mei.
\newblock Transformers as statisticians: Provable in-context learning with
  in-context algorithm selection.
\newblock In \emph{Advances in Neural Information Processing Systems},
  volume~36, pages 57125--57211, 2023.

\bibitem[Balazadeh et~al.(2024)Balazadeh, Chidambaram, Nguyen, Krishnan, and
  Syrgkanis]{balazadeh2024sequential}
Vahid Balazadeh, Keertana Chidambaram, Viet Nguyen, Rahul~G Krishnan, and
  Vasilis Syrgkanis.
\newblock Sequential decision making with expert demonstrations under
  unobserved heterogeneity.
\newblock \emph{Advances in Neural Information Processing Systems},
  37:\penalty0 65476--65498, 2024.

\bibitem[Balke and Pearl(1997)]{balke1997bounds}
Alexander Balke and Judea Pearl.
\newblock Bounds on treatment effects from studies with imperfect compliance.
\newblock \emph{Journal of the American Statistical Association}, 92\penalty0
  (439):\penalty0 1171--1176, 1997.

\bibitem[Battocchi et~al.(2019)Battocchi, Dillon, Hei, Lewis, Oka, Oprescu, and
  Syrgkanis]{econml}
Keith Battocchi, Eleanor Dillon, Maggie Hei, Greg Lewis, Paul Oka, Miruna
  Oprescu, and Vasilis Syrgkanis.
\newblock {EconML}: {A Python Package for ML-Based Heterogeneous Treatment
  Effects Estimation}.
\newblock https://github.com/py-why/EconML, 2019.
\newblock Version 0.15.0.

\bibitem[Bischl et~al.(2021)Bischl, Casalicchio, Feurer, Gijsbers, Hutter,
  Lang, Gomes~Mantovani, van Rijn, and Vanschoren]{bischl2021cc18}
Bernd Bischl, Giuseppe Casalicchio, Matthias Feurer, Pieter Gijsbers, Frank
  Hutter, Michel Lang, Rafael Gomes~Mantovani, Jan van Rijn, and Joaquin
  Vanschoren.
\newblock {OpenML} benchmarking suites.
\newblock In \emph{Proceedings of the Neural Information Processing Systems
  Track on Datasets and Benchmarks}, 2021.

\bibitem[Brown et~al.(2020)Brown, Mann, Ryder, Subbiah, Kaplan, Dhariwal,
  Neelakantan, Shyam, Sastry, Askell, Agarwal, Herbert-Voss, Krueger, Henighan,
  Child, Ramesh, Ziegler, Wu, Winter, Hesse, Chen, Sigler, Litwin, Gray, Chess,
  Clark, Berner, McCandlish, Radford, Sutskever, and Amodei]{brown2020language}
Tom Brown, Benjamin Mann, Nick Ryder, Melanie Subbiah, Jared~D Kaplan, Prafulla
  Dhariwal, Arvind Neelakantan, Pranav Shyam, Girish Sastry, Amanda Askell,
  Sandhini Agarwal, Ariel Herbert-Voss, Gretchen Krueger, Tom Henighan, Rewon
  Child, Aditya Ramesh, Daniel Ziegler, Jeffrey Wu, Clemens Winter, Chris
  Hesse, Mark Chen, Eric Sigler, Mateusz Litwin, Scott Gray, Benjamin Chess,
  Jack Clark, Christopher Berner, Sam McCandlish, Alec Radford, Ilya Sutskever,
  and Dario Amodei.
\newblock Language models are few-shot learners.
\newblock In \emph{Advances in Neural Information Processing Systems},
  volume~33, pages 1877--1901, 2020.

\bibitem[Bynum et~al.(2025)Bynum, Puli, Herrero-Quevedo, Nguyen,
  Fernandez-Granda, Cho, and Ranganath]{bynum2025black}
Lucius~EJ Bynum, Aahlad~Manas Puli, Diego Herrero-Quevedo, Nhi Nguyen, Carlos
  Fernandez-Granda, Kyunghyun Cho, and Rajesh Ranganath.
\newblock Black box causal inference: Effect estimation via meta prediction.
\newblock \emph{arXiv:2503.05985}, 2025.

\bibitem[Chan et~al.(2022)Chan, Santoro, Lampinen, Wang, Singh, Richemond,
  McClelland, and Hill]{chan2022data}
Stephanie Chan, Adam Santoro, Andrew Lampinen, Jane Wang, Aaditya Singh, Pierre
  Richemond, James McClelland, and Felix Hill.
\newblock Data distributional properties drive emergent in-context learning in
  transformers.
\newblock In \emph{Advances in Neural Information Processing Systems},
  volume~35, 2022.

\bibitem[Chernozhukov et~al.(2018)Chernozhukov, Chetverikov, Demirer, Duflo,
  Hansen, Newey, and Robins]{chernozhukov2017doubledebiased}
Victor Chernozhukov, Denis Chetverikov, Mert Demirer, Esther Duflo, Christian
  Hansen, Whitney Newey, and James Robins.
\newblock Double/debiased machine learning for treatment and structural
  parameters.
\newblock \emph{The Econometrics Journal}, 21\penalty0 (1), 2018.
\newblock \doi{10.1111/ectj.12097}.

\bibitem[Choo et~al.(2025)Choo, Squires, Bhattacharyya, and
  Sontag]{choo2024probably}
Davin Choo, Chandler Squires, Arnab Bhattacharyya, and David Sontag.
\newblock Probably approximately correct high-dimensional causal effect
  estimation given a valid adjustment set.
\newblock In \emph{Proceedings of the Fourth Conference on Causal Learning and
  Reasoning}, volume 275 of \emph{Proceedings of Machine Learning Research},
  2025.

\bibitem[Coda-Forno et~al.(2023)Coda-Forno, Binz, Akata, Botvinick, Wang, and
  Schulz]{coda2023meta}
Julian Coda-Forno, Marcel Binz, Zeynep Akata, Matt Botvinick, Jane Wang, and
  Eric Schulz.
\newblock Meta-in-context learning in large language models.
\newblock \emph{Advances in Neural Information Processing Systems},
  36:\penalty0 65189--65201, 2023.

\bibitem[Curth(2021)]{curth2021catenets}
Alicia Curth.
\newblock {CATENets: Sklearn‑style Implementations of Neural Network‑based
  Conditional Average Treatment Effect (CATE) Estimators}.
\newblock {https://github.com/AliciaCurth/CATENets}, 2021.
\newblock GitHub repository, commit 821bfb0. Accessed: 2025‑05‑11.

\bibitem[Curth and van~der Schaar(2021{\natexlab{a}})]{curth2021inductive}
Alicia Curth and Mihaela van~der Schaar.
\newblock On inductive biases for heterogeneous treatment effect estimation.
\newblock In \emph{Advances in Neural Information Processing Systems},
  volume~34, pages 15883--15894, 2021{\natexlab{a}}.

\bibitem[Curth and van~der Schaar(2021{\natexlab{b}})]{curth2021nonparametric}
Alicia Curth and Mihaela van~der Schaar.
\newblock Nonparametric estimation of heterogeneous treatment effects: From
  theory to learning algorithms.
\newblock In \emph{Proceedings of the 24th International Conference on
  Artificial Intelligence and Statistics (AISTATS)}. PMLR, 2021{\natexlab{b}}.

\bibitem[Curth et~al.(2021)Curth, Svensson, Weatherall, and van~der
  Schaar]{curth2021really}
Alicia Curth, David Svensson, James Weatherall, and Mihaela van~der Schaar.
\newblock {Really Doing Great at Estimating CATE? A Critical Look at ML
  Benchmarking Practices in Treatment Effect Estimation}.
\newblock In \emph{Proceedings of the Neural Information Processing Systems
  Track on Datasets and Benchmarks}, volume~1, 2021.

\bibitem[Dai et~al.(2023)Dai, Sun, Dong, Hao, Ma, Sui, and Wei]{dai2022can}
Damai Dai, Yutao Sun, Li~Dong, Yaru Hao, Shuming Ma, Zhifang Sui, and Furu Wei.
\newblock {Why Can GPT Learn In-Context? Language Models Secretly Perform
  Gradient Descent as Meta-Optimizers}.
\newblock In \emph{Findings of the Association for Computational Linguistics:
  ACL 2023}, pages 4005--4019. Association for Computational Linguistics, 2023.
\newblock \doi{10.18653/v1/2023.findings-acl.247}.

\bibitem[Defazio et~al.(2024)Defazio, Yang, Khaled, Mishchenko, Mehta, and
  Cutkosky]{defazio2024road}
Aaron Defazio, Xingyu Yang, Ahmed Khaled, Konstantin Mishchenko, Harsh Mehta,
  and Ashok Cutkosky.
\newblock The road less scheduled.
\newblock \emph{Advances in Neural Information Processing Systems},
  37:\penalty0 9974--10007, 2024.

\bibitem[Dong et~al.(2024)Dong, Li, Dai, Zheng, Ma, Li, Xia, Xu, Wu, Chang,
  et~al.]{dong2024survey}
Qingxiu Dong, Lei Li, Damai Dai, Ce~Zheng, Jingyuan Ma, Rui Li, Heming Xia,
  Jingjing Xu, Zhiyong Wu, Baobao Chang, et~al.
\newblock A survey on in-context learning.
\newblock In \emph{Proceedings of the 2024 Conference on Empirical Methods in
  Natural Language Processing}, pages 1107--1128, 2024.

\bibitem[Doob(1949)]{doob1949application}
Joseph~L Doob.
\newblock Application of the theory of martingales.
\newblock \emph{Le calcul des probabilites et ses applications}, pages 23--27,
  1949.

\bibitem[Dorie et~al.(2019)Dorie, Hill, Shalit, Scott, and
  Cervone]{dorie2019automated}
Vincent Dorie, Jennifer Hill, Uri Shalit, Marc Scott, and Dan Cervone.
\newblock Automated versus do-it-yourself methods for causal inference: Lessons
  learned from a data analysis competition.
\newblock \emph{Statistical Science}, 34\penalty0 (1):\penalty0 43--68, 2019.

\bibitem[Fischer et~al.(2023)Fischer, Feurer, and Bischl]{fischer2023ctr}
Sebastian~Felix Fischer, Matthias Feurer, and Bernd Bischl.
\newblock {OpenML-CTR23} -- {A} curated tabular regression benchmarking suite.
\newblock In \emph{AutoML Conference (Workshop)}, 2023.

\bibitem[Foster and Syrgkanis(2023)]{foster2023orthogonal}
Dylan~J Foster and Vasilis Syrgkanis.
\newblock Orthogonal statistical learning.
\newblock \emph{The Annals of Statistics}, 51\penalty0 (3):\penalty0 879--908,
  2023.

\bibitem[Funk et~al.(2011)Funk, Westreich, Wiesen, St{\"u}rmer, Brookhart, and
  Davidian]{funk2011doubly}
Michele~Jonsson Funk, Daniel Westreich, Chris Wiesen, Til St{\"u}rmer, M~Alan
  Brookhart, and Marie Davidian.
\newblock Doubly robust estimation of causal effects.
\newblock \emph{American Journal of Epidemiology}, 173\penalty0 (7):\penalty0
  761--767, 2011.

\bibitem[Garnelo et~al.(2018{\natexlab{a}})Garnelo, Rosenbaum, Maddison,
  Ramalho, Saxton, Shanahan, Teh, Rezende, and Eslami]{garnelo2018conditional}
Marta Garnelo, Dan Rosenbaum, Christopher Maddison, Tiago Ramalho, David
  Saxton, Murray Shanahan, Yee~Whye Teh, Danilo Rezende, and S.~M.~Ali Eslami.
\newblock Conditional neural processes.
\newblock In \emph{Proceedings of the 35th International Conference on Machine
  Learning}, volume~80, pages 1704--1713, 2018{\natexlab{a}}.

\bibitem[Garnelo et~al.(2018{\natexlab{b}})Garnelo, Schwarz, Rosenbaum, Viola,
  Rezende, Eslami, and Teh]{garnelo2018neural}
Marta Garnelo, Jonathan Schwarz, Dan Rosenbaum, Fabio Viola, Danilo~J Rezende,
  SM~Eslami, and Yee~Whye Teh.
\newblock Neural processes.
\newblock \emph{arXiv:1807.01622}, 2018{\natexlab{b}}.

\bibitem[Gijsbers et~al.(2024)Gijsbers, Bueno, Coors, LeDell, Poirier, Thomas,
  Bischl, and Vanschoren]{gijsbers2024amlb}
Pieter Gijsbers, Marcos~LP Bueno, Stefan Coors, Erin LeDell, S\'ebastien
  Poirier, Janek Thomas, Bernd Bischl, and Joaquin Vanschoren.
\newblock {AMLB}: An {AutoML} benchmark.
\newblock \emph{Journal of Machine Learning Research}, 25\penalty0
  (101):\penalty0 1--65, 2024.

\bibitem[Grinsztajn et~al.(2022)Grinsztajn, Oyallon, and
  Varoquaux]{grinsztajn2022why}
L\'eo Grinsztajn, Edouard Oyallon, and Ga\"el Varoquaux.
\newblock Why do tree-based models still outperform deep learning on typical
  tabular data?
\newblock In \emph{Advances in Neural Information Processing Systems}, 2022.

\bibitem[Guo et~al.(2017)Guo, Pleiss, Sun, and Weinberger]{guo2017calibration}
Chuan Guo, Geoff Pleiss, Yu~Sun, and Kilian~Q Weinberger.
\newblock On calibration of modern neural networks.
\newblock In \emph{International Conference on Machine Learning}, pages
  1321--1330. PMLR, 2017.

\bibitem[Hahn et~al.(2020)Hahn, Murray, and Carvalho]{hahn2020bayesian}
P~Richard Hahn, Jared~S Murray, and Carlos~M Carvalho.
\newblock Bayesian regression tree models for causal inference: Regularization,
  confounding, and heterogeneous effects (with discussion).
\newblock \emph{Bayesian Analysis}, 15\penalty0 (3):\penalty0 965--1056, 2020.

\bibitem[Helli et~al.(2024)Helli, Schnurr, Hollmann, M{\"u}ller, and
  Hutter]{helli2024drift}
Kai Helli, David Schnurr, Noah Hollmann, Samuel M{\"u}ller, and Frank Hutter.
\newblock {Drift-resilient TabPFN: In-context learning temporal distribution
  shifts on tabular data}.
\newblock \emph{Advances in Neural Information Processing Systems},
  37:\penalty0 98742--98781, 2024.

\bibitem[Henry et~al.(2020)Henry, Dachapally, Pawar, and Chen]{henry2020query}
Alex Henry, Prudhvi~Raj Dachapally, Shubham~Shantaram Pawar, and Yuxuan Chen.
\newblock Query-key normalization for transformers.
\newblock In \emph{Findings of the Association for Computational Linguistics:
  EMNLP 2020}, pages 4246--4253. Association for Computational Linguistics,
  November 2020.
\newblock \doi{10.18653/v1/2020.findings-emnlp.379}.

\bibitem[Hernan and Robins(2023)]{hernan2010causal}
M.A. Hernan and J.M. Robins.
\newblock \emph{Causal Inference: What If}.
\newblock Chapman \& Hall/CRC monographs on statistics \& applied probability.
  Taylor \& Francis, 2023.
\newblock ISBN 9781315374932.

\bibitem[Hill(2011)]{hill2011bayesian}
Jennifer~L Hill.
\newblock Bayesian nonparametric modeling for causal inference.
\newblock \emph{Journal of Computational and Graphical Statistics}, 20\penalty0
  (1):\penalty0 217--240, 2011.

\bibitem[Hillstrom(2008)]{hillstrom2008mining}
Kevin Hillstrom.
\newblock Minethatdata e-mail analytics and data mining challenge dataset.
\newblock
  {https://blog.minethatdata.com/2008/03/minethatdata-e-mail-analytics-and-data.html},
  2008.
\newblock Accessed: 2025-05-11.

\bibitem[Hollmann et~al.(2023)Hollmann, M{\"u}ller, Eggensperger, and
  Hutter]{hollmann2023tabpfn}
Noah Hollmann, Samuel M{\"u}ller, Katharina Eggensperger, and Frank Hutter.
\newblock Tab{PFN}: A transformer that solves small tabular classification
  problems in a second.
\newblock In \emph{The Eleventh International Conference on Learning
  Representations}, 2023.

\bibitem[Hollmann et~al.(2025)Hollmann, M{\"u}ller, Purucker, Krishnakumar,
  K{\"o}rfer, Hoo, Schirrmeister, and Hutter]{hollmann2025accurate}
Noah Hollmann, Samuel M{\"u}ller, Lennart Purucker, Arjun Krishnakumar, Max
  K{\"o}rfer, Shi~Bin Hoo, Robin~Tibor Schirrmeister, and Frank Hutter.
\newblock Accurate predictions on small data with a tabular foundation model.
\newblock \emph{Nature}, 637\penalty0 (8045):\penalty0 319--326, 2025.

\bibitem[Imani and White(2018)]{imani2018improving}
Ehsan Imani and Martha White.
\newblock Improving regression performance with distributional losses.
\newblock In \emph{International conference on machine learning}, pages
  2157--2166. PMLR, 2018.

\bibitem[Imbens and Angrist(1994)]{angrist1995identification}
Guido~W. Imbens and Joshua~D. Angrist.
\newblock Identification and estimation of local average treatment effects.
\newblock \emph{Econometrica}, 62\penalty0 (2):\penalty0 467--475, 1994.
\newblock ISSN 00129682, 14680262.

\bibitem[Imbens and Rubin(1997)]{imbens1997bayesian}
Guido~W Imbens and Donald~B Rubin.
\newblock Bayesian inference for causal effects in randomized experiments with
  noncompliance.
\newblock \emph{The Annals of Statistics}, pages 305--327, 1997.

\bibitem[Imbens and Rubin(2015)]{imbens2015causal}
Guido~W Imbens and Donald~B Rubin.
\newblock \emph{Causal Inference in Statistics, Social, and Biomedical
  Sciences}.
\newblock Cambridge University Press, 2015.

\bibitem[Jesson et~al.(2020)Jesson, Mindermann, Shalit, and
  Gal]{jesson2020identifying}
Andrew Jesson, S{\"o}ren Mindermann, Uri Shalit, and Yarin Gal.
\newblock Identifying causal-effect inference failure with uncertainty-aware
  models.
\newblock \emph{Advances in Neural Information Processing Systems},
  33:\penalty0 11637--11649, 2020.

\bibitem[Jordan et~al.(1999)Jordan, Ghahramani, Jaakkola, and
  Saul]{jordan1999introduction}
Michael~I Jordan, Zoubin Ghahramani, Tommi~S Jaakkola, and Lawrence~K Saul.
\newblock An introduction to variational methods for graphical models.
\newblock \emph{Machine Learning}, 37:\penalty0 183--233, 1999.

\bibitem[Kamkari et~al.(2023)Kamkari, Balazadeh, Zehtab, and
  Krishnan]{kamkari2023order}
Hamidreza Kamkari, Vahid Balazadeh, Vahid Zehtab, and Rahul~G Krishnan.
\newblock Order-based structure learning with normalizing flows.
\newblock \emph{arXiv:2308.07480}, 2023.

\bibitem[Ke et~al.(2023)Ke, Chiappa, Wang, Bornschein, Goyal, Rey, Weber,
  Botvinick, Mozer, and Rezende]{ke2022learning}
Nan~Rosemary Ke, Silvia Chiappa, Jane~X Wang, Jorg Bornschein, Anirudh Goyal,
  Melanie Rey, Theophane Weber, Matthew Botvinick, Michael~Curtis Mozer, and
  Danilo~Jimenez Rezende.
\newblock Learning to induce causal structure.
\newblock In \emph{International Conference on Learning Representations}, 2023.

\bibitem[Kennedy(2023)]{kennedy2020optimal}
Edward~H. Kennedy.
\newblock Towards optimal doubly robust estimation of heterogeneous causal
  effects.
\newblock \emph{Electronic Journal of Statistics}, 17\penalty0 (2):\penalty0
  3008--3049, 2023.

\bibitem[Khemakhem et~al.(2021)Khemakhem, Monti, Leech, and
  Hyvarinen]{khemakhem2021causal}
Ilyes Khemakhem, Ricardo Monti, Robert Leech, and Aapo Hyvarinen.
\newblock Causal autoregressive flows.
\newblock In \emph{International Conference on Artificial Antelligence and
  Statistics}, pages 3520--3528. PMLR, 2021.

\bibitem[Kim et~al.(2019)Kim, Mnih, Schwarz, Garnelo, Eslami, Rosenbaum,
  Vinyals, and Teh]{kim2019attentive}
Hyunjik Kim, Andriy Mnih, Jonathan Schwarz, Marta Garnelo, Ali Eslami, Dan
  Rosenbaum, Oriol Vinyals, and Yee~Whye Teh.
\newblock Attentive neural processes.
\newblock In \emph{International Conference on Learning Representations}, 2019.

\bibitem[Kingma(2014)]{kingma2014adam}
Diederik~P Kingma.
\newblock Adam: A method for stochastic optimization.
\newblock \emph{arXiv preprint arXiv:1412.6980}, 2014.

\bibitem[Konstantinov et~al.(2023)Konstantinov, Kirpichenko, and
  Utkin]{konstantinov2023heterogeneous}
Andrei Konstantinov, Stanislav Kirpichenko, and Lev Utkin.
\newblock Heterogeneous treatment effect with trained kernels of the
  nadaraya--watson regression.
\newblock \emph{Algorithms}, 16\penalty0 (5):\penalty0 226, 2023.

\bibitem[K{\"u}nzel et~al.(2019)K{\"u}nzel, Sekhon, Bickel, and
  Yu]{kunzel2019metalearners}
S{\"o}ren~R. K{\"u}nzel, Jasjeet~S. Sekhon, Peter~J. Bickel, and Bin Yu.
\newblock Metalearners for estimating heterogeneous treatment effects using
  machine learning.
\newblock \emph{Proceedings of the National Academy of Sciences}, 116\penalty0
  (10):\penalty0 4156--4165, 2019.

\bibitem[LaLonde(1986)]{lalonde1986evaluating}
Robert~J LaLonde.
\newblock Evaluating the econometric evaluations of training programs with
  experimental data.
\newblock \emph{The American Economic Review}, pages 604--620, 1986.

\bibitem[Lau et~al.(2024)Lau, Choi, Balazadeh, Chidambaram, Syrgkanis, and
  Krishnan]{lau2024personalized}
Allison Lau, Younwoo Choi, Vahid Balazadeh, Keertana Chidambaram, Vasilis
  Syrgkanis, and Rahul~G Krishnan.
\newblock Personalized adaptation via in-context preference learning.
\newblock \emph{arXiv preprint arXiv:2410.14001}, 2024.

\bibitem[Lee et~al.(2023)Lee, Xie, Pacchiano, Chandak, Finn, Nachum, and
  Brunskill]{lee2023supervised}
Jonathan Lee, Annie Xie, Aldo Pacchiano, Yash Chandak, Chelsea Finn, Ofir
  Nachum, and Emma Brunskill.
\newblock Supervised pretraining can learn in-context reinforcement learning.
\newblock \emph{Advances in Neural Information Processing Systems},
  36:\penalty0 43057--43083, 2023.

\bibitem[{Lenta LLC}(2020)]{lenta2020}
{Lenta LLC}.
\newblock Lenta uplift dataset.
\newblock {https://github.com/maks-sh/scikit-uplift}, 2020.
\newblock Accessed: 2025-05-11.

\bibitem[Li et~al.(2023{\natexlab{a}})Li, Ding, and Mealli]{li2023bayesian}
Fan Li, Peng Ding, and Fabrizia Mealli.
\newblock Bayesian causal inference: a critical review.
\newblock \emph{Philosophical Transactions of the Royal Society A},
  381\penalty0 (2247):\penalty0 20220153, 2023{\natexlab{a}}.

\bibitem[Li et~al.(2023{\natexlab{b}})Li, Ildiz, Papailiopoulos, and
  Oymak]{li2023transformers}
Yingcong Li, Muhammed~Emrullah Ildiz, Dimitris Papailiopoulos, and Samet Oymak.
\newblock Transformers as algorithms: Generalization and stability in
  in-context learning.
\newblock In \emph{International Conference on Machine Learning}, pages
  19565--19594. PMLR, 2023{\natexlab{b}}.

\bibitem[Liang et~al.(2024)Liang, Balasubramanian, and
  Lai]{liang2024transformers}
Haodong Liang, Krishnakumar Balasubramanian, and Lifeng Lai.
\newblock Transformers handle endogeneity in in-context linear regression.
\newblock \emph{arXiv preprint arXiv:2410.01265}, 2024.

\bibitem[Linero and Antonelli(2023)]{linero2023and}
Antonio~R Linero and Joseph~L Antonelli.
\newblock The how and why of bayesian nonparametric causal inference.
\newblock \emph{Wiley Interdisciplinary Reviews: Computational Statistics},
  15\penalty0 (1):\penalty0 e1583, 2023.

\bibitem[Liu et~al.(2024)Liu, Bellamy, and Beam]{liu2024dag}
Manqing Liu, David~R Bellamy, and Andrew~L Beam.
\newblock {DAG-aware transformer for causal effect estimation}.
\newblock \emph{arXiv:2410.10044}, 2024.

\bibitem[Liu and Ye(2025)]{liu2025tabpfn}
Siyang Liu and Han-Jia Ye.
\newblock {TabPFN Unleashed: A Scalable and Effective Solution to Tabular
  Classification Problems}.
\newblock In \emph{Forty-second International Conference on Machine Learning},
  2025.

\bibitem[Lorch et~al.(2022)Lorch, Sussex, Rothfuss, Krause, and
  Sch{\"o}lkopf]{lorch2022amortized}
Lars Lorch, Scott Sussex, Jonas Rothfuss, Andreas Krause, and Bernhard
  Sch{\"o}lkopf.
\newblock Amortized inference for causal structure learning.
\newblock \emph{Advances in Neural Information Processing Systems},
  35:\penalty0 13104--13118, 2022.

\bibitem[Louizos et~al.(2017)Louizos, Shalit, Mooij, Sontag, Zemel, and
  Welling]{louizos2017causal}
Christos Louizos, Uri Shalit, Joris~M Mooij, David Sontag, Richard Zemel, and
  Max Welling.
\newblock Causal effect inference with deep latent-variable models.
\newblock In \emph{Advances in Neural Information Processing Systems},
  volume~30, 2017.

\bibitem[Ma et~al.(2025)Ma, Thomas, Hosseinzadeh, Kamkari, Labach, Cresswell,
  Golestan, Yu, Caterini, and Volkovs]{ma2024tabdpt}
Junwei Ma, Valentin Thomas, Rasa Hosseinzadeh, Hamidreza Kamkari, Alex Labach,
  Jesse~C Cresswell, Keyvan Golestan, Guangwei Yu, Anthony~L Caterini, and
  Maksims Volkovs.
\newblock {TabDPT: Scaling Tabular Foundation Models on Real Data}.
\newblock In \emph{Advances in Neural Information Processing Systems},
  volume~38, 2025.

\bibitem[Ma et~al.(2024)Ma, Melnychuk, Schweisthal, and
  Feuerriegel]{ma2024diffpo}
Yuchen Ma, Valentyn Melnychuk, Jonas Schweisthal, and Stefan Feuerriegel.
\newblock {DiffPO: A causal diffusion model for learning distributions of
  potential outcomes}.
\newblock In \emph{Advances in Neural Information Processing Systems},
  volume~37, pages 43663--43692, 2024.

\bibitem[MacKinnon et~al.(2007)MacKinnon, Fairchild, and
  Fritz]{mackinnon2007mediation}
David~P MacKinnon, Amanda~J Fairchild, and Matthew~S Fritz.
\newblock Mediation analysis.
\newblock \emph{Annu. Rev. Psychol.}, 58\penalty0 (1):\penalty0 593--614, 2007.

\bibitem[Mahajan et~al.(2024{\natexlab{a}})Mahajan, Gladrow, Hilmkil, Zhang,
  and Scetbon]{mahajan2024zero}
Divyat Mahajan, Jannes Gladrow, Agrin Hilmkil, Cheng Zhang, and Meyer Scetbon.
\newblock Zero-shot learning of causal models.
\newblock \emph{arXiv:2410.06128}, 2024{\natexlab{a}}.

\bibitem[Mahajan et~al.(2024{\natexlab{b}})Mahajan, Mitliagkas, Neal, and
  Syrgkanis]{mahajan2024empirical}
Divyat Mahajan, Ioannis Mitliagkas, Brady Neal, and Vasilis Syrgkanis.
\newblock Empirical analysis of model selection for heterogeneous causal effect
  estimation.
\newblock In \emph{The Twelfth International Conference on Learning
  Representations}, 2024{\natexlab{b}}.

\bibitem[Maksim~Shevchenko(2020)]{user-guide-for-uplift-modeling}
Irina~Elisova Maksim~Shevchenko.
\newblock User guide for uplift modeling and casual inference.
\newblock {https://www.uplift-modeling.com/en/latest/user\_guide/index.html},
  2020.

\bibitem[Manski(1993)]{manski1993identification}
Charles~F Manski.
\newblock Identification problems in the social sciences.
\newblock \emph{Sociological methodology}, pages 1--56, 1993.

\bibitem[McCarter(2024)]{mccarter2025exactly}
Calvin McCarter.
\newblock {What exactly has TabPFN learned to do?}
\newblock In \emph{The Third Blogpost Track at ICLR 2024}, 2024.

\bibitem[McElfresh et~al.(2023)McElfresh, Khandagale, Valverde, Prasad~C,
  Ramakrishnan, Goldblum, and White]{mcelfresh2023neural}
Duncan McElfresh, Sujay Khandagale, Jonathan Valverde, Vishak Prasad~C, Ganesh
  Ramakrishnan, Micah Goldblum, and Colin White.
\newblock When do neural nets outperform boosted trees on tabular data?
\newblock In \emph{Advances in Neural Information Processing Systems}, 2023.

\bibitem[{Megafon PJSC}(2020)]{megafon2020}
{Megafon PJSC}.
\newblock Megafon uplift dataset.
\newblock {https://github.com/maks-sh/scikit-uplift}, 2020.
\newblock Accessed: 2025-05-11.

\bibitem[Miller(2018)]{miller2018detailed}
Jeffrey~W Miller.
\newblock {A detailed treatment of Doob's theorem}.
\newblock \emph{arXiv:1801.03122}, 2018.

\bibitem[M{\"u}ller et~al.(2022)M{\"u}ller, Hollmann, Arango, Grabocka, and
  Hutter]{müller2023transformers}
Samuel M{\"u}ller, Noah Hollmann, Sebastian~Pineda Arango, Josif Grabocka, and
  Frank Hutter.
\newblock {Transformers Can Do Bayesian Inference}.
\newblock In \emph{International Conference on Learning Representations}, 2022.

\bibitem[Neal et~al.(2020)Neal, Huang, and Raghupathi]{neal2020realcause}
Brady Neal, Chin-Wei Huang, and Sunand Raghupathi.
\newblock {RealCause: Realistic causal inference benchmarking}.
\newblock \emph{arXiv:2011.15007}, 2020.

\bibitem[Neal(2012)]{neal2012bayesian}
Radford~M Neal.
\newblock \emph{Bayesian learning for neural networks}, volume 118.
\newblock Springer Science \& Business Media, 2012.

\bibitem[Nilforoshan et~al.(2023)Nilforoshan, Moor, Roohani, Chen,
  {\v{S}}urina, Yasunaga, Oblak, and Leskovec]{nilforoshan2023zero}
Hamed Nilforoshan, Michael Moor, Yusuf Roohani, Yining Chen, Anja {\v{S}}urina,
  Michihiro Yasunaga, Sara Oblak, and Jure Leskovec.
\newblock Zero-shot causal learning.
\newblock In \emph{Advances in Neural Information Processing Systems},
  volume~36, pages 6862--6901, 2023.

\bibitem[Oganisian and Roy(2021)]{oganisian2021practical}
Arman Oganisian and Jason~A Roy.
\newblock A practical introduction to bayesian estimation of causal effects:
  Parametric and nonparametric approaches.
\newblock \emph{Statistics in Medicine}, 40\penalty0 (2):\penalty0 518--551,
  2021.

\bibitem[Olsson et~al.(2022)Olsson, Elhage, Nanda, Joseph, DasSarma, Henighan,
  Mann, Askell, Bai, Chen, Conerly, Drain, Ganguli, Hatfield-Dodds, Hernandez,
  Johnston, Jones, Kernion, Lovitt, Ndousse, Amodei, Brown, Clark, Kaplan,
  McCandlish, and Olah]{olsson2022context}
Catherine Olsson, Nelson Elhage, Neel Nanda, Nicholas Joseph, Nova DasSarma,
  Tom Henighan, Ben Mann, Amanda Askell, Yuntao Bai, Anna Chen, Tom Conerly,
  Dawn Drain, Deep Ganguli, Zac Hatfield-Dodds, Danny Hernandez, Scott
  Johnston, Andy Jones, Jackson Kernion, Liane Lovitt, Kamal Ndousse, Dario
  Amodei, Tom Brown, Jack Clark, Jared Kaplan, Sam McCandlish, and Chris Olah.
\newblock In-context learning and induction heads.
\newblock \emph{arXiv:2209.11895}, 2022.

\bibitem[Ovadia et~al.(2019)Ovadia, Fertig, Ren, Nado, Sculley, Nowozin,
  Dillon, Lakshminarayanan, and Snoek]{ovadia2019can}
Yaniv Ovadia, Emily Fertig, Jie Ren, Zachary Nado, D.~Sculley, Sebastian
  Nowozin, Joshua Dillon, Balaji Lakshminarayanan, and Jasper Snoek.
\newblock Can you trust your model\textquotesingle s uncertainty? evaluating
  predictive uncertainty under dataset shift.
\newblock In \emph{Advances in Neural Information Processing Systems},
  volume~32, 2019.

\bibitem[Pearl(2009)]{pearl2009causality}
Judea Pearl.
\newblock \emph{Causality}.
\newblock Cambridge University Press, 2009.

\bibitem[Peters et~al.(2014)Peters, Mooij, Janzing, and
  Sch{\"o}lkopf]{peters2014causal}
Jonas Peters, Joris~M Mooij, Dominik Janzing, and Bernhard Sch{\"o}lkopf.
\newblock Causal discovery with continuous additive noise models.
\newblock \emph{The Journal of Machine Learning Research}, 15\penalty0
  (1):\penalty0 2009--2053, 2014.

\bibitem[Peters et~al.(2017)Peters, Janzing, and
  Sch{\"o}lkopf]{peters2017elements}
Jonas Peters, Dominik Janzing, and Bernhard Sch{\"o}lkopf.
\newblock \emph{Elements of causal inference: Foundations and learning
  algorithms}.
\newblock The MIT Press, 2017.

\bibitem[Peyrard and Cho(2025)]{peyrard2025meta}
Maxime Peyrard and Kyunghyun Cho.
\newblock Meta-statistical learning: Supervised learning of statistical
  inference.
\newblock \emph{arXiv:2502.12088}, 2025.

\bibitem[Preston(2009)]{preston2009note}
Chris Preston.
\newblock {A note on standard Borel and related spaces}.
\newblock \emph{Journal of Contemporary Mathematical Analysis}, 44\penalty0
  (1):\penalty0 63--71, 2009.

\bibitem[Qu et~al.(2025)Qu, Holzm{\"u}ller, Varoquaux, and
  Morvan]{qu2025tabicl}
Jingang Qu, David Holzm{\"u}ller, Ga{\"e}l Varoquaux, and Marine~Le Morvan.
\newblock {TabICL: A Tabular Foundation Model for In-Context Learning on Large
  Data}.
\newblock In \emph{Forty-second International Conference on Machine Learning},
  2025.

\bibitem[Radcliffe(2007)]{radcliffe2007using}
Nicholas Radcliffe.
\newblock Using control groups to target on predicted lift: Building and
  assessing uplift models.
\newblock Technical report, Stochastic Solutions, 2007.

\bibitem[Ramey et~al.(1992)Ramey, Bryant, Wasik, Sparling, Fendt, and
  La~Vange]{ramey1992infant}
Craig~T Ramey, Donna~M Bryant, Barbara~H Wasik, Joseph~J Sparling, Kaye~H
  Fendt, and Lisa~M La~Vange.
\newblock Infant health and development program for low birth weight, premature
  infants: Program elements, family participation, and child intelligence.
\newblock \emph{Pediatrics}, 89\penalty0 (3):\penalty0 454--465, 1992.

\bibitem[{Retail Hero}(2020)]{retailhero2020}
{Retail Hero}.
\newblock Retail hero (x5) uplift dataset.
\newblock {https://github.com/maks-sh/scikit-uplift}, 2020.
\newblock Accessed: 2025-05-11.

\bibitem[Robertson et~al.(2025)Robertson, Reuter, Guo, Hollmann, Hutter, and
  Sch{\"o}lkopf]{robertson2025pfn}
Jake Robertson, Arik Reuter, Siyuan Guo, Noah Hollmann, Frank Hutter, and
  Bernhard Sch{\"o}lkopf.
\newblock Do-pfn: In-context learning for causal effect estimation.
\newblock \emph{arXiv preprint arXiv:2506.06039}, 2025.

\bibitem[Rosenbaum and Rubin(1983)]{rosenbaum1983central}
Paul~R Rosenbaum and Donald~B Rubin.
\newblock The central role of the propensity score in observational studies for
  causal effects.
\newblock \emph{Biometrika}, 70\penalty0 (1):\penalty0 41--55, 1983.

\bibitem[Rubin(1974)]{rubin1974estimating}
Donald~B Rubin.
\newblock Estimating causal effects of treatments in randomized and
  nonrandomized studies.
\newblock \emph{Journal of Educational Psychology}, 66\penalty0 (5):\penalty0
  688, 1974.

\bibitem[Rubin(1978)]{rubin1978bayesian}
Donald~B Rubin.
\newblock Bayesian inference for causal effects: The role of randomization.
\newblock \emph{The Annals of Statistics}, pages 34--58, 1978.

\bibitem[Rubin(2005)]{rubin2005causal}
Donald~B Rubin.
\newblock Causal inference using potential outcomes: Design, modeling,
  decisions.
\newblock \emph{Journal of the American Statistical Association}, 100\penalty0
  (469):\penalty0 322--331, 2005.

\bibitem[Sawant et~al.(2018)Sawant, Namballa, Sadagopan, and
  Nassif]{sawant2018contextual}
Neela Sawant, Chitti~Babu Namballa, Narayanan Sadagopan, and Houssam Nassif.
\newblock Contextual multi-armed bandits for causal marketing.
\newblock \emph{arXiv preprint arXiv:1810.01859}, 2018.

\bibitem[Scetbon et~al.(2024)Scetbon, Jennings, Hilmkil, Zhang, and
  Ma]{scetbon2024fixed}
Meyer Scetbon, Joel Jennings, Agrin Hilmkil, Cheng Zhang, and Chao Ma.
\newblock A fixed-point approach for causal generative modeling.
\newblock In \emph{Proceedings of the 41st International Conference on Machine
  Learning}, volume 235, pages 43504--43541, 2024.

\bibitem[Schuler et~al.(2018)Schuler, Baiocchi, Tibshirani, and
  Shah]{schuler2018comparison}
Alejandro Schuler, Michael Baiocchi, Robert Tibshirani, and Nigam Shah.
\newblock A comparison of methods for model selection when estimating
  individual treatment effects.
\newblock \emph{arXiv:1804.05146}, 2018.

\bibitem[Shalit et~al.(2017)Shalit, Johansson, and
  Sontag]{shalit2017estimating}
Uri Shalit, Fredrik~D. Johansson, and David Sontag.
\newblock Estimating individual treatment effect: generalization bounds and
  algorithms.
\newblock In \emph{Proceedings of the 34th International Conference on Machine
  Learning}, volume~70 of \emph{Proceedings of Machine Learning Research},
  pages 3076--3085, 2017.

\bibitem[Shazeer(2020)]{shazeer2020glu}
Noam Shazeer.
\newblock {GLU variants improve transformer}.
\newblock \emph{arXiv:2002.05202}, 2020.

\bibitem[Shi et~al.(2019)Shi, Blei, and Veitch]{shi2019adapting}
Claudia Shi, David Blei, and Victor Veitch.
\newblock Adapting neural networks for the estimation of treatment effects.
\newblock In \emph{Advances in Neural Information Processing Systems},
  volume~32, 2019.

\bibitem[Shimoni et~al.(2018)Shimoni, Yanover, Karavani, and
  Goldschmnidt]{shimoni2018benchmarking}
Yishai Shimoni, Chen Yanover, Ehud Karavani, and Yaara Goldschmnidt.
\newblock Benchmarking framework for performance-evaluation of causal inference
  analysis.
\newblock \emph{arXiv:1802.05046}, 2018.

\bibitem[Srivastava(1998)]{srivastava1998course}
Sashi~Mohan Srivastava.
\newblock \emph{{A course on Borel sets}}.
\newblock Springer, 1998.

\bibitem[Thomas et~al.(2024)Thomas, Ma, Hosseinzadeh, Golestan, Yu, Volkovs,
  and Caterini]{thomas2024retrieval}
Valentin Thomas, Junwei Ma, Rasa Hosseinzadeh, Keyvan Golestan, Guangwei Yu,
  Maksims Volkovs, and Anthony Caterini.
\newblock Retrieval \& fine-tuning for in-context tabular models.
\newblock In \emph{Advances in Neural Information Processing Systems},
  volume~37, pages 108439--108467, 2024.

\bibitem[Vetter et~al.(2025)Vetter, Gloeckler, Gedon, and
  Macke]{vetter2025effortless}
Julius Vetter, Manuel Gloeckler, Daniel Gedon, and Jakob~H Macke.
\newblock Effortless, simulation-efficient bayesian inference using tabular
  foundation models.
\newblock \emph{arXiv:2504.17660}, 2025.

\bibitem[Von~Oswald et~al.(2023)Von~Oswald, Niklasson, Randazzo, Sacramento,
  Mordvintsev, Zhmoginov, and Vladymyrov]{von2023transformers}
Johannes Von~Oswald, Eyvind Niklasson, Ettore Randazzo, Jo{\~a}o Sacramento,
  Alexander Mordvintsev, Andrey Zhmoginov, and Max Vladymyrov.
\newblock Transformers learn in-context by gradient descent.
\newblock In \emph{International Conference on Machine Learning}, pages
  35151--35174. PMLR, 2023.

\bibitem[von Oswald et~al.(2023)von Oswald, Schlegel, Meulemans, Kobayashi,
  Niklasson, Zucchet, Scherrer, Miller, Sandler, y~Arcas, Vladymyrov, Pascanu,
  and Sacramento]{von2023uncovering}
Johannes von Oswald, Maximilian Schlegel, Alexander Meulemans, Seijin
  Kobayashi, Eyvind Niklasson, Nicolas Zucchet, Nino Scherrer, Nolan Miller,
  Mark Sandler, Blaise~Agüera y~Arcas, Max Vladymyrov, Razvan Pascanu, and
  João Sacramento.
\newblock Uncovering mesa-optimization algorithms in transformers.
\newblock \emph{arXiv:2309.05858}, 2023.

\bibitem[Wang et~al.(2021)Wang, Wu, Weimer, and Zhu]{Wang_FLAML_A_Fast_2021}
Chi Wang, Qingyun Wu, Markus Weimer, and Erkang Zhu.
\newblock {FLAML: A Fast and Lightweight AutoML Library}.
\newblock In \emph{Proceedings of Machine Learning and Systems}, volume~3,
  2021.

\bibitem[Xia et~al.(2021)Xia, Lee, Bengio, and Bareinboim]{xia2021causal}
Kevin Xia, Kai-Zhan Lee, Yoshua Bengio, and Elias Bareinboim.
\newblock The causal-neural connection: Expressiveness, learnability, and
  inference.
\newblock \emph{Advances in Neural Information Processing Systems},
  34:\penalty0 10823--10836, 2021.

\bibitem[Xia et~al.(2023)Xia, Pan, and Bareinboim]{xia2022neural}
Kevin~Muyuan Xia, Yushu Pan, and Elias Bareinboim.
\newblock Neural causal models for counterfactual identification and
  estimation.
\newblock In \emph{The Eleventh International Conference on Learning
  Representations}, 2023.

\bibitem[Xie et~al.(2022)Xie, Raghunathan, Liang, and Ma]{xie2021explanation}
Sang~Michael Xie, Aditi Raghunathan, Percy Liang, and Tengyu Ma.
\newblock {An Explanation of In-context Learning as Implicit Bayesian
  Inference}.
\newblock In \emph{International Conference on Learning Representations}, 2022.

\bibitem[Yadlowsky et~al.(2023)Yadlowsky, Doshi, and
  Tripuraneni]{yadlowsky2023pretraining}
Steve Yadlowsky, Lyric Doshi, and Nilesh Tripuraneni.
\newblock Pretraining data mixtures enable narrow model selection capabilities
  in transformer models.
\newblock \emph{arXiv:2311.00871}, 2023.

\bibitem[Ye et~al.(2025)Ye, Liu, and Chao]{ye2025closer}
Han-Jia Ye, Si-Yang Liu, and Wei-Lun Chao.
\newblock {A closer look at TabPFN v2: Strength, limitation, and extension}.
\newblock \emph{arXiv:2502.17361}, 2025.

\bibitem[Yuan et~al.(2013)Yuan, Wang, and Zhao]{yuan2013real}
Shuai Yuan, Jun Wang, and Xiaoxue Zhao.
\newblock Real-time bidding for online advertising: measurement and analysis.
\newblock In \emph{Proceedings of the seventh international workshop on data
  mining for online advertising}, pages 1--8, 2013.

\bibitem[Zhang et~al.(2023)Zhang, Jennings, Hilmkil, Pawlowski, Zhang, and
  Ma]{zhang2023towards}
Jiaqi Zhang, Joel Jennings, Agrin Hilmkil, Nick Pawlowski, Cheng Zhang, and
  Chao Ma.
\newblock Towards causal foundation model: on duality between causal inference
  and attention.
\newblock \emph{arXiv:2310.00809}, 2023.

\bibitem[Zheng et~al.(2018)Zheng, Aragam, Ravikumar, and Xing]{zheng2018dags}
Xun Zheng, Bryon Aragam, Pradeep~K Ravikumar, and Eric~P Xing.
\newblock {DAGs with NO TEARS: Continuous Optimization for Structure Learning}.
\newblock In \emph{Advances in Neural Information Processing Systems},
  volume~31, 2018.

\bibitem[Émilie Diemert et~al.(2018)Émilie Diemert, Teytaud, Oblé, and
  Meynet]{diemert2018criteo}
Émilie Diemert, Olivier Teytaud, Guillaume Oblé, and Florent Meynet.
\newblock A large scale benchmark for uplift modeling.
\newblock In \emph{Proceedings of the AdKDD Workshop}, 2018.

\end{thebibliography}
\clearpage

\section*{NeurIPS Paper Checklist}

\begin{enumerate}

\item {\bf Claims}
    \item[] Question: Do the main claims made in the abstract and introduction accurately reflect the paper's contributions and scope?
    \item[] Answer: \answerYes{} 
    \item[] Justification: We provide a concrete summary of contributions at the end of the introduction.

\item {\bf Limitations}
    \item[] Question: Does the paper discuss the limitations of the work performed by the authors?
    \item[] Answer: \answerYes{} 
    \item[] Justification: We summarize our limitations, alongside future directions, in \cref{sec:conclusion}. We are also fully transparent in the limitations of our theory, and also some of the practical limitations of our method detailed in \cref{sec:experiments}.

\item {\bf Theory assumptions and proofs}
    \item[] Question: For each theoretical result, does the paper provide the full set of assumptions and a complete (and correct) proof?
    \item[] Answer: \answerYes{}{} 
    \item[] Justification: We provide all the assumptions and the complete proof for \cref{prop:cepo-ppd-consistency} in \cref{appx:theory}. We also provide a proof sketch in the main paper.

    \item {\bf Experimental result reproducibility}
    \item[] Question: Does the paper fully disclose all the information needed to reproduce the main experimental results of the paper to the extent that it affects the main claims and/or conclusions of the paper (regardless of whether the code and data are provided or not)?
    \item[] Answer: \answerYes{} 
    \item[] Justification: We provide the code, along with the model checkpoints and Jupyter Notebooks to replicate all the experiments in the main paper and appendix.

\item {\bf Open access to data and code}
    \item[] Question: Does the paper provide open access to the data and code, with sufficient instructions to faithfully reproduce the main experimental results, as described in supplemental material?
    \item[] Answer: \answerYes{} 
    \item[] Justification: We publish \method\ as a standalone PyPI package (\url{https://pypi.org/project/causalpfn/}), along with the instructions to reproduce all the results in the paper. The training data is fully public and is sufficiently reproducible from the given implementation details provided. 

\item {\bf Experimental setting/details}
    \item[] Question: Does the paper specify all the training and test details (e.g., data splits, hyperparameters, how they were chosen, type of optimizer, etc.) necessary to understand the results?
    \item[] Answer: \answerYes{} 
    \item[] Justification: The paper specifies all of the important details in the main text, in addition to extra details in \cref{sec:appx-exp-details}. Moreover, the package we release contains all of the necessary hyperparameters used for inference.

\item {\bf Experiment statistical significance}
    \item[] Question: Does the paper report error bars suitably and correctly defined or other appropriate information about the statistical significance of the experiments?
    \item[] Answer: \answerYes{} 
    \item[] Justification: We provide the 1-sigma standard errors of the mean values in the CATE \& ATE results table, across different realizations of each benchmark dataset. We also demonstrate error bars in the calibration plots, which show 1-sigma standard errors of the mean calibration curves, across multiple samples of the synthetic datasets.

\item {\bf Experiments compute resources}
    \item[] Question: For each experiment, does the paper provide sufficient information on the computer resources (type of compute workers, memory, time of execution) needed to reproduce the experiments?
    \item[] Answer: \answerYes{} 
    \item[] Justification: We use a single A100 GPU for 7 days to train the base \method. We also use an H100 GPU for 2 days to produce the TabPFN fine-tuned results in \cref{tab:tabpfn_comparison}. Apart from that, all of the other experiments are run on either a desktop RTX6000, A100, or an H100.
    
\item {\bf Code of ethics}
    \item[] Question: Does the research conducted in the paper conform, in every respect, with the NeurIPS Code of Ethics \url{https://neurips.cc/public/EthicsGuidelines}?
    \item[] Answer: \answerYes{} 
    \item[] Justification: All the datasets used in the paper were either synthetically generated or publicly available. The authors confirm that the research conducted in the paper complies with the NeurIPS Code of Ethics, to the best of their knowledge.

\item {\bf Broader impacts}
    \item[] Question: Does the paper discuss both potential positive societal impacts and negative societal impacts of the work performed?
    \item[] Answer: \answerYes{} 
    \item[] Justification: Causal effect estimation is a fundamental problem with many societal benefits across public policy, healthcare, and economics. While we do not directly try to solve any critical societal issues, by developing a strong causal estimator, we believe it may lead to positive impact in the future.
    
\item {\bf Safeguards}
    \item[] Question: Does the paper describe safeguards that have been put in place for responsible release of data or models that have a high risk for misuse (e.g., pretrained language models, image generators, or scraped datasets)?
    \item[] Answer: \answerNA{} 
    \item[] Justification: We believe our model poses no such risks, as it is a method for causal effect estimation for tabular observational datasets, with reliable credible intervals. 

\item {\bf Licenses for existing assets}
    \item[] Question: Are the creators or original owners of assets (e.g., code, data, models), used in the paper, properly credited and are the license and terms of use explicitly mentioned and properly respected?
    \item[] Answer: \answerYes{} 
    \item[] Justification: All of the project developers are authors in the paper and are properly credited for their contribution.

\item {\bf New assets}
    \item[] Question: Are new assets introduced in the paper well documented and is the documentation provided alongside the assets?
    \item[] Answer: \answerYes{} 
    \item[] Justification: We attach and release all of the assets and code related to this document. All of the code is well-documented and transparent.

\item {\bf Crowdsourcing and research with human subjects}
    \item[] Question: For crowdsourcing experiments and research with human subjects, does the paper include the full text of instructions given to participants and screenshots, if applicable, as well as details about compensation (if any)? 
    \item[] Answer: \answerNA{} 
    \item[] Justification: Our research did not involve crowdsourcing nor research with human subjects.

\item {\bf Institutional review board (IRB) approvals or equivalent for research with human subjects}
    \item[] Question: Does the paper describe potential risks incurred by study participants, whether such risks were disclosed to the subjects, and whether Institutional Review Board (IRB) approvals (or an equivalent approval/review based on the requirements of your country or institution) were obtained?
    \item[] Answer: \answerNA{} 
    \item[] Justification: Our research did not involve crowdsourcing nor research with human subjects.

\item {\bf Declaration of LLM usage}
    \item[] Question: Does the paper describe the usage of LLMs if it is an important, original, or non-standard component of the core methods in this research? Note that if the LLM is used only for writing, editing, or formatting purposes and does not impact the core methodology, scientific rigorousness, or originality of the research, declaration is not required.
    \item[] Answer: \answerNA{} 
    \item[] Justification: We did not use LLMs for any important and original contributions in this research. 

\end{enumerate}

\clearpage

\appendix

\section*{Appendix Contents}
\addcontentsline{toc}{section}{Appendix Contents}
\startcontents[sections]
\printcontents[sections]{}{1}{\setcounter{tocdepth}{2}}
\clearpage
\section{Notation, Definitions, and Assumptions}
\label{appx:notation}

\xhdr{Sample Space} 
Let $\cB$ denote the Borel $\sigma$-algebra on $\R$. Let $Z \!=\! \lpar \vect{X}, T, Y \rpar$ collect the observed variables, taking values in a standard Borel space $(\cZ, \cB_\cZ)$. In particular, $\vect{X} \in \cX$, $T \in \cT$ where $\cT$ is finite, and $Y \in \R$. To reason about counterfactuals, define the augmented variable $\tilde{Z} \!=\! \lpar \vect{X}, T, \{Y_t\}_{t \in \cT}, Y \rpar$ on a standard Borel space $(\tilde{\cZ}, \cB_{\tilde{\cZ}})$. 

\xhdr{Data–Generating Parameters}
Let $(\Psi,\cB_\Psi)$ be a standard Borel parameter space. For $\psi\in\Psi$, a data–generating process (DGP) is a probability measure $\dist^\psi$ on $(\tilde\cZ,\cB_{\tilde\cZ})$, which induces the observational marginal $\dist_{\obs}^\psi$ on $(\cZ,\cB_{\cZ})$. We use $\psi$ to denote both the random parameter (when distributed according to a prior) and its realized value, when clear from context.

We impose a mild regularity condition to ensure measurability of parameter–to–law maps. Let $\cP(\tilde\cZ)$ and $\cP(\cZ)$ denote the spaces of probability measures on $\tilde\cZ$ and $\cZ$, respectively, endowed with the Borel $\sigma$-algebras generated by the weak topologies.

\begin{assumption}[Measurability]\label{assump:reg-measurability}
The map $\psi\mapsto \dist^\psi \in \cP(\tilde\cZ)$ is measurable, in the sense that $\psi \mapsto \dist^\psi\lpar B \rpar$ is $\cB_\Psi$-measurable for each $B \in \cB_{\tilde{\cZ}}$. Similarly, $\psi\mapsto \dist_{\obs}^\psi \in \cP(\cZ)$ is $\cB_\Psi$-measurable and its image set $\{\dist_{\obs}^{\psi}:\psi\in\Psi\}$ is a Borel subset of $\cP(\cZ)$.
\end{assumption}

\xhdr{Prior and Posterior Distributions} 
Let $\pi$ be a prior on $(\Psi,\cB_\Psi)$. Define the joint law $\dist^\pi$ of $\lpar(\tilde Z_i)_{i\ge1},\psi\rpar$ by first sampling $\psi \sim \pi$ and then, conditional on $\psi$, sampling $(\tilde{Z}_i)_{i \geq 1}$ i.i.d.\ from $\dist^\psi$. We use $\dist^\pi_{\vect X}$ to denote its marginal distribution on $\vect X$. 

Let $\data^n_\obs \coloneqq \lpar Z_1, Z_2, \ldots, Z_n \rpar$ denote the first $n$ observed variables (the $\cZ$-marginals of the corresponding $\tilde Z_i$). We write $\pi\lpar\cdot \given \data_\obs^n\rpar$ for the posterior on $\Psi$ induced by $\dist^\pi$. 

\xhdr{Parametric CEPOs and CEPO Posterior Predictive} 
For each $t\in\cT$ and $\pi$-almost every $\psi$, regular conditional distributions for $(Y_t\mid \vect X)$ exist because all relevant spaces are standard Borel. Thus, there is a Borel version of the conditional expectation $\vect x\mapsto \E^{\dist^\psi}\lbar Y_t \given \vect X=\vect x\rbar$. We fix a version $\mu_t(\cdot\midsem\psi)$ that is jointly measurable in $(\vect X,\psi)$ and call it the conditional expected potential outcome (CEPO):
\begin{equation}\label{eq:cepo-appx}
    \mu_t\lpar \vect x \midsem \psi \rpar \; \coloneqq\;  \E^{\dist^\psi} \!\lbar\, Y_t \given \vect X = \vect x\, \rbar, \qquad \text{for }\dist_{\vect X}^\pi\text{-almost every } \vect x.
\end{equation}

\begin{assumption}[Integrability] \label{assump:reg-integrability}
For every $t \in \cT$ and $\dist^\pi_{\vect X}$-almost every $\vect x$:
\begin{equation}
    \E^\pi \lbar\, |\mu_{t}(\vect x \midsem \psi)|\, \rbar < \infty.
\end{equation}
\end{assumption}

For any query $(t,\vect x)$ and dataset $\data_\obs^n$, the CEPO posterior predictive distribution (CEPO-PPD), a probability measure on $\R$, is the pushforward of the posterior $\pi\lpar \psi \given \data_\obs^n\rpar$ through $\psi\mapsto \mu_t(\vect x;\psi)$:
\begin{equation}\label{eq:cepo-ppd-appx}
    \pi^{\mu_t} \lpar B \given \vect{x}, \data_\obs^n \rpar \coloneqq \int_\Psi \bbI\lpar \mu_t\lpar \vect x \midsem \psi \rpar \in B\rpar \pi \lpar \diff \psi \given \data_\obs^n\rpar, \qquad B\in \cB.
\end{equation}

\xhdr{Model}
Given a query $(t,\vect{x})$ and context $\data_{\obs}^n$, a model with parameters $\theta$ yields a predictive distribution $q_\theta\lpar \,\cdot\given \vect{x},t,\data_{\obs}^n\rpar$  on $\R$ for the CEPO values. 

\xhdr{Observational Quotient Space} Standard consistency results such as Doob's consistency theorem \citep{miller2018detailed} are concerned with the parameters of the \emph{observational} distribution. To leverage such results, we characterize the set of DGPs with the same observational distributions as follows:
\begin{definition}[Observational Quotient Space]
    \label{def:q-space}
    Let $\Phi \coloneqq \Psi / \sim$ be the set of equivalence classes under: $\psi_1 \sim \psi_2$ iff $\dist^{\psi_1}_\obs = \dist^{\psi_2}_\obs$. Let $[\cdot]: \Psi\to\Phi$ be the quotient map and equip $\Phi$ with the quotient $\sigma$-algebra $\cB_{\Phi}\coloneqq\{A\subseteq\Phi:  [\cdot]^{-1}(A)\in\cB_\Psi\}$. We call $(\Phi, \cB_\Phi)$ the observational quotient space.
\end{definition}
We write $\phi = [\psi]$ to denote the equivalence class corresponding to the parameter $\psi$ and may interchangeably use $\dist^{[\psi]}$, $\dist^\phi$, or $\dist^{\psi}_\obs$ to denote the corresponding observational distribution. Note $(\Phi,\cB_\Phi)$ is also standard Borel. Indeed, identify $\Phi$ with the image $R\coloneqq\{\dist_{\obs}^{\psi}:\psi\in\Psi\}\subseteq\cP(\cZ)$ via the measurable bijection $[\psi]\mapsto \dist_{\obs}^{\psi}$. By Assumption~\ref{assump:reg-measurability}, $R$ is a Borel subset of the standard Borel space $\cP(\cZ)$ (the space of probability measures on $\cZ$ with the weak topology is standard Borel whenever $(\cZ,\cB_\cZ)$ is), hence $R$ and thus $\Phi$ are standard Borel~\citep{srivastava1998course}.

\xhdr{Identifiability}
Finally, we re-state the definition of identifiability from \cref{sec:background} more formally. This will be a necessary condition to prove our results in the next sections.

\begin{repdefinition}{def:identification}[CEPO-Identifiability] We call a prior $\pi$ on $(\Psi,\cB_\Psi)$ CEPO-identifiable, if for each value of $t \in \cT$, there exists a map $f_t: \cX \times \Phi \to \R$, such that for $\pi$-almost all parameters $\psi$ and $\dist^\pi_{\vect X}$-almost all values of $\vect x$, the CEPO value $\mu_t(\vect x \midsem \psi) = f_t(\vect x, [\psi])$.
\end{repdefinition}
Note that the above definition is compatible with the standard identifiability in the literature~\citep{peters2017elements}, since there is a bijection between each equivalence class $[\psi]$ and the observational distribution $\dist_\obs^\psi$.
\section{Consistency Result} \label{appx:theory}

\subsection{Re-Statement of \texorpdfstring{\cref{prop:cepo-ppd-consistency}}{Proposition 1}}

\begin{repproposition}{prop:cepo-ppd-consistency}[Formal]
Under \cref{assump:reg-measurability,assump:reg-integrability}, there exist sets $\cX_0\subseteq\cX$ and $\Psi^\star\subseteq\Psi$ with  $\dist^\pi_{\vect X}\lpar \cX_0 \rpar = 1$  and $\pi\lpar \Psi^\star \rpar = 1$, such that for all $t \in \cT, \vect x \in \cX_0$, and $\psi^\star \in \Psi^\star$, if $Z_1, Z_2, \ldots \sim \dist^{\psi^\star}_\obs$ i.i.d., then 
\begin{equation} \label{eq:cepo-ppd-prop-eq}
  \lim_{n \to \infty} \E_{\mu\sim \pi^{\mu_{t}}\lpar \cdot \given \vect x, \data_\obs^n \rpar} \lbar \mu \rbar
  =
  \mu_{t}(\vect x\midsem\psi^\star)\quad \dist_\obs^{\psi^\star}\text{-a.s.},
\end{equation}
\textbf{if and only if} the prior $\pi$ is CEPO-identifiable.
\end{repproposition}

\subsection{Preliminaries for the Proof of \texorpdfstring{\cref{prop:cepo-ppd-consistency}}{Proposition 1}}
Here, we introduce some concepts to simplify the statement of the proof. We start by presenting a corollary of Doob's consistency theorem without proof (Corollary 2.3 of \texorpdfstring{\citet{miller2018detailed}}{}), which we will heavily leverage for the proof of \cref{prop:cepo-ppd-consistency}. The result is re-stated to match our parallel notation: 

\begin{theorem}[Corollary of Doob's Consistency Theorem]\label{thm:doob} Suppose $(\cZ, \cB_\cZ)$ and $(\Phi, \cB_\Phi)$ are two standard Borel spaces. Let $\nu$ be a probability measure on $(\Phi, \cB_\Phi)$. For each $\phi \in \Phi$, let $\dist^\phi$ be a probability measure on $\lpar \cZ, \cB_\cZ \rpar$. Consider a measurable map $g: \Phi \to \R$ and assume:
\begin{enumerate}
    \item[\textbf{(i)}] \textbf{Measurability.} $\phi \mapsto \dist^\phi \lpar B \rpar$ is measurable for every $B \in \cB_\cZ$.
    \item[\textbf{(ii)}] \textbf{No Redundancy.} $\phi \neq \phi' \implies  \dist^\phi \neq  \dist^{\phi'}$.
    \item[\textbf{(iii)}] \textbf{Integrability.} $\E^\nu\lbar |g({\phi})|\rbar < \infty$.
\end{enumerate}
Moreover, define the extended joint probability measure $\nu_{\text{tot}}$ on $\lpar(Z_1, Z_2, \ldots), \phi\rpar$ by first drawing $\phi \sim \nu$, and then, conditioned on $\phi$, sampling $Z_1, Z_2, \ldots$ i.i.d. from $\dist^\phi$. Then, there exists $\Phi_0 \subseteq \Phi$ with $\nu(\Phi_0)=1$, such that for any $\phi_0 \in \Phi_0$ and $Z_1, Z_2, \ldots \sim \dist^{\phi_0}$ i.i.d., we have 
\begin{equation}
    \lim_{n \to \infty} \E^{\nu_{\text{tot}}} \lbar g({\phi}) \given Z_1, \ldots, Z_n \rbar = g(\phi_0) \quad \dist^{\phi_0}\text{-a.s.}
\end{equation}
\end{theorem}
\xhdr{The Joint Measure $\Pi$} 
For technical convenience, we define a joint measure $\Pi$ on variables $\psi, ( \tilde Z_i )_{i \ge 1}, [\psi],$ and $( Z_i )_{i \ge 1}$ , as the pushforward measure of $\dist^\pi$ by the following map: 
\begin{equation} \label{eq:the-big-pushforward}
    \lpar \psi, ( \tilde Z_i )_{i \ge 1} \rpar \mapsto \lpar \psi, ( \tilde Z_i )_{i \ge 1}, [\psi], ( Z_i )_{i \ge 1} \rpar.
\end{equation}
In particular, we have the following equalities:
\begin{equation}
    \Pi \lpar ( Z_i )_{i \ge 1} \given \psi, [\psi] \rpar = \dist^\psi_\obs \lpar ( Z_i )_{i \ge 1} \rpar = \dist^{[\psi]} \lpar ( Z_i )_{i \ge 1} \rpar  = \Pi \lpar ( Z_i )_{i \ge 1} \given [\psi] \rpar,
\end{equation}
which results in the conditional independence
\begin{equation} \label{eq:cond-indep-PI}
    ( Z_i )_{i \ge 1} \indep_\Pi \psi \mid [\psi].
\end{equation}
Since all spaces involved are standard Borel, regular conditional distributions exist; hence the above conditional laws are well defined~\citep{preston2009note}.

\textit{(Notation Remark)}
We remove the superscript $\Pi$ in expectations and simply write $\E$ when we take expectations w.r.t.\ $\Pi$. Also, we reuse the symbol $\Pi$ for the joint measure and for any of its marginals or conditionals; the intended meaning will be clear from context.

\subsection{Proof of \texorpdfstring{\cref{prop:cepo-ppd-consistency}}{Proposition 1}}

For any given $t$ and $\vect{x}$, define the expected CEPOs under observational equivalence class $\phi \in \Phi$ as\footnote{Equivalently, $g_t(\vect x \midsem \phi)=\E\lbar \mu_t(\vect x \midsem \psi)\given  [\psi] = \phi \rbar$. } 
\begin{equation}\label{eq:obs-functional}
    g_t(\vect x \midsem \phi) \coloneqq \E\lbar \mu_t(\vect x \midsem \psi) \given \phi \rbar.
\end{equation}
We can use \cref{thm:doob} to establish a consistency result connecting the CEPO-PPDs and the functions $g_t$ defined in \cref{eq:obs-functional}:
\begin{lemma}\label{lemma:expected-cepo-cons}
Under \cref{assump:reg-measurability,assump:reg-integrability}, there exist sets $\cX_0\subseteq\cX$ and $\Psi_0\subseteq\Psi$ with  $\dist^\pi_{\vect X}\lpar \cX_0 \rpar = 1$  and $\pi\lpar \Psi_0 \rpar = 1$, such that for all $t \in \cT, \vect x \in \cX_0$, and $\psi_0 \in \Psi_0$, if $Z_1, Z_2, \ldots \sim \dist^{\psi_0}_\obs$ i.i.d., then 
\begin{equation}
  \lim_{n \to \infty} \E_{\mu \sim \pi^{\mu_{t}}\lpar \cdot \given \vect x , \data_\obs^n\rpar} \lbar\mu \rbar
  =
  g_{t}(\vect x \midsem [\psi_0])\quad \dist_\obs^{\psi_0}\text{-a.s.} \label{eq:lemma-expected-cepo-cons}
\end{equation}
where $\mu$ denotes the identity map on $\R$.
\end{lemma}
\begin{proof}    
    From \cref{assump:reg-integrability}, a subset $\cX_0 \subseteq \cX$ exists with $\dist^\pi_{\vect X}(\cX_0) = 1$, such that for all $t \in \cT$ and $\vect x \in \cX_0$, we have $\E^\pi \lbar |\mu_{t}(\vect x \midsem \psi) |\rbar < \infty$. Fix a value of $\vect x_0 \in \cX_0$ and $t_0 \in \cT$. A similar integrability statement can be made for $g_{t_0}(\vect x_0\midsem \phi)$:
    \begin{align}
        \E \lbar \abs*{g_{t_0}(\vect x_0\midsem \phi)} \rbar &= \E \lbar \abs{\E \lbar \mu_{t_0}(\vect x_0 \midsem \psi) \given \phi \rbar}\rbar& \text{from \cref{eq:obs-functional}}\nonumber \\
        & \leq \E \lbar \E \lbar \abs{\mu_{t_0}(\vect x_0 \midsem \psi)} \given \phi \rbar \rbar& \text{(Jensen's inequality)}\nonumber \\
        & = \E \lbar \abs{\mu_{t_0}(\vect x_0 \midsem \psi)}\rbar < \infty.& \text{(total expectation)}\label{eq:integ3}
    \end{align}
    
    Now, we use \cref{thm:doob} to obtain the desired results for the function $g_{t_0}(\vect x_0 \midsem \phi)$ by plugging in $(\cZ, \cB_\cZ)$ directly from our notation and $(\Phi, \cB_\Phi)$ from  \cref{def:q-space}. Moreover, we replace $\nu$ and $\nu_{\text{tot}}$ by the marginals of $\Pi$ on the random variables $\phi$ and $\lpar (Z_i)_{i \geq 1}, \phi \rpar$, respectively. Finally, it is easy to see that all the required assumptions hold:
    \begin{enumerate}
        \item[\stepindicator{i}] \textbf{Measurability.} Follows from the measurability in \cref{assump:reg-measurability}.
        \item[\stepindicator{ii}] \textbf{No Redundancy.} Follows from the definition of the quotient space in \cref{def:q-space}.
        \item[\stepindicator{iii}] \textbf{Integrability.} Follows from \cref{eq:integ3}.
    \end{enumerate}
    As a result of \cref{thm:doob}, there exists a set $\Phi_0 \subseteq \Phi$ with $\Pi(\Phi_0) = 1$, such that for any $\phi_0 \in \Phi_0$ and $Z_1, Z_2, \ldots \sim \dist^{\phi_0}$ i.i.d., we have
    \begin{equation} \label{eq:new-doob}
        \lim_{n \to \infty} \E \lbar g_{t_0}(\vect x_0 \midsem \phi) \given \data_\obs^n \rbar = g_{t_0}(\vect x_0 \midsem \phi_0)\quad \dist^{\phi_0}\text{-a.s.}
    \end{equation}
We can simplify the expectation in the L.H.S. of \cref{eq:new-doob} as follows:
\begin{align}
    \E \lbar g_{t_0}(\vect x_0\midsem \phi) \given \data_\obs^n \rbar =&~ \E \lbar \E \lbar \mu_{t_0}(\vect x_0\midsem \psi)  \given \phi \rbar \given \data^n_{\obs} \rbar& \text{from \cref{eq:obs-functional}}\nonumber \\
    =&~ \E \lbar\E \lbar \mu_{t_0}(\vect x_0\midsem \psi)  \given \data^n_{\obs}, \phi \rbar \given \data^n_{\obs} \rbar& \text{from \cref{eq:cond-indep-PI}} \nonumber\\
    =&~ \E \lbar \mu_{t_0}(\vect x_0\midsem \psi)  \given \data^n_{\obs} \rbar & \text{(tower property)} \nonumber \\
     \overset{(\star)}{=}&~ \E_{\mu \sim \pi^{\mu_{t_0}}\lpar \cdot \given \vect x_0, \data_\obs^n\rpar} \lbar \mu \rbar, \label{eq:final111}
\end{align}
where $(\star)$ follows from the fact that CEPO-PPD $\pi^{\mu_{t_0}}$ is the pushforward of the posterior $\Pi(\psi | \data^n_\obs) = \pi(\psi | \data^n_\obs)$ under the map $\psi \mapsto \mu_{t_0}(\vect x_0 \midsem \psi)$.

We then define $\Psi_0$ as the preimage of $\Phi_0$ under the quotient mapping. It is easy to verify that $\pi(\Psi_0) = \Pi(\Psi_0) = \Pi(\Phi_0) = 1$. For any $\psi_0 \in \Psi_0$, set $\phi_0 = [\psi_0]$. Combining \cref{eq:new-doob,eq:final111}, and repeating the entire argument for all $t_0 \in \cT$ and $\vect x_0 \in \cX_0$ concludes the proof.
\end{proof}

\cref{lemma:expected-cepo-cons} establishes a consistency result between the CEPO-PPDs and functions $g_t$ we defined on the quotient space. With the consistency proven in the observational quotient space, all that remains is to connect the R.H.S. of \cref{eq:lemma-expected-cepo-cons} to the original CEPOs. This is where identifiability comes into play. In what follows, fix $t_0 \in \cT$ and $\vect x_0 \in \cX_0$:

\xhdr{CEPO-Identifiability $\Rightarrow$ Consistency}  Under CEPO-identifiability (\cref{def:identification}), there exists $\Psi_1 \subseteq \Psi$ with $\pi(\Psi_1) = 1$, where for all $\psi', \psi'' \in \Psi_1$ that  $[\psi'] = [\psi'']$, we have $\mu_{t_0}(\vect x_0 \midsem \psi') = \mu_{t_0}(\vect x_0\midsem \psi'')$. Define $\Psi^\star \coloneqq \Psi_1 \cap \Psi_0$ and note that $\pi(\Psi^\star) = 1$. Consequently, for any $\psi^\star \in \Psi^\star$, we also have $\psi^\star \in \Psi_1$, and 
\begin{equation}\label{eq:iden-to-consistency}
    g_{t_0}(\vect x_0 \midsem [\psi^\star]) \overset{\cref{eq:obs-functional}}{=} \E\lbar \mu_{t_0}(\vect x_0 \midsem \psi) \given [\psi] = [\psi^\star]\rbar = \mu_{t_0}(\vect x_0 \midsem \psi^\star).
\end{equation} 
Combining \cref{eq:iden-to-consistency} with \cref{lemma:expected-cepo-cons} and repeating the argument for all $t_0 \in \cT$ and $\vect x_0 \in \cX_0$ proves the first side of \cref{prop:cepo-ppd-consistency}.

\xhdr{Consistency $\Rightarrow$ CEPO-Identifiability} When consistency holds, from \cref{eq:cepo-ppd-prop-eq}, a set $\Psi^\star \subseteq \Psi$ exists with $\pi(\Psi^\star) = 1$,  where for all $\psi^\star \in \Psi^\star$, if $Z_1, Z_2, \ldots \sim \dist^{\psi^\star}_\obs$, then
\begin{align}
    \lim_{n \to \infty} \E_{\mu \sim \pi^{\mu_{t_0}}\lpar \cdot \given \vect x_0, \data_\obs^n\rpar} \lbar \mu \rbar
  =
  \mu_{t_0}(\vect x_0\midsem\psi^\star)\quad \dist_\obs^{\psi^\star}\text{-a.s.}
\end{align}
Moreover, according to \cref{lemma:expected-cepo-cons}, there exists a set $\Psi_0 \subseteq \Psi$ with $\pi(\Psi_0) = 1$, such that for all $\psi_0 \in \Psi_0$, if $Z_1, Z_2, \ldots \sim \dist^{\psi_0}_\obs$, then
\begin{equation}
  \lim_{n \to \infty} \E_{\mu \sim \pi^{\mu_{t_0}}\lpar \cdot \given \vect x_0, \data_\obs^n\rpar} \lbar \mu \rbar
  =
  g_{t_0}(\vect x_0 \midsem [\psi_0])\quad \dist_\obs^{\psi_0}\text{-a.s.} \label{eq:lemma-expected-cepo-cons-2}
\end{equation}
Using these two identities, we can define $\Psi_1 \coloneqq \Psi^\star \cap \Psi_0$, where $\pi(\Psi_1) = 1$, and the following holds for every $\psi_1 \in \Psi_1$:
\begin{equation}\label{eq:as-equality}
    \mu_{t_0}(\vect x_0 \midsem \psi_1) = g_{t_0}(\vect x_0 \midsem [\psi_1]).
\end{equation}
Hence, the prior $\pi$ is indeed CEPO-identifiable, as we can use the functional $g$ in place of $f$ in \cref{def:identification}. Repeating this process for all $t_0 \in \cT$ and $\vect x_0 \in \cX_0$ concludes the proof of \cref{prop:cepo-ppd-consistency}.

\section{Validity of the Causal Data-Prior Loss} \label{appx:theory2}
Here, we show that the causal data-prior loss is equivalent to the expected forward KL divergence between the CEPO-PPDs and the parameterized distribution $q_\theta$. For the theoretical justification, we assume a fixed observation size $n$ and define $\data_\obs \coloneqq \data_\obs^n$ with a dropped superscript for simplicity.

\begin{assumption}[Existence of Densities]\label{assump:densities}
We assume each CEPO-PPD $\pi^{\mu_t} \lpar \cdot \given \vect{x}, \data_\obs^n \rpar$ admits a density w.r.t.\ Lebesgue measure and use the same symbol for its density. Moreover, we assume $q_\theta\lpar \,\cdot \given \vect x,t,\data_\obs^n\rpar$ is a probability measure with full support on $\R$, which admits a density w.r.t.\ Lebesgue measure. Similar to CEPO-PPDs, we use the same symbol for the measure and its density. 
\end{assumption}

\begin{definition}\label{def:kl-loss}
Let $\dist^\pi_\obs$ be the marginal distribution of $\dist^\pi$ on $(Z_i)_{i \geq 1}$. Then, the expected forward-KL divergence between $\pi^{\mu_t}$ and $q_\theta$ is defined as
\begin{equation}
   \cL_{t}^{\mathrm{KL}}(\theta)\coloneqq \E_{\data_\obs \cup \{\vect{x}\} \ \sim\ \dist^\pi_\obs} \lbar \kl{\pi^{\mu_t}\lpar \cdot \given \vect{x}, \data_\obs \rpar}{q_\theta\lpar \cdot \given \vect{x}, t, \data_\obs \rpar} \rbar, \label{eq:kl-loss}
\end{equation}
where $\data_\obs \cup \{\vect{x}\} \ \sim\ \dist^\pi_\obs$ refers to first drawing $\psi\sim \pi$, and
then sampling $\data_\obs=(Z_1,\ldots,Z_n)$ i.i.d. from $\dist^\psi_\obs$ and an independent query point
$\vect x\sim \dist^\psi_{\vect X}$.
\end{definition}

\begin{proposition}
Under \cref{assump:densities}, the causal data-prior loss from \cref{def:causal-data-prior-loss} and the expected forward-KL divergence in \cref{def:kl-loss} have the same optima. In other words, for all $t \in \cT$,
\begin{equation}\label{eq:equivalence-def}
    \arg \min_\theta \cL_{t}^{\mathrm{KL}}(\theta) = \arg \min_\theta \cL_{t}(\theta).
\end{equation}
\end{proposition}
\begin{proof}
Fix a $t \in \cT$. We note that
\begin{align}
     {\cL}^{\mathrm{KL}}_{t}(\theta) =  \E_{\data_\obs \cup \{\vect{x}\} \ \sim\ \dist^\pi_\obs} \lbar  \E_{\mu \sim \pi^{\mu_t} \lpar \cdot \given \vect{x}, \data_\obs\rpar} \lbar \log \frac{\pi^{\mu_t} \lpar \mu \given \vect{x}, \data_\obs\rpar}{q_\theta\lpar \mu \given \vect{x}, t, \data_\obs\rpar} \rbar \rbar. \label{eq:kl1}
\end{align}
From \cref{eq:cepo-ppd-appx}, we know that $\pi^{\mu_t}\lpar \cdot \given \vect{x},\data_\obs\rpar$ is the pushforward of the posterior $\pi\lpar \cdot \given\data_\obs\rpar$ by the function $\psi \mapsto  \mu_t(\vect{x}\midsem \psi)$. Hence, for any measurable function $h:\R \to \R$, we get
\begin{align}
    \E_{\mu \sim \pi^{\mu_t} \lpar \cdot \given \vect{x}, \data_\obs\rpar} \lbar h(\mu)\rbar = \E_{\psi \sim \pi \lpar \cdot \given \data_\obs\rpar} \lbar h(\mu_t(\vect{x}\midsem \psi)) \rbar.\label{eq:lous}
\end{align}
Setting $h(\mu)=\log \frac{\pi^{\mu_t} \lpar \mu \given \vect{x}, \data_\obs\rpar}{q_\theta\lpar \mu \given \vect{x}, t, \data_\obs\rpar}$ in \cref{eq:lous} and combining with \cref{eq:kl1} yields
\begin{align}
     {\cL}^{\mathrm{KL}}_{t}(\theta) &=  \E_{\data_\obs \cup \{\vect{x}\} \ \sim\ \dist^\pi_\obs} \lbar \E_{\psi \sim \pi\lpar \cdot \given \data_\obs\rpar} \lbar \log \frac{\pi^{\mu_t} \lpar\mu_t(\vect{x}\midsem \psi) \given \vect{x}, \data_\obs\rpar}{q_\theta\lpar \mu_t(\vect{x}\midsem \psi) \given \vect{x}, t, \data_\obs\rpar} \rbar \rbar \\
     &=  \E_{\data_\obs \cup \{\vect{x}\} \ \sim\ \dist^\pi_\obs,\; \psi \sim \pi\lpar \cdot \given \data_\obs\rpar} \lbar \log \frac{\pi^{\mu_t} \lpar \mu_t(\vect{x}\midsem \psi) \given \vect{x}, \data_\obs\rpar}{q_\theta\lpar \mu_t(\vect{x}\midsem \psi) \given \vect{x}, t, \data_\obs\rpar} \rbar. \label{eq:mid-kl}
\end{align}
Next, we use the Bayes' rule to derive
\begin{equation}
    \underbrace{\dist^\pi_\obs \lpar \data_\obs \rpar}_{\text{evidence}} \underbrace{\pi\lpar \psi \given \data_\obs\rpar}_{\text{posterior}} = \underbrace{\pi(\psi)}_{\text{prior}} \underbrace{\dist^\psi_\obs \lpar \data_\obs \rpar}_{\text{likelihood}}. \label{eq:bayesss}
\end{equation}
Combining \cref{eq:mid-kl} and \cref{eq:bayesss}, we get
\begin{align}
      {\cL}^{\mathrm{KL}}_{t}(\theta) &=  \E_{\psi \sim \pi,\ \data_\obs \cup \{\vect{x}\} \ \sim\ \dist^\psi_\obs} \lbar \log \frac{\pi^{\mu_t}\lpar \mu_t(\vect{x}\midsem \psi) \given \vect{x}, \data_\obs\rpar}{q_\theta\lpar \mu_t(\vect{x}\midsem \psi) \given \vect{x}, t, \data_\obs\rpar} \rbar \\
      & = \E_{\psi \sim \pi,\ \data_\obs \cup \{\vect{x}\} \ \sim\ \dist^\psi_\obs} \lbar - \log {q_\theta\lpar \mu_t(\vect{x}\midsem \psi) \given \vect{x}, t, \data_\obs\rpar} \rbar + \text{constant term in } \theta\\
      & =  {\cL}_{t}(\theta)+ \text{constant term in } \theta,\label{eq:equiv}
\end{align}
which concludes the proof.
\end{proof}

\textit{(Remark 1)} In general, the forward-KL divergence loss cannot be estimated without estimating the true CEPO-PPD. However, with this identity established, we can justify the use of the equivalent causal data-prior loss which is easily estimable. 

\textit{(Remark 2)} The theoretical equivalence is proved only for a fixed treatment $t \in \cT$ and a fixed, finite sample size $n \in \{1,2,\ldots\}$.  
In practice, the training loss is minimized while \emph{randomizing} both  
$t$ and the sample size $n$. If the optimizer attains a near-optima of this
randomized objective, the approximation $q_{\theta}\lpar \cdot \given \vect{x},t,\data_{\obs}\rpar \approx \pi^{\mu_{t}}\lpar \cdot \given \vect{x},\data_{\obs}\rpar$ can effectively extend to all the treatment values and to almost every sample size we care about in practice.
\section{Experimental Details}
\label{sec:appx-exp-details}
\subsection{Prior Generation \& Simulating DGPs} \label{appx:prior_generation}

As illustrated in \cref{fig:prior-sampling}, our prior generation consists of retrieving or synthesizing a base table, subsampling covariates $\vect{X}$ and CEPOs $\mu_0$ and $\mu_1$, synthesizing treatments $T$, potential outcomes $Y_t$, and finally, observed outcomes $Y$. We break down each of the components:

\xhdr{Data Sources for the Base Tables} 
We draw the base tables from two sources: \stepindicator{i} real‐world tables from OpenML, and \stepindicator{ii} fully synthetic data.

\begin{itemize}
    \item[\stepindicator{i}] We use the OpenML collections used in ~\citet{grinsztajn2022why}, AMLB~\citep{gijsbers2024amlb}, and TabZilla~\citep{mcelfresh2023neural}, all listed in \citet{ma2024tabdpt}.  
    To widen coverage, we also add tables from CTR23~\citep{fischer2023ctr} and CC18~\citep{bischl2021cc18}.  
    All OpenML IDs are in \href{https://github.com/vdblm/CausalPFN/blob/main/assets/openml_catalogue.csv}{this link}.\footnote{\url{https://github.com/vdblm/CausalPFN/blob/main/assets/openml_catalogue.csv}}  
    Data leakage is ruled out as none of the tables that share covariates or outcomes with our test sets (Lalonde, IHDP, ACIC, Criteo, Megafon, Hillstrom, Lenta, X5) are included in training. Moreover, the propensities are sampled purely synthetically, following the approach described below. 

    \item[\stepindicator{ii}] For additional diversity, we generate synthetic tables using the random neural networks used to train TabPFN v1, with the same hyperparameters described in \citet{hollmann2023tabpfn}. Inputs, from a standard Gaussian distribution, are fed into the network, and a subset of the outputs and hidden neurons are selected to construct the tabular data. Some columns are discretized at random to produce categorical and ordinal variables to reflect the structure of real-world tabular domains. While TabPFN v2~\citep{hollmann2025accurate} is a newer and stronger model, its training data is not publicly available, so we restrict ourselves to the v1 generator to ensure transparent evaluation and leakage control.
\end{itemize}

\xhdr{CEPOs with Heterogeneity Control} Once the base table is given, we randomly select two columns and name them $\mu_{\text{raw}, 0}$ and  $\mu_{\text{raw}, 1}$. However, in practice, we observe that directly using such columns for CEPOs can result in large variances (\emph{heterogeneity}) for CATE.  We therefore apply a light‑weight post‑processing inspired by \texttt{RealCause} \citep{neal2020realcause}. 

The post-processing requires a heterogeneity hyperparameter $\gamma$, which we sample uniformly from $[0, 1]$ during prior generation. Then, for $N$ units (rows) extracted from the base table, let ${\tau}_{\text{raw}}^{(n)}=\mu^{(n)}_{\text{raw}, 1}-\mu^{(n)}_{\text{raw}, 0}$ be the CATE for unit $n \in [N]$, and ${\lambda}_{\text{raw}}=\tfrac1N\sum_{n=1}^N{\tau}_{\text{raw}}^{(n)}$ the sample ATE.
We draw i.i.d. $\{\alpha^{(n)}\}_{n=1}^N\sim\mathrm{Unif}[0,1]$ and construct the final $\gamma$-augmented CEPOs as
\begin{align}
\mu_1^{(n)} &\coloneqq
        \lbar\alpha^{(n)}+(1-\alpha^{(n)})\gamma\rbar{\mu}_{\text{raw}, 1}^{(n)}
        +(1-\gamma)(1-\alpha^{(n)})({\mu}_{\text{raw}, 0}^{(n)}+{\lambda_{\text{raw}}}),\\
        \mu_0^{(n)} &\coloneqq
        \lbar(1-\alpha^{(n)})+\alpha^{(n)}\gamma\rbar{\mu}_{\text{raw},0}^{(n)}
        +(1-\gamma)\alpha^{(n)}({\mu}_{\text{raw},1}^{(n)}-{\lambda_{\text{raw}}}).
\end{align}
A simple algebraic check shows
\begin{equation}    
\tau^{(n)} \;\coloneqq\;
        \mu_1^{(n)}-\mu_0^{(n)}
        \;=\;
        \gamma\,\tau_{\text{raw}}^{(n)} + (1-\gamma){\lambda_{\text{raw}}},
\quad
\var[\tau \mid \vect{x}] = \gamma^{2}\var[{\tau}_{\text{raw}} \mid \vect{x}].
\end{equation}
Hence, while preserving the average treatment effect, $\gamma=0$ yields a dataset with a zero variance CATE (fully homogeneous), whereas $\gamma=1$ recovers the original heterogeneity.

\xhdr{Outcomes} 
After constructing the CEPO columns $\mu_0(\vect{x})$ and $\mu_1(\vect{x})$, we need to turn them into potential outcomes by adding zero‑mean noises. To avoid tying the data to a specific parametric noise model, we introduce two additional \emph{nuisance} columns, $\eta_0(\vect x)$ and $\eta_1(\vect x)$, sampled from the base table. Let $\epsilon_t$ be random scalars, independent from $\vect x$, with $\E[\epsilon_t]=0$.
We define the potential outcomes as
\begin{equation}
Y_t \;=\; \mu_t(\vect{x}) \;+\; \eta_t(\vect{x})\,\epsilon_t, 
\quad t\in\cT.
\end{equation}
This construction preserves the conditional means, that is
$\E[Y_t \mid \vect{x}] = \mu_t(\vect{x})$. The input‑dependent scale factors $\eta_t(\vect{x})$ allow for heteroscedastic noises and capture a richer family of outcome distributions than additive parametric noise models. For our training, we sample  $\epsilon_t$ from a Gaussian with a variance uniformly drawn from $(0, \var(\mu_t)]$. This choice of noise values ensures a similar noise scale to the scale of CEPOs, resulting in training data with a more informative signal-to-noise ratio.

\xhdr{Propensities with Positivity Control} Given a covariate vector $\vect{x}$, the strong ignorability assumption requires the propensity values \(0 < \dist(T = 1 \mid \vect X = \vect x) < 1\). Hence, due to the invertibility of the {sigmoid} function, it is sufficient to generate treatment logits, through any function $f: \vect X \to \R$, and then apply a {sigmoid} function to get values within $(0, 1)$. To simulate different degrees of confounding, we choose $f$ by randomly selecting one of the following mechanisms:
\begin{itemize}
    \item[\stepindicator{i}] \textbf{Randomized treatments (RCT).}  
    Treatments are independent of covariates, i.e., $f$ is constant. We sample $c \sim \texttt{Logistic}(0, 1)$ and set ${f}(\vect{x}) = c$ to get uniform propensities. 
    
    \item[\stepindicator{ii}] \textbf{Linear logits.}  
    Draw the random vector $\vect{w}$ from a standard Gaussian and set  
    ${f}(\vect{x})=\vect{w}^{\top}\vect{x}$.
    
    \item[\stepindicator{iii}] \textbf{Non‑linear logits.}  
    Feed $\vect{x}$ into a randomly initialized MLP, similar architecture to that of \citet{hollmann2023tabpfn}, to get $f(\vect x)$. 
\end{itemize}
Empirically, we observe that the above procedure yields an artificially high level of positivity, which is not reflective of real-world scenarios. We therefore apply a light‑weight post‑processing transform, inspired by \texttt{RealCause} \citep{neal2020realcause}, to better control the positivity level. Concretely, we sample a parameter $\xi \in [0, 1]$ and \emph{exacerbate} extreme propensity scores to mimic poor positivity:
\begin{equation}
\dist(T=1 \mid \vect X = \vect x)
\;\coloneqq\;
\xi\,\sigmoid\lpar f(\vect{x})\rpar
\;+\;
(1-\xi)\,
\mathbb{I} \lpar f(\vect{x})> 0\rpar.
\end{equation}
Here, $\xi=1$ leaves the original positivity intact. However, for smaller
$\xi$ values, the support of the treated and control groups become increasingly
disjoint, leading to low‑positivity scenarios.

\xhdr{Treatment Assignment} Finally, each unit’s treatment is drawn as $T\sim\bernoulli\lpar\sigmoid \lpar f(\vect{X})\rpar\rpar$, and the observed outcome $Y$ is also derived by selecting the assigned potential outcome $Y \coloneqq Y_T$.

Collectively, all of the steps above simulate different DGPs, with various levels of positivity and heterogeneity, extracted from real and synthetic sources of tabular data. This procedure creates a broad prior $\pi$ for \method, which is necessary for the model to work well in practice.
\subsection{Model Details} \label{appx:arch}
\xhdr{Architecture \& Training} 
We represent each context row $(t, \vect x, y)$ and query row $(t, \vect x)$ as single tokens by summing up (1) a treatment embedding for $t$, (2) a covariate embedding for $\vect{x}$ (padded to length $F = 100$), and (3) an outcome embedding for $y$ (only for context rows). We use linear layers for embeddings and omit the positional encodings to preserve the permutation invariance of the context set, similar to other PFN-style transformers. 

All tokens—context and query—are passed into a 20‐layer transformer, with a hidden size of 384,  QK-normalization (RMS)\footnote{Different from \citet{henry2020query}, we perform normalization \emph{after} the query and key projection.}, and a parallel SwiGLU-activated~\citep{shazeer2020glu} feed-forward block.

The transformer’s query outputs are then projected to a 1024‐dimensional logit vector, then softmaxed at a fixed temperature of $\theta_T = 1.0$ to form a discrete CEPO posterior over the interval $[-10, 10]$. We then scale the interval to match the scale of the outcomes and clip the out-of-range values. At inference time, we return the posterior mean as the point estimate and sample 10,000 times to estimate credible intervals at any desired significance level $\alpha$.

The full model has approximately 20M parameters and is trained in two stages: \stepindicator{i} a predictive phase that mimics standard predictive PFN training from \citet{ma2024tabdpt}, and \stepindicator{ii} a causal phase that optimizes the CEPO loss. We use AdamW~\citep{kingma2014adam} with warmup and cosine annealing for the predictive phase, and switch to the schedule-free optimizer~\citep{defazio2024road} in the causal phase. The model is trained with a maximum context length of 16K in the first phase and 2,048 in the second. We use four A100 GPUs trained for at most one week for the initial phase, and two days on an H100 for the second phase.  

Finally, to enhance parallel training, we batch both the queries and the tables. That is, rather than sampling only one DGP and one query token, each gradient update samples $B_t$ DGPs, draws $B_q$ queries per DGP, and concatenates everything into a single tensor. The tensor is then passed through the transformer to get $B_t B_q$ CEPO-PPDs. The final loss is averaged over all the batches. See \cref{alg:training} for a detailed demonstration of \method's training algorithm.  
\begin{algorithm}[H]
  \caption{Parallel training of \method.}
  \label{alg:training}
  {
  \footnotesize
  \begin{algorithmic}[1]
    \Require Prior $\pi$, DGPs and CEPO values $\dist^\psi_\obs$, $\mu_t(\cdot\!\midsem\psi)$, model $q_\theta$, DGP batch size $B_t$, query batch size $B_q$, fixed feature length $F$, and histogram loss $\texttt{HL}$ \cref{eq:hl-loss}~\citep{imani2018improving}.
    \While{not converged}
      \State Sample $\psi[1], \hdots, \psi[B_t] \sim \pi$
      \State Sample $\data_\obs[i] \sim \dist^{\psi[i]}_\obs, \forall 1 \le i \le B_t$
      \State Randomly sample query treatments $t^{(i, j)}$ for ${1 \le i \le B_t, 1 \le j \le B_q}$
      \State Sample query covariates $\vect{x}^{(i, j)} \sim \dist^\psi_\obs[i]$ for ${1 \le i \le B_t, 1 \le j \le B_q}$
      \State Set ${\mu}^{(i, j)} \gets \mu_{t^{(i, j)}}\lpar \vect{x}^{(i, j)}\midsem \psi[i] \rpar$
      \State Pad $\vect{x}^{(i, j)}$ with zeros such that $\vect{x}^{(i, j)} \in \mathbb{R}^F$ 
      \State $\widehat{\cL} \gets \frac{1}{B_t\cdot B_q} \sum_{i,j}
             \texttt{HL}\left[{\mu}^{(i, j)} \Vert q_\theta( \cdot \mid \vect{x}^{(i, j)},t^{(i, j)},\data_\obs[i])\right]$
      \State Update $\theta$ using the gradients $\nabla_\theta\widehat{\cL}$
    \EndWhile
  \end{algorithmic}
  }
\end{algorithm}

\xhdr{Handling Large Tables at Inference Time}  
\method's default maximum context length is set to 4,096 at inference, but real‑world tables may contain millions of rows. Training PFN‑style transformers on such long contexts can be challenging due to hardware or architectural constraints.  While some tabular foundation models such as TabICL~\citep{qu2025tabicl} modify the architecture itself, \citet{thomas2024retrieval} show that, retrieving a small relevant subset of rows for each query at inference time allows a model with a short context length to better generalize to longer contexts.

We adopt this retrieval philosophy in \method\ to enable causal effect estimation on large tables. First, we fit a lightweight gradient boosting regressor on the context data to produce weak CATE estimates for each covariate. This regressor estimates CATE by regressing outcomes on the treatment and covariates and then taking the difference in predicted outcomes between $T=1$ and $T=0$. This step is applied \emph{only} when the table is too large to fit within the model's maximum context window. We then (i) sort both the context rows and the queries based on their weak CATE estimates, which effectively stratifies the data; (ii) partition the ordered queries into consecutive mini‑batches; and (iii) for each query batch, use a fast bisection search to select a contiguous window of context rows whose weak CATE estimate range most closely matches that of the batch. As a result, each batch is exposed only to a neighborhood of rows with similar causal effects, allowing all CEPO predictions to be computed with short forward passes.
\subsection{Sensitivity to Dataset Size \href{https://github.com/vdblm/CausalPFN/blob/main/notebooks/ablation_dataset_size.ipynb}{
    \includegraphics[height=1.4em]{assets/jupyter_logo.pdf}
}}\label{appx:sensitivity}

During the causal phase of training, we consider sample sizes up to 2{,}048 and covariates up to 100. However, during inference, \method\ can take up to 50{,}000 samples. 

To assess the effect of context size and dimensionality on \method’s performance, we run additional experiments on synthetic polynomial datasets. The test set size is fixed at 100 in all experiments. For each \emph{(rows, covariates)} configuration, we report mean~$\pm$~standard error over 50 datasets drawn from the polynomial prior with different random seeds.

\xhdr{Effect of Sample Size}
We consider the same DGPs while increasing the number of samples and fixing the number of covariates to 10. Table~\ref{tab:sample-size} reports PEHE for CATE across baselines. \method\ exhibits faster PEHE decay with increasing rows, with a slight plateau at very large contexts.

\begin{table}[H]
\centering
\caption{Effect of sample size on PEHE (mean~$\pm$~SE). Covariates = 10; averages over 50 datasets.}
\label{tab:sample-size}
\begin{adjustbox}{max width=\textwidth}
\begin{tabular}{lcccccccc}
\toprule
\textbf{Method} & \multicolumn{8}{c}{\textbf{Number of Rows}} \\
\cmidrule(lr){2-9}
& 10 & 20 & 50 & 100 & 200 & 500 & 5{,}000 & 10{,}000 \\
\midrule
\method & 1.34$\pm$0.02 & 1.27$\pm$0.02 & 1.10$\pm$0.02 & 0.89$\pm$0.02 & 0.74$\pm$0.03 & 0.46$\pm$0.01 &
0.29$\pm$0.01 & 0.31$\pm$0.01 \\
DA-Learner & 1.33$\pm$0.02 & 1.30$\pm$0.02 & 1.16$\pm$0.01 & 1.00$\pm$0.01 & 0.91$\pm$0.03 & 0.85$\pm$0.01 &
0.84$\pm$0.02 & 0.82$\pm$0.02 \\
S-Learner & 1.44$\pm$0.01 & 1.40$\pm$0.02 & 1.35$\pm$0.02 & 1.21$\pm$0.02 & 1.18$\pm$0.05 & 1.07$\pm$0.03 & 
1.00$\pm$0.04 & 1.03$\pm$0.04 \\
T-Learner & 1.33$\pm$0.02 & 1.30$\pm$0.02 & 1.15$\pm$0.01 & 0.97$\pm$0.01 & 0.88$\pm$0.02 & 0.81$\pm$0.01 &
0.81$\pm$0.02 &
0.81$\pm$0.02 \\
X-Learner & 1.35$\pm$0.02 & 1.32$\pm$0.02 & 1.20$\pm$0.02 & 1.04$\pm$0.01 & 0.94$\pm$0.03 & 0.87$\pm$0.01 & 
0.87$\pm$0.02 & 0.84$\pm$0.02 \\
\bottomrule
\end{tabular}
\end{adjustbox}
\end{table}

\xhdr{Effect of Covariate Size}
Next, we fix the number of samples to 1{,}000 and vary the number of covariates. Table~\ref{tab:covariate-size} compares \method’s performance to other methods in terms of PEHE. 
Although \method\ consistently outperforms the baselines, the performance gap narrows as the number of covariates grows—likely due to training exposure being limited to up to 100 dimensions, which could be mitigated by training with higher-dimensional inputs.

\begin{table}[H]
\centering
\caption{Effect of covariate size on PEHE (mean~$\pm$~SE). Samples = 1{,}000; averages over 50 datasets.}
\label{tab:covariate-size}
\begin{adjustbox}{max width=\textwidth}
\begin{tabular}{lcccccccc}
\toprule
\textbf{Method} & \multicolumn{8}{c}{\textbf{Number of Covariates}} \\
\cmidrule(lr){2-9}
& 1 & 5 & 10 & 20 & 50 & 100 & 500 & 1{,}000 \\
\midrule
\method & 0.08$\pm$0.00 & 0.17$\pm$0.01 & 0.40$\pm$0.01 & 0.67$\pm$0.01 & 0.87$\pm$0.02 & 1.01$\pm$0.02 &
1.28$\pm$0.02 & 1.32$\pm$0.02 \\
DA-Learner & 0.21$\pm$0.01 & 0.59$\pm$0.01 & 0.85$\pm$0.01 & 1.04$\pm$0.01 & 1.14$\pm$0.01 & 1.22$\pm$0.01 &
1.30$\pm$0.02 &
1.32$\pm$0.02 \\
S-Learner & 0.54$\pm$0.04 & 0.83$\pm$0.02 & 1.09$\pm$0.02 & 1.17$\pm$0.02 & 1.21$\pm$0.02 & 1.23$\pm$0.01 & 
1.30$\pm$0.02 & 1.33$\pm$0.02 \\
T-Learner & 0.27$\pm$0.01 & 0.60$\pm$0.01 & 0.83$\pm$0.01 & 0.98$\pm$0.01 & 1.09$\pm$0.01 & 1.18$\pm$0.01 & 
1.28$\pm$0.02 &
1.32$\pm$0.02 \\
X-Learner & 0.29$\pm$0.01 & 0.64$\pm$0.01 & 0.88$\pm$0.01 & 1.06$\pm$0.01 & 1.15$\pm$0.01 & 1.23$\pm$0.01 &
1.30$\pm$0.02 & 1.33$\pm$0.02 \\
\bottomrule
\end{tabular}
\end{adjustbox}
\end{table}
\subsection{Discussion on Inference Speed}\label{appx:speed}

Many applied settings prioritize throughput and latency over marginal gains in asymptotic accuracy. Real-time bidding must estimate incremental ad effects and decide bids within strict millisecond budgets~\cite{yuan2013real}. Likewise, e-commerce personalization depends on rapid uplift estimation within short user sessions, where serving latency directly affects conversion~\cite{sawant2018contextual}.

Although \method\ requires substantial offline training, it is designed for zero-shot deployment on new tables with no test-time fitting or adaptation. At inference, interventional queries reduce to a small and fixed number of forward passes, and the computation parallelizes well across large batches (e.g., using mixed precision, caching).

Accordingly, \cref{fig:teaser} does not claim that \method\ is intrinsically faster than every baseline; rather, it reflects practitioner-facing wall-clock time from data arrival to effects returned. Baselines that require per-dataset refitting or tuning incur this cost at deployment, whereas \method\ does not.
\subsection{Baseline Hyperparameters and Results without Hyperparameter Tuning \href{https://github.com/vdblm/CausalPFN/blob/main/notebooks/causal_effect_full.ipynb}{
    \includegraphics[height=1.4em]{assets/jupyter_logo.pdf}
}} \label{appx:hyper-setup}

\xhdr{No Hyperparameter Tuning} \Cref{tab:cate-no-hyper} summarizes the performance of all methods without hyperparameter tuning. \method\ attains the best (second-best) average rank on CATE (ATE).

\xhdr{EconML Hyperparameters} For the results without hyperparameter tuning in \cref{tab:cate-no-hyper}, we ran the models with the recommended hyperparameters in the Jupyter notebooks from EconML \citep{econml}. For the tuned results in \cref{tab:causal_effect_results}, we performed hyperparameter tuning using the FLAML (AutoML) library~\citep{Wang_FLAML_A_Fast_2021} on both the propensity and outcome models with \stepindicator{i} Time budget of $900$ seconds, \stepindicator{ii} K-fold cross-validation with $K = 3$, \stepindicator{iii} Early stopping, and \stepindicator{iv} base estimators \texttt{["lgbm", "xgboost", "xgb\_limitdepth", "rf", "kneighbor", "extra\_tree", "lrl1", "lrl2"]}.
For Forest DR-Learner and Forest DML, we additionally expanded the covariates with cubic terms (polynomial degree~3), with an additional tuning of the final model.

\xhdr{CATE Nets} For the results without hyperparameter tuning in \cref{tab:cate-no-hyper}, we ran the models with the default hyperparameters and a batch size of $512$. For the tuned results in \cref{tab:causal_effect_results}, we perform a grid search on the hyperparameters for the neural architecture: \stepindicator{i} Number of layers $\in$ $\{2, 3\}$, \stepindicator{ii} Representation dimension $\in$ $\{128, 256\}$, \stepindicator{iii} Number of hidden output layers $\in$ $\{1, 2\}$, and \stepindicator{iv} Width of the hidden output layers $\in$ $\{128, 256\}$. The rest of the hyperaparameters are left unchanged.

\xhdr{BART \& GRF} 
The GRF implementation includes an internal \texttt{tune} option.  
We enable this option in \cref{tab:causal_effect_results} and disable it for the untuned experiment in \cref{tab:cate-no-hyper}. BART, on the other hand, offers no comparable hyperparameter-tuning.  
Its only alternative, a full cross-fit, is prohibitively slow and uses a rudimentary Bayesian routine.
Thus, the BART scores appear unchanged in \cref{tab:causal_effect_results,tab:cate-no-hyper}.

\begin{table}[H]
\centering
\captionsetup{font=footnotesize}
\caption{\textbf{CATE \& ATE results.} PEHE \textit{(left half)} alongside ATE relative error and its overall average \textit{(right half)}. PEHE for Lalonde {\tiny CPS/PSID} is shown in thousands.  Best numbers are in \firstbest{blue}; second best are in \secondbest{purple}. Cells with ``—'' indicate that the method is not applicable.}
\label{tab:cate-no-hyper}
\begin{adjustbox}{max width=\textwidth}
\LARGE
\begin{tabular}{lccccc|ccccc}
\toprule
\multirow{2}{*}{\textbf{Method}} &
\multicolumn{5}{c|}{\textbf{Mean PEHE $\pm$ Standard Error} $(\downarrow \text{better})$} &
\multicolumn{5}{c}{\textbf{Mean ATE Relative Error $\pm$ Standard Error} $(\downarrow \text{better})$} \\
\cmidrule(r){2-6}\cmidrule(l){7-11}
& IHDP & ACIC 2016 & \multicolumn{1}{c}{Lalonde {\footnotesize CPS}} & \multicolumn{1}{c}{Lalonde {\footnotesize PSID}} & \multicolumn{1}{c|}{Avg.} & 
IHDP & ACIC 2016 & Lalonde {\footnotesize CPS} & Lalonde {\footnotesize PSID} & \multicolumn{1}{c}{Avg.} \\
& & & ($\times10^{3}$) & ($\times10^{3}$) & Rank & & & & & Rank\\
\midrule
\textbf{CausalPFN} &
\firstbest{0.58$\pm$0.07}&
\secondbest{0.92$\pm$0.11}&
\firstbest{8.96$\pm$0.02}&
{14.40$\pm$0.20}&
\firstbest{2.17$\pm$0.09}& 
0.20$\pm$0.04& 
\secondbest{0.05$\pm$0.01}& 
\firstbest{0.13$\pm$0.01}& 
0.22$\pm$0.02& 
\secondbest{4.26$\pm$0.18}\\

DA-Learner&
2.98$\pm$0.51&
1.88$\pm$0.24& 
\secondbest{9.01$\pm$0.02}& 
\secondbest{13.96$\pm$0.19}& 
\secondbest{3.64$\pm$0.18}&
0.22$\pm$0.04& 
0.09$\pm$0.03& 
\secondbest{0.22$\pm$0.01}& 
\secondbest{0.08$\pm$0.01}& 
\firstbest{4.15$\pm$0.19}\\

T-Learner&
2.94$\pm$0.49&
2.06$\pm$0.20& 
9.29$\pm$0.02& 
\firstbest{13.91$\pm$0.18}& 
4.01$\pm$0.18&
0.22$\pm$0.04& 
0.11$\pm$0.03& 
0.40$\pm$0.01& 
\firstbest{0.07$\pm$0.01}& 
4.62$\pm$0.18\\

DragonNet &
2.13$\pm$0.24& 
2.23$\pm$0.20& 
10.83$\pm$0.15& 
16.40$\pm$0.27& 
5.62$\pm$0.17& 
0.21$\pm$0.04& 
0.09$\pm$0.02& 
0.56$\pm$0.03& 
0.44$\pm$0.02& 
6.04$\pm$0.17 \\

IPW &
—& 
—& 
—& 
—& 
—&  
0.23$\pm$0.04 & 
0.24$\pm$0.05 & 
\secondbest{0.22$\pm$0.01}& 
\firstbest{0.07$\pm$0.01}& 
4.33$\pm$0.20\\

TarNet &
\secondbest{1.89$\pm$0.15}& 
2.26$\pm$0.20& 
12.00$\pm$0.04& 
18.71$\pm$0.16& 
6.87$\pm$0.11& 
0.21$\pm$0.04& 
0.06$\pm$0.02& 
0.90$\pm$0.01& 
0.72$\pm$0.01& 
7.54$\pm$0.14\\

X-Learner&
3.70$\pm$0.62& 
1.71$\pm$0.31& 
12.28$\pm$0.03&
21.72$\pm$0.16& 
8.13$\pm$0.16&
0.19$\pm$0.03& 
0.07$\pm$0.02&
0.83$\pm$0.01&
0.92$\pm$0.01& 
7.92$\pm$0.17\\

RA-Net &
2.08$\pm$0.19 &
2.42$\pm$0.22 & 
12.86$\pm$0.12 & 
20.13$\pm$0.41 & 
8.18$\pm$0.16& 
0.20$\pm$0.04 & 
0.07$\pm$0.03 & 
0.96$\pm$0.02 & 
0.71$\pm$0.04 & 
7.95$\pm$0.17 \\

BART &
2.50$\pm$0.39& 
\firstbest{0.68$\pm$0.11}& 
12.81$\pm$0.05& 
21.36$\pm$0.16& 
8.20$\pm$0.17& 
0.44$\pm$0.09& 
\firstbest{0.04$\pm$0.01}& 
0.99$\pm$0.01& 
0.86$\pm$0.01& 
8.72$\pm$0.18 \\

GRF &
4.26$\pm$0.69& 
{1.36$\pm$0.30}& 
12.18$\pm$0.06& 
21.84$\pm$0.16& 
8.21$\pm$0.17& 
\secondbest{0.18$\pm$0.03}& 
0.07$\pm$0.02& 
0.81$\pm$0.02& 
0.85$\pm$0.02& 
7.78$\pm$0.17\\

S-Learner &
3.91$\pm$0.68 &
2.23$\pm$0.28 & 
12.88$\pm$0.02 & 
22.68$\pm$0.13 & 
9.29$\pm$0.18&
0.28$\pm$0.05 & 
0.12$\pm$0.05 & 
1.00$\pm$0.00 & 
1.03$\pm$0.00 & 
9.99$\pm$0.18 \\

Forest DR Learner&
3.90$\pm$0.66&
1.68$\pm$0.35& 
26.08$\pm$4.96& 
22.55$\pm$0.25& 
9.51$\pm$0.18& 
0.19$\pm$0.04&
0.08$\pm$0.04&
1.39$\pm$0.28& 
0.87$\pm$0.03& 
8.35$\pm$0.17\\

Forest DML&
4.40$\pm$0.72& 
1.47$\pm$0.32& 
15.12$\pm$0.15& 
23.12$\pm$0.15& 
10.51$\pm$0.18& 
\firstbest{0.09$\pm$0.02}&
\secondbest{0.05$\pm$0.02}& 
1.12$\pm$0.02& 
1.02$\pm$0.01& 
9.37$\pm$0.23\\

\bottomrule
\end{tabular}
\end{adjustbox}
\end{table}

All inference, including baselines, performed on an 80 GB H100 GPU, 32 CPUs, and 256 GB RAM.
\subsection{Marketing Experiments \href{https://github.com/vdblm/CausalPFN/blob/main/notebooks/qini.ipynb}{
    \includegraphics[height=1.4em]{assets/jupyter_logo.pdf}
}} \label{appx:qini}

\xhdr{Datasets} Apart from Hill$^{(1)}$ and Hill$^{(2)}$, which were explained in the main text. We also run experiments on the following datasets:
\begin{enumerate}
    \item \textbf{Criteo.} 25M ad‑exposure records from Criteo’s online \emph{incrementality tests}: a randomly selected \emph{held‑out} audience is shielded from seeing an advert, while the treated audience is shown the ad; the target is a post‑impression conversion flag. We use a readily provided 2.5M stratified subset of this dataset from \texttt{sklift}.
    \item \textbf{Retail‑Hero (X5).} Transaction logs from the X5 Retail Group hackathon.  Customers are randomly offered personalized coupons (treatment); the outcome records whether the customer subsequently purchased the promoted items. 
    \item \textbf{Lenta.} SMS‑based promotion experiment run by the grocery chain Lenta.  The treatment group receives a marketing text, and the outcome is a visit after the campaign window. 
    \item \textbf{Megafon (Mega).} Synthetic yet domain‑faithful data released for the MegaFon Uplift competition.  Users are randomly offered a telecom upsell offer (treatment), and the outcome indicates whether they accepted the offer.
\end{enumerate}

\xhdr{Qini Evaluation} To build Qini curves we follow \texttt{scikit‑uplift}’s recommended five‑fold \emph{stratified} split based on the outcome and the treatment \citep{user-guide-for-uplift-modeling}.  In each fold, we hold out 20\% of the data as test rows and train the baseline models on the remaining 80\%.  For \method\ we use that same 80\% as context tokens and treat the held‑out 20\% as queries.  We then rank the rows based on their CATE estimates to compute the Qini curves and the corresponding Qini scores.

\xhdr{Context Length Challenges} In all the marketing experiments, we have increased the model’s maximum context length from the default 4,096 to 50,000 tokens. This context length is sufficient for the subsampled datasets in \cref{tab:uplift}. However, extending beyond 50K for the \emph{full‑table} runs is not feasible in GPU memory. We thus use the retrieval approach explained in \cref{appx:arch} to achieve CATE estimates for this setting. \cref{tab:uplift-full} shows \method's performance (with the retrieval approach) compared to the baselines on the full-table datasets. We conjecture that the relative under‑performance compared to \cref{tab:uplift} is due to this retrieval heuristic. 
\begin{table}[H]
    \centering
    \captionof{table}{\footnotesize \textbf{Normalized Qini scores} ($\uparrow$ better). Scores are normalized per dataset such that the top-performing model achieves 1.0 (highlighted in \firstbest{bold}). All datasets use full stratified subsamples: Hill$^{(1)}$ and Hill$^{(2)}$ (64K rows), Criteo (2.5M rows), X5 (200K rows), Lenta (687K rows), and Mega (600K rows).}
    \label{tab:uplift-full}
    \footnotesize
    \setlength{\tabcolsep}{4pt}
      \begin{tabular}{lccccccc}
        \toprule
        \textbf{Method} &
        \textbf{Hill$^{(1)}$} &
        \textbf{Hill$^{(2)}$} &
        \textbf{Criteo} &
        \textbf{X5} &
        \textbf{Lenta} &
        \textbf{Mega} &
        \textbf{Avg.} \\
        \midrule
        
        S Learner  
        & \firstbest{1.000} 
        & \firstbest{1.000} 
        & \firstbest{1.000} 
        & \firstbest{1.000} 
        & \firstbest{1.000} 
        & 0.913
        & \firstbest{0.985}\\
        
        X~Learner         
        & 0.975 
        & 0.980 
        & 0.994
        & 0.965
        & 0.868 
        & 0.997 
        & 0.963\\
        
        DA~Learner        
        & 0.985
        & 0.964 
        & 0.955 
        & 0.969 
        & 0.903 
        & \firstbest{1.000} 
        & 0.963\\
        
        T~Learner         
        & 0.991 
        & 0.972 
        & 0.902 
        & 0.953 
        & 0.833 
        & 0.987 
        & 0.940\\
        
        \method        
        & 0.992
        & 0.968 
        & 0.939 
        & 0.746
        & 0.947 
        & 0.954 
        & 0.924\\
        \bottomrule
      \end{tabular}
\end{table}
\subsection{Calibration, Coverage, and Credible Intervals} \label{appx:uncertainty}

\begin{figure}[t]
    \centering
    \includegraphics[width=\textwidth]{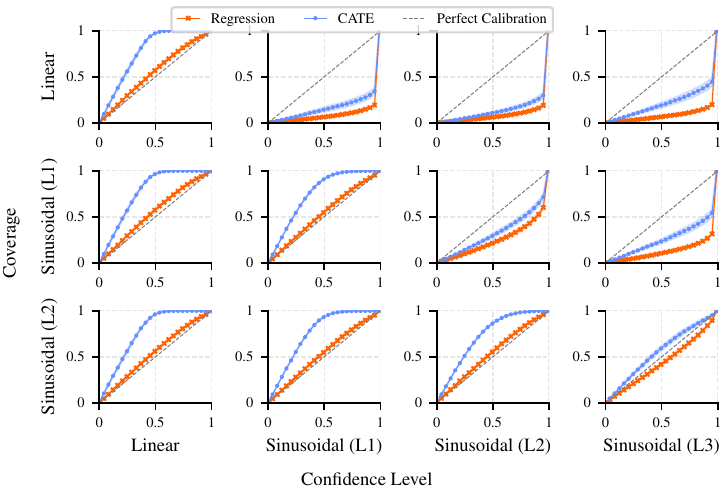}
    \caption{CATE and regression calibration curves for synthetic sinusoidal datasets, {\bf before calibration.} Models are trained on Linear/Sinusoidal (L1)/Sinusoidal (L2) datasets and tested on Linear/Sinusoidal (L1)/Sinusoidal (L2)/Sinusoidal (L3) benchmarks.}
    \label{fig:ablation-sinusoid}
\end{figure}

\begin{figure}[t]
    \centering
    \includegraphics[width=\textwidth]{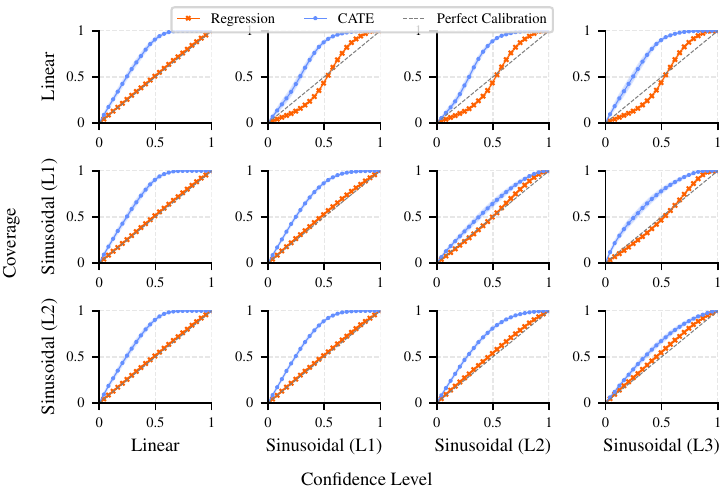}
    \caption{CATE and regression calibration curves for synthetic sinusoidal datasets, {\bf after calibration.} Models are trained on Linear/Sinusoidal (L1)/Sinusoidal (L2) datasets and tested on Linear/Sinusoidal (L1)/Sinusoidal (L2)/Sinusoidal (L3) benchmarks.}
    \label{fig:ablation-sinusoid-calibrated}
\end{figure}

\xhdr{The Synthetic DGPs} For the calibration results in \cref{fig:calibration}, we use two families of synthetic DGPs, polynomials and sinusoidals. As a general recipe, each DGP defines a treatment logit function $f(\vect{x}) \in \mathbb{R}$ and assigns treatments by sampling from the $\bernoulli \lpar \sigmoid (f(\vect x))\rpar$. Moreover, each DGP specifies two CEPO functions $\mu_0, \mu_1: \cX \to \R$. It then samples the potential outcomes by $y_t = \mu_t\lpar \vect x \rpar + \epsilon_t$ for $t \in \{0, 1\}$, where the noise terms $\epsilon_t \sim \mathrm{Normal}(0,1)$, $\mathrm{Laplace}(0,1)$, or $\mathrm{Uniform}(-1,1)$ with equal probability. We now describe each DGP family in more detail:

\begin{enumerate}[label=(\alph*),leftmargin=*]
\item \textbf{Polynomial.}  
      We first draw the number of features $d \sim \mathrm{Unif}\{10,\dots,20\}$ and sample covariate vectors $\vect{x} \sim\mathrm{Unif}[-2,2]^d$. We then fix a maximum degree $\deg \in \{1,2,3,4\}$, augment covariates with powers
      $\vect{x}_{\text{ext}}
          = (x_1,\dots,x_d,x_1^2,\dots,x_d^{\deg})$,
      sample weights
      $\vect{w}_{\mu_0}, \vect{w}_{\mu_1}, \vect{w}_T  \sim \mathrm{Unif}[-5,5]^{d\times\deg + 1}$, and
      define 
      \begin{equation}
          f(\vect{x})=\vect{w}_T^{\top}\vect{x}_{\text{ext}}, \quad \mu_t(\vect{x})=\vect{w}_{\mu_t}^{\top}\vect{x}_{\text{ext}}\; \text{ for }\; t \in \{0, 1\}.
      \end{equation}
      Degrees 1, 2, 3,  and 4 give the Linear, Quadratic, Cubic, and Quartic
      sub‑families; each degree adds new terms and is therefore a
      super-set of all lower degrees. We train on one degree family and test on the others to probe generalization.

\item \textbf{Sinusoidal.}  
      We draw the number of features $d \sim \mathrm{Unif}\{5,\dots,10\}$ and sample covariate vectors $\vect{x} \sim\mathrm{Unif}[-3,5]^d$. We then sample weight vectors
      $\vect{w}_{\mu_0}, \vect{w}_{\mu_1}, \vect{w}_T \sim  \mathrm{Unif}[-10,6]^d$, and a frequency $\omega \in \mathbb{R}^+$. We define the treatment logit function and the CEPOs as
      \begin{equation}
         f(\vect{x}) \;=\;
         \sin\lpar\omega\, \lcbar\vect{w}_T^{\top}\vect{x}\rcbar\rpar
         +\vect{w}_T^{\top}\vect{x},
      \quad
      \mu_t(\vect{x}) \;=\;
         \sin \lpar\omega\,\lcbar \vect{w}_{\mu_t}^{\top}\vect{x}\rcbar\rpar
         +\vect{w}_{\mu_t}^{\top}\vect{x} \; \text{ for }\; t \in \{0, 1\}.
      \end{equation}
      For training DGPs, we create three sub-families: Linear ($\omega = 0$), L1 ($\omega\!\in\![0,1]$) and 
      L2 ($\omega\!\in\!(1,2]$). For test-time DPGs, we use the following:
      Linear ($\omega = 0$),
      L1 ($\omega\!\in\![0.5,1]$),
      L2 ($\omega\!\in\!(1.5,2]$), and
      L3 ($\omega\!\in\!(2.5,3]$). This allows us to measure extrapolation to unseen frequencies. For example, an L2-trained model has seen DGPs from L1 and L2, but not L3.
\end{enumerate}
\xhdr{Synthetic Experiments on Sinusoidal} \cref{fig:ablation-sinusoid} shows both the regression curve $\widehat{cov}_\mu$ (orange) and the CATE curve $\widehat{cov}_\tau$. The model is overly confident in OOD scenarios (e.g., L2 tested on an L1 trained model) and either well-calibrated or conservative otherwise. The figure also shows that the regression curve is always below the blue CATE curve. Once calibration is done on the regression curve, as shown in \cref{fig:ablation-sinusoid-calibrated}, the $\mathrm{ICE}_\mu$ becomes smaller, resulting in a well-calibrated or conservative model, even on OOD scenarios.

\begin{figure}[t]
    \centering
    \includegraphics[scale=1.1]{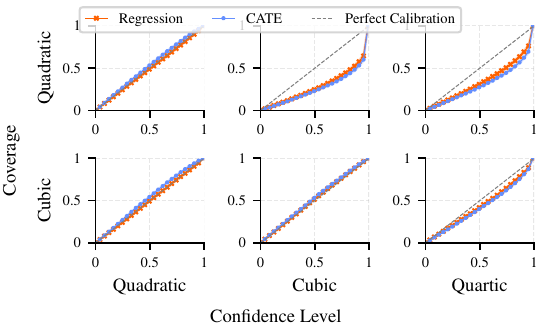}
    \caption{CATE and regression calibration curves for synthetic polynomial datasets, {\bf before calibration.} Models are trained on Quadratic/Cubic datasets and tested on Quadratic/Cubic/Quartic ones.}
    \label{fig:ablation-polynomial}
\end{figure}

\begin{figure}[t]
    \centering
    \includegraphics[scale=1.1]{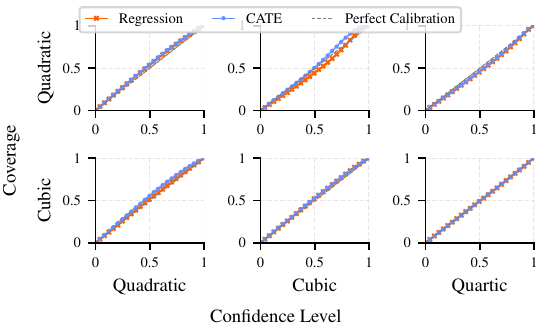}
    \caption{CATE and regression calibration curves for synthetic polynomial datasets, {\bf after calibration.} Models are trained on Quadratic/Cubic datasets and tested on Quadratic/Cubic/Quartic ones.}
    \label{fig:ablation-polynomial-calibrated}
\end{figure}

\xhdr{Synthetic Experiments on Polynomial} Similar to the sinusoidal setting, the uncalibrated curves in \cref{fig:ablation-polynomial} show that the model becomes overly confident when tested on OOD data (e.g., testing a model trained on Quadratic data on Cubic DGP). However, applying the regression calibration results in near-perfect CATE calibration, as shown in \cref{fig:ablation-polynomial-calibrated}.


\begin{figure}[t]
    \centering
    \includegraphics[scale=1.1]{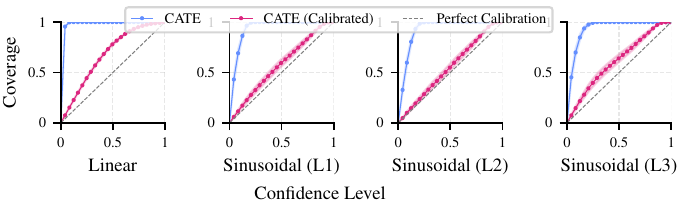}
    \caption{\method's CATE calibration on sinusoidal datasets, {\bf before and after calibration.}}
    \label{fig:causalpfn-sinusoid}
\end{figure}

\begin{figure}[t]
    \centering
    \includegraphics[scale=1.1]{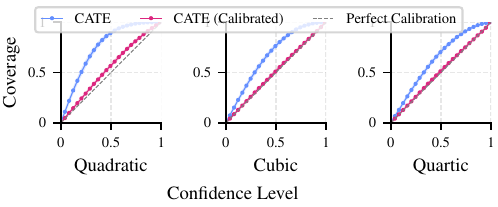}
    \caption{\method's CATE calibration on polynomial datasets, {\bf before and after calibration.}}
    \label{fig:causalpfn-polynomial}
\end{figure}

\begin{figure}[t]
    \centering
    \includegraphics[scale=1.1]{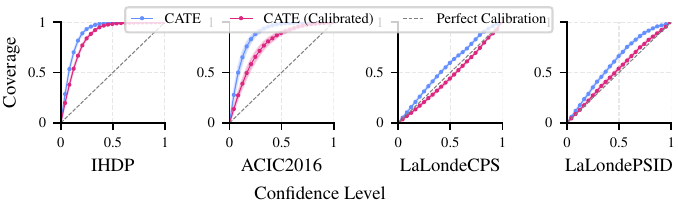}
    \caption{\method's CATE calibration on standard benchmarks, {\bf before and after calibration.}}
    \label{fig:causalpfn-benchmarks}
\end{figure}

\xhdr{Calibration of the Large-scale \method} We evaluate the calibration curves of the large-scale pre-trained \method\ on both synthetic and standard benchmarks in \cref{fig:causalpfn-sinusoid,fig:causalpfn-polynomial,fig:causalpfn-benchmarks}. The model generally appears conservative. This may be attributed to the Gaussian smoothing used in the histogram loss; yet, this smoothing is necessary to achieve stability in training. Regardless, across all datasets, post-hoc regression calibration improves reliability: the calibrated (pink) curves adhere far more closely to the diagonal than their uncalibrated (blue) counterparts.  In \cref{fig:causalpfn-sinusoid,fig:causalpfn-polynomial} the improvement is almost perfect, while in \cref{fig:causalpfn-benchmarks} it corrects the base model’s strong conservatism on IHDP and ACIC 2016 and achieves near-ideal alignment on the Lalonde datasets.

\clearpage
\section{Concurrent Work on PFNs for Causal Inference} \label{appx:concurrent}

Do-PFN~\citep{robertson2025pfn} is a concurrent approach that extends TabPFN to interventional queries by learning the interventional posterior predictive distribution, i.e., a distribution over $Y_t$ given $(\vect X,\data_\obs)$. In contrast, \method\ targets the \emph{expectation} of the interventional distribution (e.g., $\E\lbar Y_t \given \vect X{=}\vect x\rbar$), thus removing outcome (aleatoric) noise from the prediction target. This is especially relevant for uncertainty quantification, where we aim to isolate epistemic uncertainty about the causal effect.

More importantly, as described, Do-PFN does not explicitly enforce identifiability: the training prior can include \emph{observationally equivalent} DGPs (distinct processes with the same $\dist(\vect X,T,Y)$ but different effects). As formalized in \cref{prop:cepo-ppd-consistency}, if the training prior admits such cases, then any learner that conditions only on observational data cannot, in general, have its posterior predictive concentrate on the true effect, even with unlimited samples and model capacity. \method\ avoids this by constructing a prior that satisfies the ignorability (identifiability) condition, ensuring that CEPOs are functionals of $\dist_\obs$ (one effect per observational law). Empirically, \method\ outperforms Do-PFN on standard benchmarks in both PEHE and ATE relative error (Table~\ref{tab:do-pfn}).

\begin{table}[H]
\centering
\caption{Head-to-head comparison on benchmarks (mean $\pm$ SE; $\downarrow$ is better). For PEHE, Lalonde CPS/PSID values are reported $\times 10^{3}$.}
\label{tab:do-pfn}
\begin{adjustbox}{max width=\textwidth}
\begin{tabular}{llcccc}
\toprule
& & \textbf{IHDP} & \textbf{ACIC 2016} & \textbf{Lalonde CPS} & \textbf{Lalonde PSID} \\
\midrule
\multirow{2}{*}{\textbf{PEHE} ($\downarrow$)} 
& CausalPFN & $0.58\pm0.07$ & $0.92\pm0.11$ & $8.96\pm0.02$ & $14.40\pm0.20$ \\
& Do-PFN & $6.07\pm0.89$ & $4.11\pm0.52$ & $12.01\pm0.03$ & $20.91\pm0.14$ \\
\midrule
\multirow{2}{*}{\textbf{ATE Relative Error} ($\downarrow$)}
& CausalPFN & $0.20\pm0.04$ & $0.05\pm0.01$ & $0.13\pm0.01$ & $0.22\pm0.02$ \\
& Do-PFN & $0.57\pm0.10$ & $0.67\pm0.04$ & $0.87\pm0.01$ & $0.92\pm0.01$ \\
\bottomrule
\end{tabular}
\end{adjustbox}
\end{table}

\end{document}